\documentclass[11pt]{article}

\usepackage[utf8]{inputenc}

\usepackage{amsmath}
\usepackage{amsthm}
\usepackage{amssymb}
\usepackage{url}

\usepackage{fullpage}

\usepackage{appendix}
\usepackage{graphicx}
\usepackage{times}
\usepackage{ifpdf}
\usepackage{amsthm}
\usepackage{algorithm}
\usepackage[algo2e, ruled, vlined]{algorithm2e}
\newcommand{\ignore}[1]{}
\newcommand{\notinproc}[1]{#1}
\newcommand{\onlyinproc}[1]{}

\def\E{\textsf{E}}
\def\Exp{\textsf{Exp}}
\def\Lap{\textsf{Lap}}
\def\Var{\textsf{Var}}
\def\Bias{\textsf{Bias}}
\def\MSE{\textsf{MSE}}
\def\NRMSE{\textsf{NRMSE}}
\def\CDF{\textsf{CDF}}
\def\PDF{\textsf{PDF}}
\def\ekey{\textsf{key}}

\def\Zipf{\textsf{Zipf}}

\def\ekey{\mathop{\mathrm{key}}}
\newtheorem{thm}{Theorem}[section]
\newtheorem{theorem}[thm]{Theorem}

\newtheorem{lemma}[thm]{Lemma}

\newtheorem{infprob}[thm]{Informal\,\,Problem}

\newtheorem{definition}[thm]{Definition}

\newtheorem{corollary}[thm]{ Corollary}

\usepackage{xcolor} 

\SetCommentSty{mycommfont}

\renewcommand{\epsilon}{\varepsilon}

\title{Differentially Private Weighted Sampling}
\author{Edith Cohen \\
Google Research \\ Tel Aviv University
\and Ofir Geri  \\
Stanford University
\and Tamas Sarlos \\
Google Research
\and Uri Stemmer \\
Ben-Gurion University \\ Google Research
}

\date{}
\begin{document}
 \maketitle

\begin{abstract}
    Common datasets have the form of
    {\em elements} with {\em keys} (e.g., transactions and products)
    and the goal is to perform analytics on the aggregated form of {\em key} and {\em frequency} pairs.  A weighted sample of keys by (a function of) frequency is a highly versatile summary that provides a sparse set of representative keys and supports approximate evaluations of query statistics. 
    We propose {\em private weighted sampling} (PWS): A method that sanitizes a weighted sample as to ensure element-level differential privacy, while retaining its utility to the maximum extent possible.
    PWS maximizes the reporting probabilities of keys and estimation quality of a broad family of statistics. PWS improves over the state of the art even for the well-studied special case of {\em private histograms}, when no sampling is performed.  We empirically observe significant performance gains of 20\%-300\% increase in key reporting for common Zipfian frequency distributions and accurate estimation with $\times 2$-$ 8$ lower frequencies.
    PWS is applied as a post-processing of a non-private sample, without requiring the original data.  Therefore, it can be a seamless addition to existing implementations, such as those optimizes for distributed or streamed data. 
    We believe that due to practicality and performance, PWS may become a method of choice in applications where privacy is desired.
    \end{abstract}

\section{Introduction}

{\em Weighted sampling schemes} are often used to obtain versatile summaries of large datasets. The sample constitutes a representation of the data and also facilitates efficient estimation of many statistics. Motivated by the increasing awareness and demand for data privacy, in this work we construct {\em privacy preserving} weighted sampling schemes. The privacy notion that we work with is that of {\em differential privacy}~\cite{DMNS06}, a strong privacy notion that is considered by many researchers to be a gold-standard for privacy preserving data analysis.

Before describing our new results, we define our setting more precisely. Consider an input dataset containing $m$ elements, where each element contains a key $x$ from some domain $\mathcal{X}$. For every key $x\in \mathcal{X}$ we write $w_x$ to denote the multiplicity of $x$ in the input dataset. (We also refer to $w_x$ as the {\em frequency} of $x$ in the data.) With this notation, it is convenient to represent the input dataset in its {\em aggregated} form $D=\{(x,w_x) \}$   containing pairs of a key and its frequency $w_x\geq1$ in the data.
Examples of such datasets are plentiful: Keys are search query strings and elements are search requests, keys are products and elements are transactions for the products, keys are locations and elements are visits by individuals, or keys are training examples and elements are activities that generate them. We aim here to protect the privacy of data elements. These example datasets tend to be very sparse, where the number of distinct keys in the data is much smaller than the size $|\mathcal{X}|$ of the domain. Yet, the number of distinct keys can be very large and samples serve as small summaries that can be efficiently stored, computed, and transmitted. We therefore aim for our private sample to retain this property and in particular only include keys that are in the dataset.

The (non-private) sampling schemes we consider are specified by a (non-decreasing) sequence
$(q_i)_{i\geq 0}$ of probabilities $q_i\in [0,1]$, where $q_0 := 0$.
Such a sampling scheme takes an
input dataset $D=\{(x,w_x) \}$ and returns a {\em sample} $S\subseteq D$, where each pair $(x,w_x)$ is included in $S$ independently, with probability $q_{w_x}$. Loosely speaking, given a (non-private) sampling scheme $A$, we aim in this paper to design a {\em privacy preserving} variant of $A$ with the goal of preserving its ``utility'' to the extent possible under privacy constraints. We remark that an immediate consequence of the definition of differential privacy is that keys $x\in \mathcal{X}$ with very low frequencies cannot be included in the private sample $S$ (except with very small probability). On the other hand, keys with high frequencies can be included with probability (close to) $1$. 
Private sampling schemes can therefore retain more utility when the dataset has many keys with higher frequencies or for tasks that are less sensitive to low frequency keys.

\begin{infprob}\label{infprob}
Given a (non-private) sampling scheme $A$, specified by a sampling function $q$, design a {\em private} sampling scheme that takes a dataset $D=\{(x,w_x)\}$ 
and outputs a ``sanitized'' sample $S^* =\{(x,w^*_x)\}$. Informally, the goals are:
\begin{enumerate}
    \item Each pair $(x,w_x)\in D$ is sampled with probability ``as close as possible'' to the non-private sampling probability $q_{w_x}$.
    \item The sanitized sample $S^*$ 
    provides utility that is ``as close as possible'' to that of a corresponding non-private sample $S$.  In our constructions, the sanitized frequencies $w^*_x$ would be random variables from which we can estimate ordinal and linear statistics with (functions of) the frequency $w_x$.
\end{enumerate}
\end{infprob}


Informal Problem~\ref{infprob} generalizes one of the most basic tasks in the literature of differential privacy -- {\em privately computing histograms}. Informally, algorithms for private histograms take a dataset $D=\{(x,w_x)\}$ as input, and return, in a differentially private manner, a ``sanitized'' dataset $D^* =\{(x,w^*_x)\}$. 
It is often required that the output $D^*$ is {\em sparse}, in the sense that if $w_x=0$ then $w^*_x=0$. Commonly, we seek to minimize the expected or maximum error of estimators applied to $\boldsymbol{w}^*$ of statistics on  $\boldsymbol{w}$. One well-studied objective is to minimize $\max_{x\in X} \mid w_x-w^*_x|$. The work on private histograms dates all the way back to the paper that introduced differential privacy~\cite{DMNS06}, and it has received a lot of attention since then, e.g.,~\cite{KorolovaKMN09,HardtT10,BeimelBKN14,BeimelNS16,BunS16,BunNS19,BalcerV18,BunDRS18}. Observe that the private histogram problem is a special case of Informal Problem~\ref{infprob}, where $q\equiv1$.

At first glance, one might try to solve Informal Problem~\ref{infprob} by a reduction to the private histogram problem. Specifically, we consider the baseline where the data is first ``sanitized'' using an algorithm for private histograms, and then a (non-private) weighted sampling algorithm is applied to the sanitized data (treating the sanitized frequencies as actual frequencies).  This framework, of first sanitizing the data and then sampling it was also considered in \cite{CormodePST:ICDT2012}. We show that this baseline is sub-optimal, and improve upon it in several axes.

\subsection{Our Contributions}

Our proposed framework, {\em Private Weighted Sampling} (PWS), takes as input a non-private weighted sample $S$ that is produced  by a (non-private) weighted sampling scheme. 
We apply a ``sanitizer'' to the sample  $S$ to obtain a respective privacy-preserving sample $S^*$.  Our proposed solution has the following advantages.  

\paragraph{Practicality.}
The private version is generated from the sample $S$ as a post-processing step without the need to revisit the original dataset, which might be massive or unavailable. This means that we can augment existing implementations of non-private sampling schemes and retain their scalability and efficiency. This is particularly appealing for sampling schemes designed for massive distributed or streamed data that use small sketches and avoid a resource-heavy aggregation of the data~\cite{GM:sigmod98,EV:ATAP02,flowsketch:JCSS2014,AndoniKO:FOCS11,CCD:sigmetrics12,freqCapfill:TALG2018,JayaramW:Focs2018,CohenGeri:NeurIPS2019,CohenGeriPagh:ICML2020}. Our code is available at {\tiny \url{github.com/google-research/google-research/tree/master/private_sampling}}.

\paragraph{Benefits of end-to-end privacy analysis}
PWS achieves better utility compared to the baseline of first sanitizing the data and then sampling. 
In spirit, our gains follow from a well-known result in the literature of differential privacy stating that applying a differentially private algorithm on a random sample from the original data has the effect of boosting the privacy guarantees of the algorithm \cite{KamalikaMishra:Crypto2006,KasiviswanathanLNRS11,BunNSV15}. 
Our solution is derived from a precise {\em end-to-end} formulation of the privacy constraints that account for the benefits of the random sampling in our privacy analysis.  

\paragraph{Optimal reporting probabilities.}
PWS is optimal in that it maximizes  
the probability that each key $x$ is included in the private sample. The private reporting probability of a key $x$ depends on the privacy parameters, frequency, and sampling rate and is at most the non-private sampling probability $q_{w_x}$. The derivation is provided in Section~\ref{basic:sec}.

\paragraph{Estimation of linear statistics.}
Linear statistics according to a function of frequency have the form:
\onlyinproc{
  $\sum_x L(x) g(w_x)$,
}
\notinproc{
  \begin{equation}
  s := \sum_x L(x) g(w_x)\enspace ,
  \end{equation}
}
where $g(w_x) \geq 0$ is a non-decreasing function of frequency with $g(0):=0$.
The most common use case is when $L(x)$ is a predicate and $g(w) := w$ and the statistics is the sum of frequencies of keys that satisfy the selection $L$.
Our PWS sanitizer in Section~\ref{refined:sec} maintains optimal reporting probabilities and provides private information on frequencies of keys.
We show that generally differential privacy does not allow for unbiased estimators for statistics without significant increase in variance. We propose biased but nonnegative and low-variance estimators.    
\paragraph{Estimation of ordinal statistics.}
Ordinal statistics, such as (approximate) quantiles and top-$k$ sets, are derived from the order of keys that is induced by their frequencies. This order can be approximated by the order induced by PWS sanitized frequencies.  We show that PWS is optimal, over all DP sanitization schemes, for a broad class of ordinal statistics.  In particular, PWS maximizes the probability that {\em any} pair is concordant and maximizes the expected Kendall-$\tau$ rank correlation between the order induced by sanitized and true frequencies. 

\paragraph{Improvement over prior baselines.}
We show analytically and empirically in Section~\ref{experiments:sec} that we obtain orders of magnitude increase in reporting probability in low-frequency regimes. For estimation tasks, both PWS and prior schemes have lower error for higher frequencies but 
PWS obtains higher accuracy for frequencies that are $\times 2$-$8$ lower than prior schemes.  This is particularly helpful for datasets/selections with many mid-low frequency keys.

\paragraph{Improvement for private histograms.}
As an important special case of our results, we improve upon the state-of-the-art constructions for private (sparse) histograms~\cite{KorolovaKMN09,BunNS19}.  
These existing constructions obtain privacy properties by adding Laplace or Gaussian noise to the frequencies of the keys whereas we directly formulate and solve elementary constraints.
Let $\pi^*_i$ denote the PWS reporting probability of a key with frequency $i$, when applied to the special case of private histograms. Let $\phi_i$ denote the reporting probability of the state-of-the-art solution for private (sparse) histograms of~\cite{KorolovaKMN09,BunNS19}. Clearly $\pi^*_i$ is always at least $\phi_i$. We show that in low-frequency regimes we have $\pi^*_i/\phi_i \approx 2 i$.  Similarly for estimation tasks, PWS provides more accurate estimates in these regimes. 
Qualitatively, PWS and private histograms have high reporting probabilities and low estimation error for high frequencies.  But PWS significantly improves on 
low to medium frequencies, which is important for distributions with long tails.  We empirically show gains of 20\%-300\% in overall key reporting for Zipf-distributed frequencies.
As private histograms are one of the most important building blocks in the literature of differential privacy, we believe that our improvement is significant (both in theory and in practice).

\onlyinproc{Due to space limitations, we refer the reader to \cite{PWS:arxiv2020} for proofs and further details.}

\section{Related Work}

  The suboptimality of the Laplace mechanism for anonymization was noted by \cite{GhoshRS:sicomp2012}.  In our language, Ghosh et al.\ studied the non-sparse case, where all values, including $0$ values, can be reported with added noise. They did not consider sampling, and studied  pure differential privacy.  Instead of Laplace noise, they propose the use of a symmetric Geometric distribution and establish it is optimal for certain estimation tasks.  This can be viewed as a special case of what we do in that our schemes converge to that when there is no sampling, we use pure differential privacy, and when frequencies are large (so the effect of the sparse case constraint dissipates).
  Ghosh et al.\ establish the optimality of unbiased estimators for some frequency statistics when loss is symmetric. We show that bias is necessary in the sparse case and propose estimators that control the bias and variance.
  
  Key reporting was formulated and studied as {\em differentially private set union} problem \cite{GopiGJSSY:ICML2020}. They studied it without sampling, in a more general user privacy setting, and proposed a truncated Laplace noise mechanism similar to~\cite{KorolovaKMN09,BunNS19}.

  Recent independent work by \cite{desfontainesVG:2020} derived the optimal scheme for key reporting for sparse private histograms, a special case of our solution when there is no sampling.
\section{Preliminaries}

  We consider data in the form of a set of elements $\mathcal{E}$, where each element
$e\in\mathcal{E}$ has a key $e.\ekey \in \mathcal{X}$.  The {\em frequency} of a key $x$,  $w_x := \left|\{ e\in\mathcal{E} \mid e.\ekey=x\}\right|$, is defined as the number of elements with $e.\ekey = x$.  The {\em aggregated form} of the data, known in the DP literature as its {\em histogram}, is the set of key and frequency pairs $\{(x,w_x)\}$.  We use the vector notation $\boldsymbol{w}$ for the aggregated form. 
We will use $m:= |\mathcal{E}|$ for the number of elements and $n$ for the number of distinct keys in the data. 

\subsection{Weighted Sampling} \label{weightedsamplingprelim:sec}

We consider a very general form of without-replacement sampling schemes.  Each scheme is  specified by non-decreasing probabilities $(q_i)_{i\geq 1}$. The probability that a key is sampled depends on its frequency -- a key with frequency $i$ is sampled independently with probability $q_i$. Our proposed methods apply with any non-decreasing $(q_i)$.    

{\em Threshold sampling} is a popular class of weighted sampling schemes.  We review it for concreteness and motivation and use it in our empirical evaluation of PWS. A threshold sampling scheme (see Algorithm~\ref{alg:threshold}) is specified by  $(\mathcal{D},f,\tau)$, where $\mathcal{D}$ is a distribution, $f$ is a function of frequency, and $\tau$ is a numeric threshold value that specifies the sampling rate.
For each key we draw i.i.d.\ $u_x \sim \mathcal{D}$. The two common choices are
$\mathcal{D} = \Exp[1]$ for a probability proportional to size without replacement (ppswor) sample~\cite{Rosen1972:successive}  and $\mathcal{D} = U[0,1]$ for a Poisson Probability Proportional to Size (PPS) sample~\cite{Ohlsson_SPS:1990,Ohlsson_SPS:1998,DLT:jacm07}. 
A key $x$ is included in the sample if $u_x \leq  \tau f(w_x)$.
The probability that a key with frequency $i$ is sampled is 
\begin{equation} \label{qdef:eq}
    q_i := \Pr_{u\sim \mathcal{D}}[u < f(i)\tau]\enspace .
\end{equation}
\ignore{
A weighted sample can be used to estimate linear statistics of the form
 \[
 s := \sum_x L(x) g(w_x)\enspace .
 \]
 The per-key inverse-probability estimator~\cite{HT52} of $g(w_x)$ is defined as follows:
\begin{equation} \label{invprob:eq}
      a_{w_x} := \begin{cases}
 \frac{g(w_x)}{q_{w_x}} & x\text{ is included in the sample}\\
 0 & \text{otherwise}
 \end{cases}\enspace .
\end{equation}
   Our estimate of a query statistics will be the sum
 \[
 \hat{s} := \sum_{(x,w_x)\in S}  L(x) a_{w_x}\enspace .
 \]
 The estimate is unbiased, i.e., $\E[\hat{s}] = s$.
 }
  Threshold sampling is related to bottom-$k$ (order) sampling~\cite{Rosen1997a,Ohlsson_SPS:1990,DLT:jacm07,bottomk07:ds,bottomk:VLDB2008}  but instead of specifying the sample size $k$ we specify an inclusion threshold $\tau$.   
Ppswor is equivalent to drawing keys sequentially with probability proportional to $f(w_x)$.  The bottom-$k$ version stops after $k$ keys and the threshold version has a stopping rule that corresponds to the threshold. The bottom-$k$ version of Poisson PPS sampling is known as sequential Poisson or Priority sampling.  

\begin{algorithm2e}\caption{Threshold Sampling}\label{alg:threshold}
{\scriptsize 
\DontPrintSemicolon
\tcp{{\bf Threshold Sampler:}}
\KwIn{Dataset $\boldsymbol{w}$ of key frequency pairs $(x,w_x)$; distribution $\mathcal{D}$, function $f$, threshold $\tau$}
\KwOut{Sample $S$  of key-frequency pairs from $\boldsymbol{w}$}
\Begin{
$S \gets \emptyset$\;
\ForEach{$(x,w_x) \in \boldsymbol{w}$}
{
Draw independent $u_x\sim \mathcal{D}$\;
\If{$u_x < f(w_x)\tau$}{$S \gets S\cup\{(x,w_x)\}$
}
}
\Return{S}
}
}
\end{algorithm2e}

Since PWS applies a sanitizer to a sample, it inherits the efficiency of the base sampling scheme. Threshold sampling (via the respective bottom-$k$ schemes) can be implemented efficiently using small sketches
(of size expected sample size) on aggregated data that can be distributed or streamed \cite{DLT:jacm07,Rosen1997a,Ohlsson_SPS:1998,bottomk07:ds}.  On unaggregated datasets, it can be implemented 
using small sketches  for some functions of frequency including the moments $f(w)=w^p$ for $p\in [0,2]$ \cite{CCD:sigmetrics12,freqCapfill:TALG2018,CohenGeri:NeurIPS2019,CohenPW:NeurIPS2020}.

Our methods apply with a fixed threshold $\tau$. But the treatment extends to when the threshold is privately determined from the data. If we have a private approximation of the total count $\|f(\boldsymbol{w})\|_1 := \sum_x f(w_x)$  we can set $\tau \approx  k/\|f(\boldsymbol{w})\|_1$.  This provides (from the non-private sample that corresponds to the threshold) estimates with additive error $\|f(\boldsymbol{w})\|_1/\sqrt{k}$ for statistics with function of frequency $g=f$ and when $L$ is a predicate.  

\subsection{Differential Privacy}

The privacy requirement we consider is
{\em element-level} differential privacy.    Two datasets with aggregated forms
$\boldsymbol{w}$ and $\boldsymbol{w}'$ are neighbors if $\| \boldsymbol{w}-\boldsymbol{w}' \|_1 = 1$, that is, the frequencies of all keys but one are the same and the difference is at most $1$ for that one key.  The privacy requirements are specified using two parameters $\epsilon, \delta \geq 0$.


\begin{definition}[\cite{DMNS06}]
A mechanism $M$ is $(\varepsilon,\delta)$-differentially private if for any two neighboring inputs $\boldsymbol{w}$, $\boldsymbol{w}'$ and set of potential outputs $T$,
\begin{equation} \label{DP:eq}
\Pr[M(\boldsymbol{w}) \in T] \leq e^\varepsilon \Pr[M(\boldsymbol{w}') \in T] + \delta \enspace .
\end{equation}
\end{definition}

\begin{algorithm2e}\caption{Private Weighted Samples}\label{alg:sanitizer}
{\scriptsize
\DontPrintSemicolon
\tcp{{\bf Sanitized Keys:}}
\KwIn{$(\epsilon,\delta)$, weighted sample $S$, taken with non-decreasing probabilities $(q_i)_{i\geq 1}$}
\KwOut{Private sample of keys $S^*$}
Compute $(p_i)_{i\geq 1}$ 
\tcp*{Reporting probabilities per freq.}
\Begin(\tcp*[h]{Sanitize using scheme}){
$S^* \gets \emptyset$\;
\ForEach{$(x,w_x) \in S$}
{
With probability $p_{w_x}$, 
$S^* \gets S^* \cup \{x\}$
}
\Return{$S^*$}
}

\tcp{{\bf Sanitized keys and frequencies:}}
\KwIn{$(\epsilon,\delta)$, weighted sample $S$, taken with non-decreasing probabilities $(q_i)_{i\geq 1}$}  
\KwOut{Sanitized sample $S^*$}
Compute probability vectors $(p_{i\bullet})_{i\geq 1}$  
\tcp*{Reported values}
\Begin(\tcp*[h]{Sanitize using scheme}){
$S^* \gets \emptyset$\;
\ForEach{$(x,w_x) \in S$}
{
Draw $j \sim p_{w_x \bullet}$ \tcp*{By probability vector}
\If{$j>0$}{$S^* \gets S^* \cup \{(x,j)\}$}
}
\Return{$S^*$}
}

\tcp{{\bf Estimator:}}
\KwIn{Sanitized sample $S^*=\{(x,j_x)\}$, $\{\pi_{i,j}\}$ (where $\pi_{ij} := p_{ij} q_i$ , functions $g(i)$, $L(x)$}  
\KwOut{Estimate of the linear statistics $\sum_x L(x) g(x)$}
\Begin{
 Compute $(a_j)_{j\geq 1}$ using $\{\pi_{ij}\}$ and $g(i)$ \tcp*{Per-key estimates for $g()$}
 \Return{$\sum_{(x,j_x)\in S^*} L(x) a_{j_x}$} 
}
}
\end{algorithm2e}

\subsection{Private Weighted Samples} 

Given a (non-private) weighted sample $S$ of the data in the form of key and frequency pairs and (a representation) of the sampling probabilities $(q_i)_{i\geq 1}$ that guided the sampling, our
goal is to release as much of $S$ as we can without violating element-level differential privacy.

We consider two utility objectives.  The basic objective, {\em sanitized keys}, is to maximize the reporting probabilities of keys in $S$. The private sample in this case is simply a subset of the keys in $S$. The refined objective is to facilitate estimates of linear frequency and order statistics. The private sample includes sanitized keys from $S$ together with information on their frequencies.
The formats of the sanitizers and estimators are provided as Algorithm~\ref{alg:sanitizer}.  

\section{Sanitized Keys} \label{basic:sec}

\begin{algorithm2e}\caption{Compute $\pi_{i}$ for  Sanitizing Keys}\label{alg:reportkeys}
{\scriptsize 
\DontPrintSemicolon
\KwIn{$(\epsilon,\delta)$, non-decreasing sampling probabilities $(q_i)_{i\geq 1}$, $\text{Max$\_$Frequency}$}
$\pi_0 \gets 0$\;
\ForEach{$i=1,\ldots,\text{Max$\_$Frequency}$}
{
    $\pi_{i} \gets \min\{q_{i}, e^\varepsilon \pi_{i-1} + \delta, 1+ e^{-\varepsilon}(\pi_{i-1}+\delta-1) \}$
    }
    \Return{$(\pi_{i})_{i=1}^{\text{Max$\_$Frequency}}$}
}
\end{algorithm2e}

A sanitizer $C$ uses a representation of the non-decreasing $(q_i)_{i\geq 1}$  and computes respective probabilities $(p_i)_{i\geq 1}$.  A non-private sample $S$ can then be sanitized by considering each pair $(x,w_x)\in S$ and reporting the key $x$ independently with probability $p_{w_x}$.

We find it convenient to express constraints on $(p_i)_{i\geq 1}$ in terms of
the {\em end-to-end} reporting probability of a key $x$ with frequency $i$ (probability that $x$ is sampled and then reported):
\[\pi_i := p_i q_i =\Pr[x\in C(A(\boldsymbol{w}))] \enspace .\] 
Keys of frequency $0$ are not sampled or reported and we have $q_0=0$ and $\pi_0 := 0$.  The objective of maximizing $p_i$ corresponds to maximizing $\pi_i$.  We establish the following\notinproc{ (The proof is provided in Appendix~\ref{proofkeybasic:sec})}:
\begin{lemma} \label{basicpi:lemma}
Consider weighted sampling scheme $A$ where keys are sampled independently according to a non-decreasing $(q_i)_{i\geq 1}$ and a key sanitizer $C$ (Algorithm~\ref{alg:sanitizer}) is applied to the sample.  Then the probabilities $p_i \gets \pi_i/q_i$, where $\pi_i$ are the iterates computed in Algorithm~\ref{alg:reportkeys}, are each at the maximum under the DP constraints for $C(A())$.  Moreover, $(\pi_i)_{i\geq 1}$ is  non-decreasing.
\end{lemma}

\subsection{Structure and Properties of ${\boldsymbol{(\pi_i)_{i\geq 1}}}$} \label{keybasic:sec} 

The solution as computed in Algorithm~\ref{alg:reportkeys} applies with any non-decreasing $q_i$. We explore properties of the solution that allow us to compute and store it more efficiently and understand the reporting loss (reduction in reporting probabilities) that is due to the privacy requirement. \notinproc{Proofs are provided in Appendix~\ref{proofkeybasic:sec}.}

We provide closed-form expressions of the solution $\pi^*_i$ that corresponds to $q_i=1$ for all $i$ (aka the private histogram problem).  We will use the following definition of $L(\epsilon,\delta)$.  To simplify the presentation, we assume that $\epsilon$ and $\delta$ are such that $L$ is an integer (this assumption can be removed).
\begin{equation}\label{Ldef:eq}
 L(\epsilon,\delta) :=  \frac{1}{\varepsilon} \ln\left(
 \frac{e^\varepsilon -1 +2\delta}{\delta(e^\varepsilon +1)}
  \right) \approx \frac{1}{\varepsilon} \ln\left(\frac{\min\{1,\varepsilon/2\}}{\delta} \right)
\end{equation}
\begin{lemma} \label{privatehist:lemma}
When $q_i=1$ for all $i$, the sequence $(\pi_i)_{i\geq 1}$ computed by Algorithm~\ref{alg:reportkeys} has the form:
\begin{equation} \label{pistar:eq}
    \pi^*_i = \left\{ \begin{array}{ll} 
     \delta \frac{e^{\varepsilon i}-1 }{e^\varepsilon -1} &\; \text{\small $i\leq L+1$  }\\
     1- \delta  \frac{e^{\varepsilon (2L+2-i)} -1}{e^{\varepsilon} -1} &\; \text{\small $L+1 \leq i \leq 2L+1$  }\\
    1 &\; \text{\small $i \geq 2L+2$  }
    \end{array}\right.
\end{equation}


\end{lemma}

For the general case where the $q_i$'s can be smaller than 1, we bound the number of frequency values for which $\pi_i < q_i$. On these frequencies, the private reporting probability is strictly lower than that of the original non-private sample, and hence there is reporting loss due to privacy.
\begin{lemma} \label{totaltwothree:lemma}
There are at most $2L(\varepsilon,\delta)+1$ values $i$ such that $\pi_i < q_i$, where
$L$ is as defined in \eqref{Ldef:eq}. 
\end{lemma}

 We now consider the structure of the solution for threshold sampling.  The solution has a particularly simple form that can be efficiently computed and represented. 
 \begin{lemma} \label{niceq:lemma}
When the sampling probabilities $(q_i)_{i\geq 1}$ are those of threshold ppswor sampling with $f(i)=i$  then the solution has the form $\pi_i = \pi^*_i$ for $i< \ell$ and $\pi_i = q_i$ for $i\geq \ell$, where $\ell = \min\{i :  \pi^*_i > q_i\}$ is the lowest position with $\pi^*_i > q_i$ and $\pi^*_i$ is as defined in \eqref{pistar:eq}.
\end{lemma}

\section{Sanitized Keys and Frequencies} \label{refined:sec}

\begin{algorithm2e}\caption{Compute $(\pi_{i,j})$ for  Sanitized  Frequencies}\label{alg:reportfreq}
{\scriptsize 
\DontPrintSemicolon
\KwIn{($(\epsilon,\delta)$, non-decreasing $(q_i)_{i\geq 1}$), $\text{Max\_Frequency}$} 
\KwOut{$(\pi_{i,j})$ for $0\leq j\leq i \leq \text{Max\_Frequency}$}
$\pi_{0,0} \gets 1$, $\pi_0=0$\;
\ForEach(\tcp*[h]{Iterate over rows}){$i= 1,\ldots, \textrm{Max\_Frequency}$}{
$\pi_{i} \gets \min\{q_{i}, e^\varepsilon \pi_{i-1} + \delta, 1+ e^{-\varepsilon}(\pi_{i-1}+\delta-1) \}$\tcp*{End-to-end probability to output a key with frequency $i$}
$\pi_{i,0} \gets 1- \pi_{i}$\;
\ForEach(\tcp*[h]{Set lower bound values; use $[a]_+ :=\max\{a,0\}$}){$j=1,\ldots,i-1$}
{
$\pi_{i,j} \gets e^{-\varepsilon} \left(\sum_{h=1}^j \pi_{i-1,h} - \delta\right) - \sum_{h=1}^{j-1} \pi_{i,h}$ \\ $+ \left[ e^{-\varepsilon} \pi_{i-1,0}-\pi_{i,0}\right]_+$ \; 
$\pi_{i,j} \gets [\pi_{i,j}]_+$\;
}
$R \gets \pi_i - \sum_{h=1}^{i-1} \pi_{i,h}$ \tcp*{Remaining probability to assign}\;
\ForEach(\tcp*[h]{Set final values for $\pi_{i,j}$}){$j=i,\ldots,1$}
{
\lIf{$R=0$}{{\bf Break}}
$U \gets e^\varepsilon \sum_{h=j}^{i-1}\pi_{i-1,h} + \delta -\sum_{h=j+1}^i \pi_{i,h}$ \tcp*{Max value allowed for $\pi_{i,j}$}\; 
\eIf{$U-\pi_{i,j} \leq R$}{
$R\gets R- (U-\pi_{i,j})$\;
$\pi_{i,j} \gets U$
}
{
$\pi_{i,j} \gets \pi_{i,j}+R$\;
$R \gets 0$
}
}
}
\Return{$(\pi_{i,j})$}
}
\end{algorithm2e}

A frequency sanitizer $C$ returns keys $x$ together with sanitized information on their frequency.
We use $p_{i,j}$ for the probability that $C$ reports $j\in[t]$ for a sampled key that has frequency $i$, with $p_{i,0}$ being the probability that the sampled key is not reported. We have that
$\sum_{j=1}^{t} p_{i,j}$ is the total probability that a sampled key with frequency $i$ is reported by $C$. We use 
\[\pi_{i,j} \gets q_i p_{i,j}\]
for the end-to-end probability that a key with frequency $i$ is sampled and reported in the private sample with sanitized value $j$.  For notation convenience, we use
$\pi_{i,0} := 1-\sum_{j=1}^t \pi_{i,t}$ for the probability that a key is not reported, making $\pi_{i,\bullet}$ probability vectors.
The reader can interpret the returned value $j$ as a token from an ordered domain.  The estimators we propose depend only on the order of tokens and not their values and hence are invariant to a mapping of the domain that preserves the order. 


We express constraints on $(\pi_{i,j})_{i\geq 0,j\geq 0}$. For a solution to be  {\em realizable}, we must have end-to-end reporting probabilities that do not exceed the sampling probabilities: 
\begin{equation} \label{qbound2:eq}
    \forall i,\ \sum_{j=1}^{t} \pi_{i,j} \leq q_i\enspace .
\end{equation}
The DP constraints are provided in the sequel. Note that we must have $\sum_{j=1}^{t} \pi_{i,j}\leq \pi_i$, where $(\pi_i)_{i\geq 1}$ is the solution for sanitized keys (Algorithm~\ref{alg:reportkeys}), this because the sanitized frequencies DP constraints are a superset of the sanitized keys constraints -- we obtain the latter in the former by considering outputs that group together all outputs with a key $x$ with all possible values of $j >0$. For optimality, 
we seek solutions that (informally) {\em maximally separate} the distributions of different frequencies (minimize the overlap), over all possible DP frequency reporting schemes. 
We will see that maximum separation can (i)~always be achieved by a discrete distribution (when the maximum frequency is bounded) and (ii)~can be simultaneously achieved between any pair of frequencies.  In particular, we maintain optimal reporting, that is, $\sum_{j=1}^{t} \pi_{i,j} = \pi_i$ and $\pi_{i,0} =  \overline{\pi}_i$.  The solutions we express are such that for
$i_1 > i_2$, $\pi_{i_1,\bullet}$ (first-order) stochastically dominates $\pi_{i_2,\bullet}$: That is, for any $h$, the probability of a token $j\geq h$  is non-decreasing with frequency.  

  We present two algorithms that express $(\pi_{i,j})$.  
    Algorithm~\ref{alg:reportfreq} provides a simplified construction, where
$t$ is equal to the maximum frequency and we always report $j\leq i$ for a key with frequency $i$. The sanitizer satisfies realizability and DP and has optimal key reporting but attains maximum separation only under some restrictions. 
The values $\pi_{i,j}$ are specified in order of increasing $i$, where the row $\pi_{i,\bullet}$ is set so that the probability mass of $\pi_i$ is pushed to the extent possible to higher $j$ values. 
  
  Algorithm~\ref{alg:Creport} specifies PDFs $(f_i)$ for a frequency sanitizer. The PDFs have a discrete point mass at $0$ (that corresponds to the probability of not reporting) and are piecewise constant elsewhere. The scheme is a refinement of the scheme of Algorithm~\ref{alg:reportfreq} and, as we shall see, for any $(q_i)$ and $(\varepsilon,\delta)$, it maximally  separates sanitized values for different frequencies. The construction introduces at most $3m$ distinct breakpoints for frequencies up to $m$ and can be discretized to have an equivalent $(\pi_{i,j})$ form with $j\in [3m]$.
 \notinproc{ (More details and proofs are provided in Appendix~\ref{general:sec}.)}   

 \begin{theorem}\label{theorem:reported-freq-dist}
  The sanitizer with $(\pi_{i,j})$ expressed in Algorithm~\ref{alg:Creport} satisfies: 
  \begin{enumerate}
      \item $\forall i,\, \sum_{j=1}^i \pi_{i,j} = \pi_i$, and in particular,  \eqref{qbound2:eq} holds and the sanitizer is realizable.
      \item $(\varepsilon,\delta)$-DP
      \item \label{separation:cond} Maximum separation:  For each $i$, there is an index $c_i$ so that subject to the above and to row $\pi_{i-1,\bullet}$, for all $j'\leq c_i$, the sum $\sum_{j=1}^{j'} \pi_{ij}$ is at a minimum and for all $j'\geq c_i$, the sum $\sum_{j=j'}^{i} \pi_{ij}$ is at a maximum.

  \end{enumerate}
       \end{theorem}
       
  The $(\pi_{i,j})$ expressed by Algorithm~\ref{alg:reportfreq}  satisfy maximum separation (Property~\ref{separation:cond}) under the particular restrictions on the reported values (that only $i$ different outputs are possible for frequencies up to $i$).  \notinproc{ (The proofs are provided in Appendix~\ref{proofssanitizedKF:sec})}
   
   The $(\pi_{i,j})$ expressed by Algorithm~\ref{alg:Creport} and then discretized satisfy maximum separation (Property~\ref{separation:cond}) unconditionally, over all DP frequency sanitization schemes.


For the special case where $q_i=1$ for all $i$, Algorithm~\ref{alg:reportfreq} provides maximum separation. We 
provide a closed-form expression for the solution $\pi^*_{i,j}$.
 \begin{lemma} \label{formij:lemma}
 Let the DP parameters $(\varepsilon,\delta)$ be such that 
 $L(\varepsilon,\delta)$ as in \eqref{Ldef:eq} is integral.
Let $(\pi^*_{ij})$ be the solution computed in Algorithm~\ref{alg:reportfreq} for $q_i=1$ for all $i$. Then the matrix with entries $\pi^*_{ij}$ for $i,j\geq 1$ has a  lower triangular form, with the non-zero entries as follows:
\[
\text{For } j\in \{\max\{1, i-2L\},\ldots, i\},\ \pi^*_{ij} = \pi^*_{i-j+1} - \pi^*_{i-j}\enspace .
\]
Equivalently,
$
  \pi^*_{i,j} =
\begin{cases}
\delta e^{(i-j)\varepsilon} &\,  \text{if $0\leq i-j\leq L$}\\
\delta e^{(2L-(i-j))\varepsilon}   &\,  \text{if $L+1\leq i-j \leq 2L$ .}
\end{cases}
$
\end{lemma}

\begin{algorithm2e}\caption{Compute $(f_i)$ for  Sanitizing Frequencies}\label{alg:Creport}
{\scriptsize 
\DontPrintSemicolon
\KwIn{$(\epsilon,\delta)$, non-decreasing sampling probabilities $(q_i)_{i\geq 1}$, $\text{Max$\_$Frequency}$ }
\KwOut{$(f_i)_{i=0}^{\text{Max$\_$Frequency}} $, where $f_i:[0,i]$ 
\tcp*{PDF of sanitized frequency for frequency $i$: discrete mass at $f_i(0)$ (probability of not reporting) and density on $(0,i]$}}
$f_0(0) \gets 1$; $\pi_i \gets 0$ \tcp*{Keys with frequency $0$ are never reported}
\For(\tcp*[h]{Specify $f_i:[0,i]$}){$i\gets 1$ \KwTo $\text{Max$\_$Frequency}$}
{
 $\pi_{i} \gets \min\{q_{i}, e^\varepsilon \pi_{i-1} + \delta, 1+ e^{-\varepsilon}(\pi_{i-1}+\delta-1) \}$;  $f_i(0) \gets 1-\pi_i$ \tcp*{Reporting probability for $i$}
 $f_{i}(i-1,i] \gets \min\{\pi_i,\delta\}$\; 
\tcp{Represent a function $f_L:(0,i-1]$ that "lower bounds" $f_i$}
\If {$\max\{0,e^{-\varepsilon} f_{i-1}(0) - f_i(0)\} + \int_{0^+}^{i-1} f_{i-1}(x) dx \leq \delta$}
{$f_L(0,i-1]\gets 0$}
\Else
{
$\mathit{b}_i \gets$ $z$ that solves
       $ \max\{0,e^{-\varepsilon} f_{i-1}(0) - f_i(0)\}+\int_{0^+}^z f_{i-1}(x) dx = \delta$ \tcp*{Well defined, as from DP, we always have $\max\{0,e^{-\varepsilon} f_{i-1}(0) - f_i(0)\}\leq \delta$} 
$f_L(0,b_i]\gets 0$\;
\lFor{$x\in (\mathit{b}_i,i-1]$}{$f_L(x) \gets e^{-\varepsilon}f_{i-1}(x)$
}}
\tcp{Point where $f_i(x)-f_{i-1}(x)$ switches sign} 
        $c_i \gets$ $z$ that solves
$\int_{0}^{z} f_L(x)dx + e^\varepsilon \int_z^{i-1} f_{i-1}(x) dx = \pi_i -\min\{\pi_i,\delta\}$ \tcp*{Any solution $z\in (0,i-1]$}      
    \lFor{$x\in (c_i,i-1]$}{$f_i(x)= e^\varepsilon f_{i-1}(x)$}
    \lFor{$x \in (0,c_i]$}{$f_{i}(x) \gets f_L(x)$}
    
}
}
\Return{$(f_i)_{i=0}^{\text{Max$\_$Frequency}}$}
\end{algorithm2e}

\section{Estimation of Ordinal Statistics}

The sanitized frequencies can be used for estimation of statistics specified with respect to the actual frequencies.  In this section we consider {\em ordinal} statistics, that only depend on the order of frequencies but not on their nominal values.  Ordinal statistics include (approximate) top-$k$ set, quantiles, rank of a key, set of keys with a higher (or lower) rank than a specified key, and more.  We approximate ordinal statistics from the ordering of keys that is induced by sanitized frequencies.  The quality of estimated ordinal statistics is determined by the match between the order induced by exact frequencies and the order induced by sanitized frequencies. We say that the two orders are {\em concordant} on 
a subset of keys $\{x_i\}$, when they match on that subset. Since the output of our sanitizer is stochastic, we consider the probability of a subset being concordant.  When sanitized values are discrete and two keys have the same sanitized value, we use probability of $0.5$ that two keys are concordant.

We define \notinproc{(see Appendix~\ref{maxseparation:sec})} a measure of separation between distributions $f_{i_1}$ and $f_{i_2}$ at a certain quantile value $\alpha$ and show that the $(f_i)$ constructed by Algorithm~\ref{alg:Creport} maximize it pointwise for any $i_1,i_2,\alpha$. This measure generalizes and follows from Property~\ref{separation:cond} stated in Theorem~\ref{theorem:reported-freq-dist}. 
As a corollary we show\notinproc{ (The proof is provided in Appendix~\ref{maxseparation:sec})}:
\begin{corollary} \label{ordinalstat:coro}
The sanitizing scheme specified by the $(f_i)$ computed by Algorithm~\ref{alg:Creport} maximizes the following: 
The probability that a subset of keys is concordant, the probability that a 
key is correctly ordered with respect to all other keys, and the expected Kendall-$\tau$ rank correlation.
\end{corollary}
Note that we get optimality in a strong sense -- there is no Pareto front where concordant probability on some pairs of frequencies needs to be reduced in order to get a higher value for other pairs.

\section{Estimation of Linear Frequency Statistics}
 
The objective is to estimate statistics of the form
 \begin{equation} \label{qstat:eq}
     s := \sum_x L(x) g(w_x)\enspace .
 \end{equation}
We briefly review estimators for the non-private setting where the sample consists of pairs $(x,w_x)$ of keys and their frequency.  We use the per-key inverse-probability estimators~\cite{HT52} (also known as importance sampling).  The estimate $\widehat{g(w_x)}$ of $g(w_x)$ is $0$ if key $x$ is not included in the sample and otherwise the estimate is 
\onlyinproc{$a_{w_x} := \frac{g(w_x)}{q_{w_x}}$.}
\notinproc{
  \begin{equation} \label{invprob:est}
    a_{w_x} := \frac{g(w_x)}{q_{w_x}} \enspace .
  \end{equation}
}
 These estimates are nonnegative, a desired property for nonnegative values, and are also unbiased when $q_{w_x}>0$.   The estimate for the query statistics \eqref{qstat:eq} is
 \begin{equation} \label{sumest:eq}
     \hat{s} := \sum_{(x,w_x)} L(x) \widehat{g(w_x)} =  \sum_{(x,w_x)\in S}  L(x) a_{w_x}\enspace .
 \end{equation}
\notinproc{Since the estimate is $0$ for keys not represented in the sample, it can be computed from the sample.
 The variance of a per-key estimate for a key with frequency $i$ is $g(i)^2(\frac{1}{q_i} -1)$ and the variance of the sum estimator \eqref{sumest:eq} is
  \[ \Var[\hat{s}] = \sum_x L(x)^2 g(w_x)^2(\frac{1}{q_{w_x}} -1) \enspace .\]
  These inverse-probability estimates are optimal for the sampling scheme in that they minimize the sum of per-key variance under unbiasedness and non-negativity constraints.
   We note that the quality of the estimates depends on the match between $g(i)$ and $q_i$: Probability Proportional to Size (PPS), where $q_i\propto g(i)$ is most effective and minimizes the sum of per-key variance for the sample size. Our aim here is to optimize what we can do privately when $q$ and $g$ are given.}

  \subsection{Estimation with Sanitized Samples} \label{estssanitized:sec}
  We now consider estimation from sanitized samples $S^*$. 
We specify our estimators $(a_j)_{j\geq 1}$ in terms of the reported sanitized frequencies $j$.  The estimate is $0$ for keys that are not reported and are $a_j$ when reported with value $j$. The estimate of the statistics is
\begin{equation} \label{partialfest:eq}
        \widehat{s} := \sum_{(x,j)\in S^*}  L(x) a_{j}\ .
\end{equation}
 
As for choosing $(a_j)_{j\geq 1}$, a first attempt is the unique unbiased estimator: 
The unbiasedness constraints
\onlyinproc{$\forall i,\ \sum_{j=1}^i \pi_{ij} a_j = g(i)$}\notinproc{
  \[
 \forall i,\ \sum_{j=1}^i \pi_{ij} a_j = g(i)\enspace 
 \]}
 form a triangular system with a unique solution $(a_j)_{j\geq 1}$\onlyinproc{.}\notinproc{:
 \[a_i \gets \frac{g(i) - \sum_{j=1}^{i-1} \pi_{i,j} a_j}{\pi_{i,i}} \enspace .\]}
However, $(a_j)_{j\geq 1}$ may include negative values and estimates have high variance. We argue that bias is unavoidable with privacy:
First, the inclusion probability of keys with frequency $w_x=1$ can not exceed $\delta$. Therefore, the variance contribution of the key to any unbiased estimate is at least $1/\delta$.  Typically, $\delta$ is chosen so that  $1/\delta \gg n k$, where $n$ is the support size and $k$ the sample size, so this error can not be mitigated.  Second, we show \onlyinproc{in the full version}\notinproc{in Appendix~\ref{mustnegative:sec}} that even for the special case of $q=1$, {\em any} unbiased estimator applied to the output of  {\em any} sanitized keys and frequencies scheme with optimal reporting probabilities must assume negative values. That is,
DP schemes do not admit unbiased nonnegative estimators without compromising reporting probabilities. We therefore seek estimators that are biased but balance bias and variance and are nonnegative. In our evaluation we use the following Maximum Likelihood estimator (MLE):
\begin{equation} \label{MLest:eq}
    a_j \gets \frac{g(i)}{\pi_i}\text{, where } i =   \arg\max_h \pi_{hj} \enspace .
\end{equation}

This estimate is "right" for the frequency $i$ for which the probability of reporting $j$ is maximized.\notinproc{ The estimate can be biased up or down.}
\notinproc{Another estimator with desirable properties is proposed in Appendix~\ref{biaseddown:sec}.}

 We express the expected value, bias, Mean Squared Error (MSE), and variance of the per-key estimate for a key with frequency $i$:
 \onlyinproc{
 $\E_i := \sum_{j=1}^i  \pi_{i,j} a_j$, $\Bias_i := \E_i - g(i)$, $\MSE_i := \overline{\pi}_i g(i)^2 +  \sum_{j=1}^i  \pi_{i,j} (a_j - g(i))^2$, $\Var_i := \MSE_i - \Bias_i^2$. For the sum estimate \eqref{partialfest:eq} we get:
   $\Bias[\widehat{s}] = \sum_x L(x) \Bias_{w_x}$, $\Var[\widehat{s}] = \sum_x L(x)^2 \Var_{w_x}$,
   $\MSE[\widehat{s}] = \Var[\widehat{s}] + \Bias[\widehat{s}]^2$, and $\NRMSE[\widehat{s}] = \frac{\sqrt{\MSE[\widehat{s}]}}{s}$. 
       }
 \notinproc{      
 {\small
   \begin{align*}
   \E_i :=&\, \sum_{j=1}^i  \pi_{i,j} a_j\\
       \Bias_i :=&\, \E_i - g(i)\\
       \MSE_i :=&\, \overline{\pi}_i g(i)^2 +  \sum_{j=1}^i  \pi_{i,j} (a_j - g(i))^2 \\  
       \Var_i :=&\, \MSE_i - \Bias_i^2
    \enspace .
   \end{align*}
   }
 For the sum estimate \eqref{partialfest:eq} we get:
{\small
   \begin{align}
    \Bias[\widehat{s}] =&\, \sum_x L(x) \Bias_{w_x}\nonumber\\
    \Var[\widehat{s}] =&\, \sum_x L(x)^2 \Var_{w_x}\nonumber\\
    \MSE[\widehat{s}] =&\, \Var[\widehat{s}] + \Bias[\widehat{s}]^2\nonumber\\
    \NRMSE[\widehat{s}] =& \frac{\sqrt{\MSE[\widehat{s}]}}{s}\enspace . \label{NRMSE:eq}
 \end{align}
}}
Note that
the variance component of the normalized squared error $\MSE[\widehat{s}]/s^2$ decreases linearly with support size whereas the bias component may not. 
We therefore consider both the variance and bias of the per-key estimators and 
qualitatively seek low bias and ``bounded'' variance. We measure quality of statistics estimators using the Normalized Root Mean Squared Error (NRMSE).

  \section{Performance Analysis} \label{experiments:sec}
  
    We study the performance of PWS on the key reporting and estimation objectives and compare with a baseline method that provides the same privacy guarantees.  We use precise expressions (not simulations) to compute probabilities, bias, variance, and MSE of the different methods.
  
    \subsection{Private Histograms Baseline} \label{SbH:sec}
  
   We review the {\em Stability-based Histograms} (SbH) method of \cite{BunNS19,KorolovaKMN09,VadhanDPsurvey:2017}, which we use as a baseline. SbH, provided as Algorithm~\ref{alg:SBH},  is designed for the special case when $q_i=1$ for all frequencies. The input $S$ is the full data of pairs of  keys and positive frequencies $(x,w_x)$. The private output $S^*$ is a subset of the keys\notinproc{ in the data} with positive sanitized frequencies $(x,w^*_x)$.  
   \begin{algorithm2e}\caption{\small Stability-based Histograms (SbH) 
   }\label{alg:SBH}
{\small
\DontPrintSemicolon
\KwIn{$(\epsilon,\delta)$ , $S=\{(x,w_x)\}$ where $w_x>0$}
\KwOut{Key value pairs $S'$}
$S^*\gets \emptyset$\tcp*{Initialize private histogram}
$T \gets (1/\varepsilon)\ln(1/\delta)+1$\tcp*{Threshold}
\ForEach{$(x,w_x)\in S$}
{
    $w^*_x \gets w_x + \Lap[\frac{1}{\varepsilon}]$\tcp*{Add Laplace random variable}
    \If{$w^*_x \geq T$}{$S^*\gets S^* \cup (x,w^*_x)$ }
    }
}
\end{algorithm2e}

   
The SbH method is considered the state of the art for sparse histograms (only keys with $w_x > 0$ can be reported). The method returns non-negative $w_x^*>0$  sanitized frequencies.
For the case of no sampling, we compare PWS (with $q\equiv 1$) with SbH. We use the SbH sanitized frequencies directly for estimation.
For sampling, our baseline is {\em sampled-SbH}: The data is first sanitized using SbH and then sampled, using a weighted sampling algorithm with $q$, while treating the sanitized frequencies as actual frequencies. For estimation, we apply the estimator \eqref{sumest:eq} (which in this context is biased).
\notinproc{
To facilitate comparison with SbH and sampled-SbH we express the reporting probabilities, bias, and variance in Appendix~\ref{expressSbH:sec}.}

\begin{figure}[ht]
\centering
\includegraphics[width=0.44\textwidth]{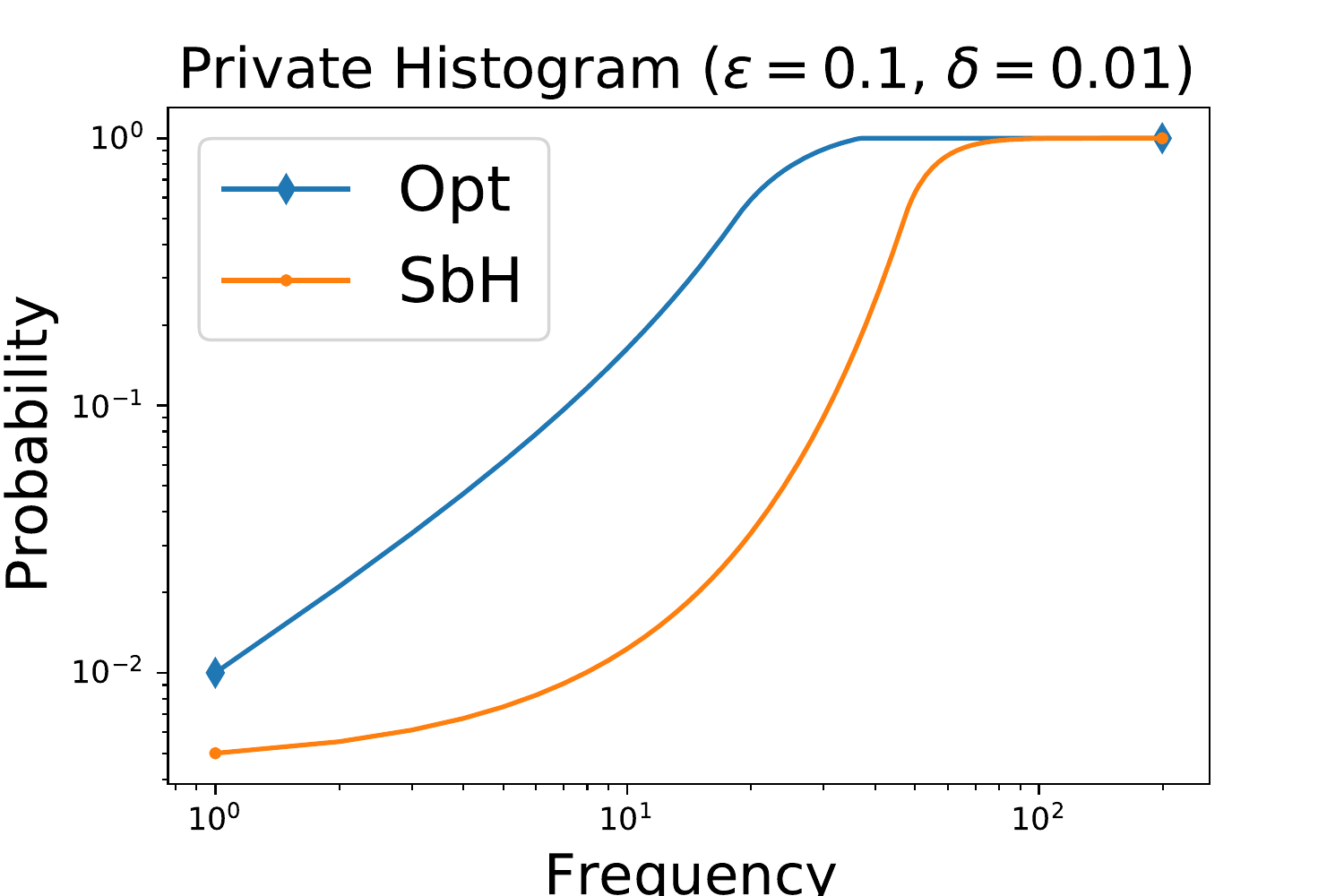}
\includegraphics[width=0.44\textwidth]{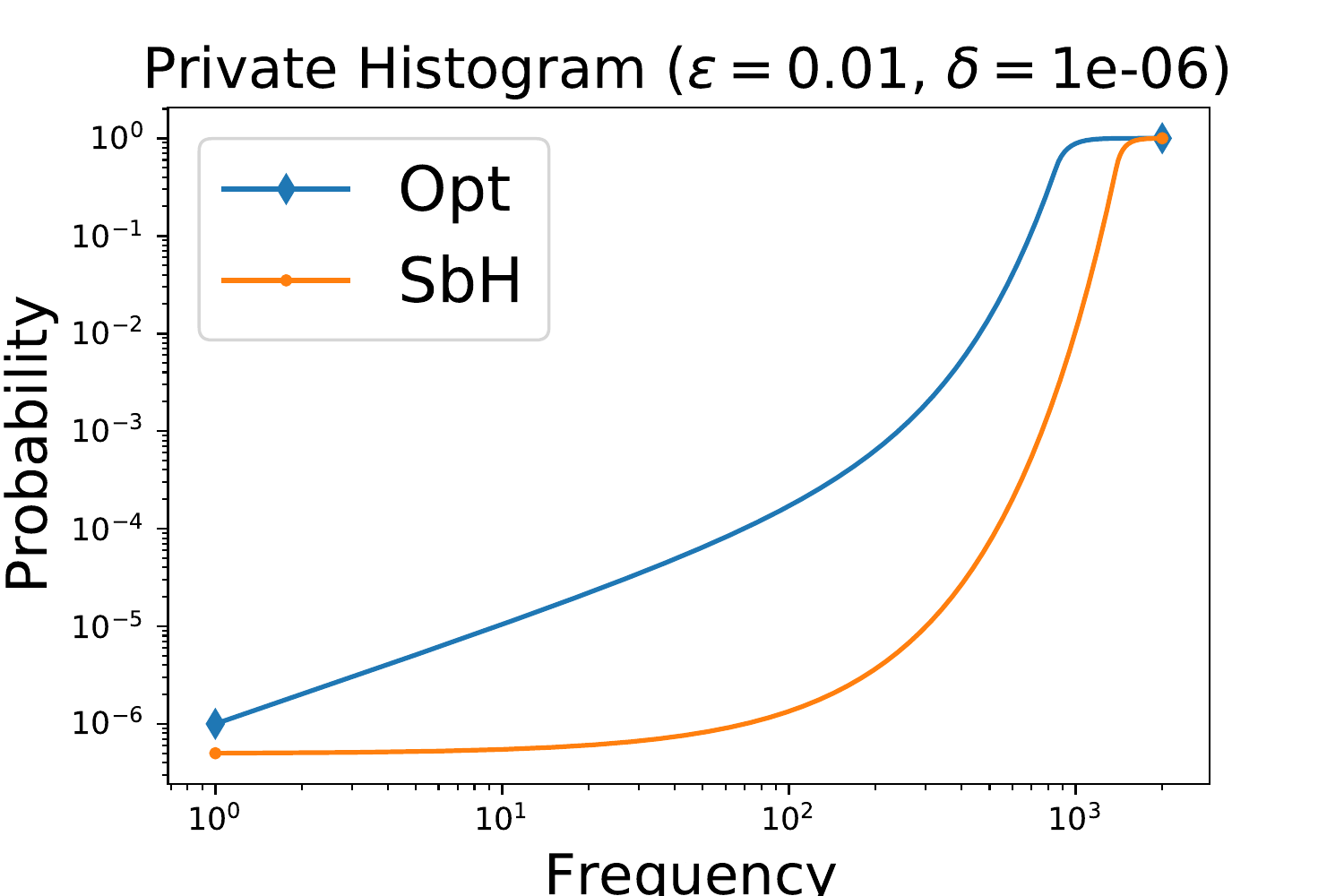}
\caption{Key reporting probability for frequency. No sampling ($q=1$) with PWS (Opt) and SbH for $(\varepsilon,\delta)=(0.1,0.01),(0.01,10^{-6})$}
\label{probfreqhist:plot}
\end{figure}
\begin{figure}[ht]
\centering
\includegraphics[width=0.30\textwidth]{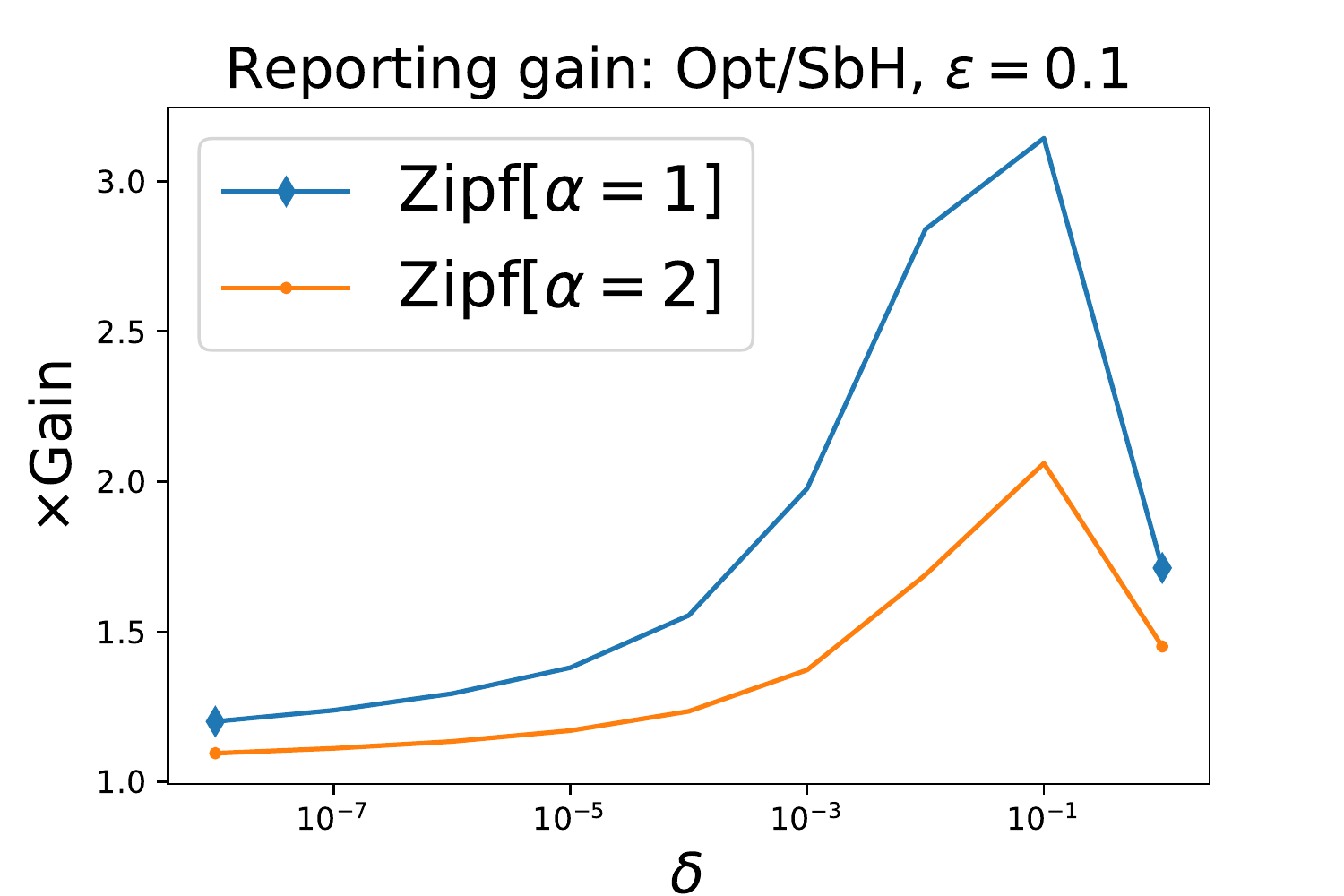}
\includegraphics[width=0.30\textwidth]{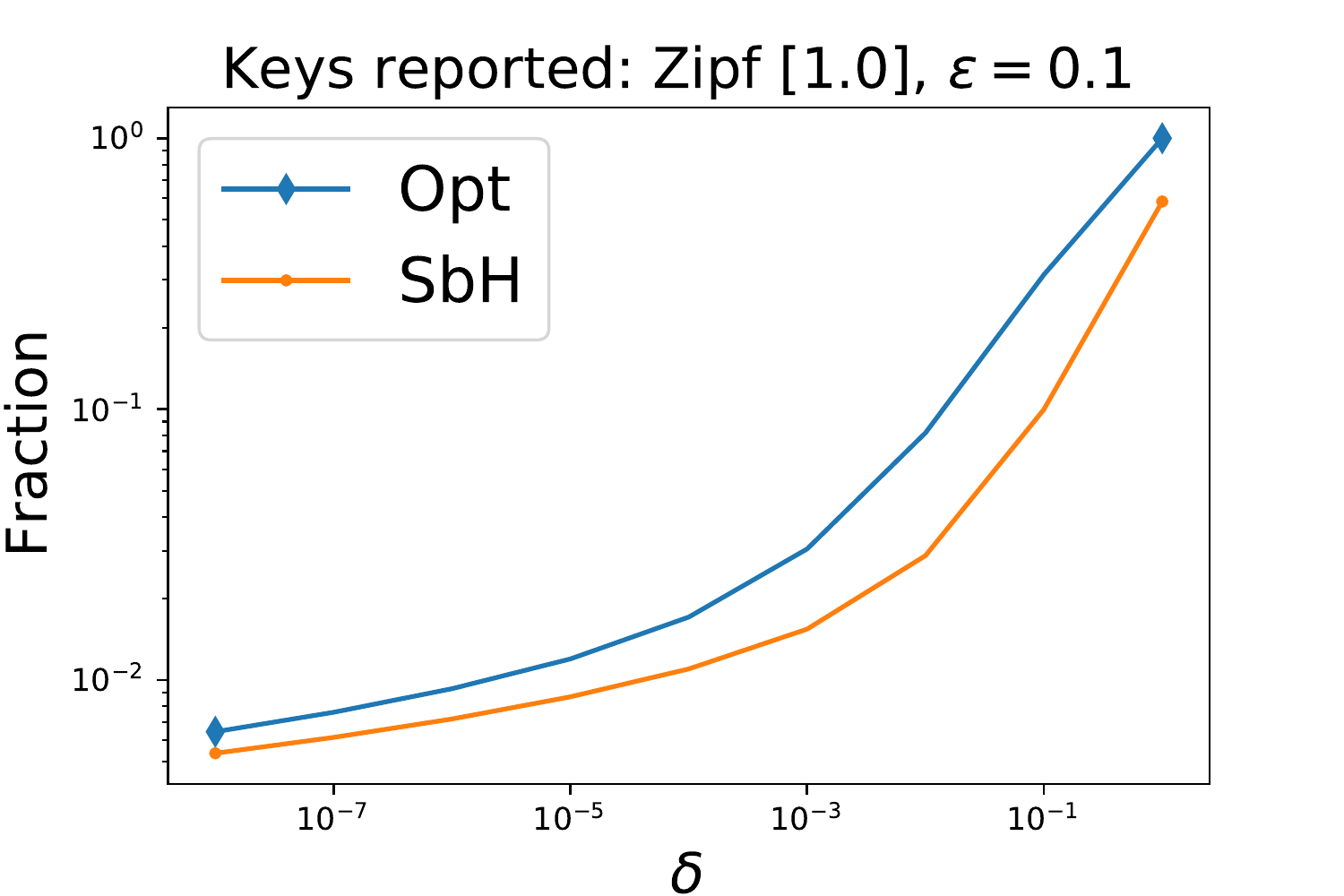}
\includegraphics[width=0.30\textwidth]{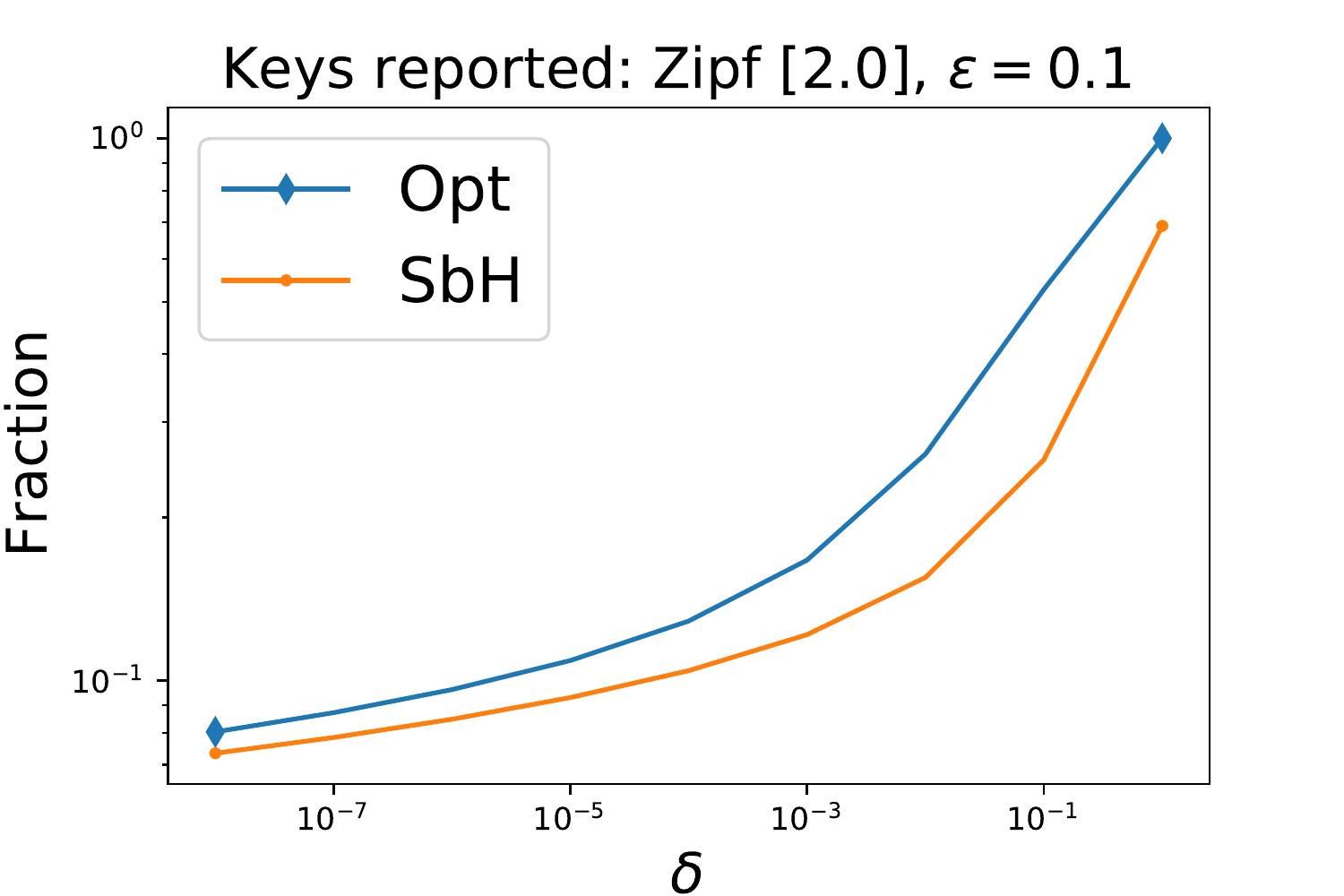}
\caption{Expected fraction of keys that are privately reported with PWS (Opt) and SbH for $\Zipf[\alpha]$ frequency distributions. For $\alpha=1,2$, \notinproc{privacy parameters }$\varepsilon=0.1$ and sweeping $\delta$ between $1$ and $10^{-8}$. Left: The respective ratio of PWS to SbH.}
\label{Zipfreport:plot}
\end{figure}

\notinproc{\subsection{Reporting Probabilities: No Sampling}}
\onlyinproc{\paragraph{Reporting Probabilities: No Sampling}}
We start with the case of no sampling and the objective of maximizing the number of privately reported keys.  We compare the PWS
(optimal) probabilities $\pi^*$ \eqref{pistar:eq} to the baseline SbH  \notinproc{\cite{BunNS19,KorolovaKMN09,VadhanDPsurvey:2017}}
reporting probabilities $\phi$\notinproc{ \eqref{SbHphi:eq}}.  
Figure~\ref{probfreqhist:plot} shows reporting probability per frequency for selected DP parameters.  We can see that with both private methods the reporting probability reaches $1$ for high frequencies but PWS (Opt) reaches the maximum earlier and is significantly higher than $\phi$ along the way. Analytically from the expressions we can see that
for $i\leq L(\varepsilon,\delta)$, 
$\pi^*_i/\phi_i \in [2,2/\varepsilon]$ with $\pi^*_i/\phi_i \approx 2i$ for lower $i$. We can also see that $\pi^*_i =1$ for  $i=2L+2 \approx \frac{2}{\varepsilon}\ln(\varepsilon/\delta)$ whereas $\phi_i > 1-\delta$ for $i\approx \frac{2}{\varepsilon} \ln(1/\delta)$.  The ratio between the frequency values when maximum reporting is reached is $\approx\ln(1/\delta)/\ln(\varepsilon/\delta)$.\notinproc{

} Figure~\ref{Zipfreport:plot} shows the expected numbers of reported keys with PWS (Opt) and SbH for frequency distributions that are $\Zipf[\alpha]$ with $\alpha=1,2$ as we sweep the privacy parameter $\delta$. Overall we see that PWS gains 20\%-300\% in the number of keys reported over baseline. 
Note that as expected, the optimal PWS reports {\em all} keys when $\delta=1$ (i.e., no privacy guarantees) but SbH incurs reporting loss. \onlyinproc{In the full version, we show similar results on two real-world datasets.}

\notinproc{
We additionally evaluate PWS and compare it to SbH on two real-world datasets:
\begin{enumerate}
    \item ABC: The words of news headlines from the Australian Broadcasting Corporation.  The keys are words and the frequency is the respective number of occurrences  \cite{abcnews}.
    \item SO: The multi-graph of Stack Overflow where keys are nodes in the graph and frequencies are undirected degrees \cite{StackOverflow}.
\end{enumerate}
Figure~\ref{RealWorldNoSampling:plot} shows the expected numbers of reported keys on these datasets. We note that the results are similar to what was observed on the synthetic Zipf datasets.
}

\notinproc{
\begin{figure}[ht]
\centering
\includegraphics[width=0.30\textwidth]{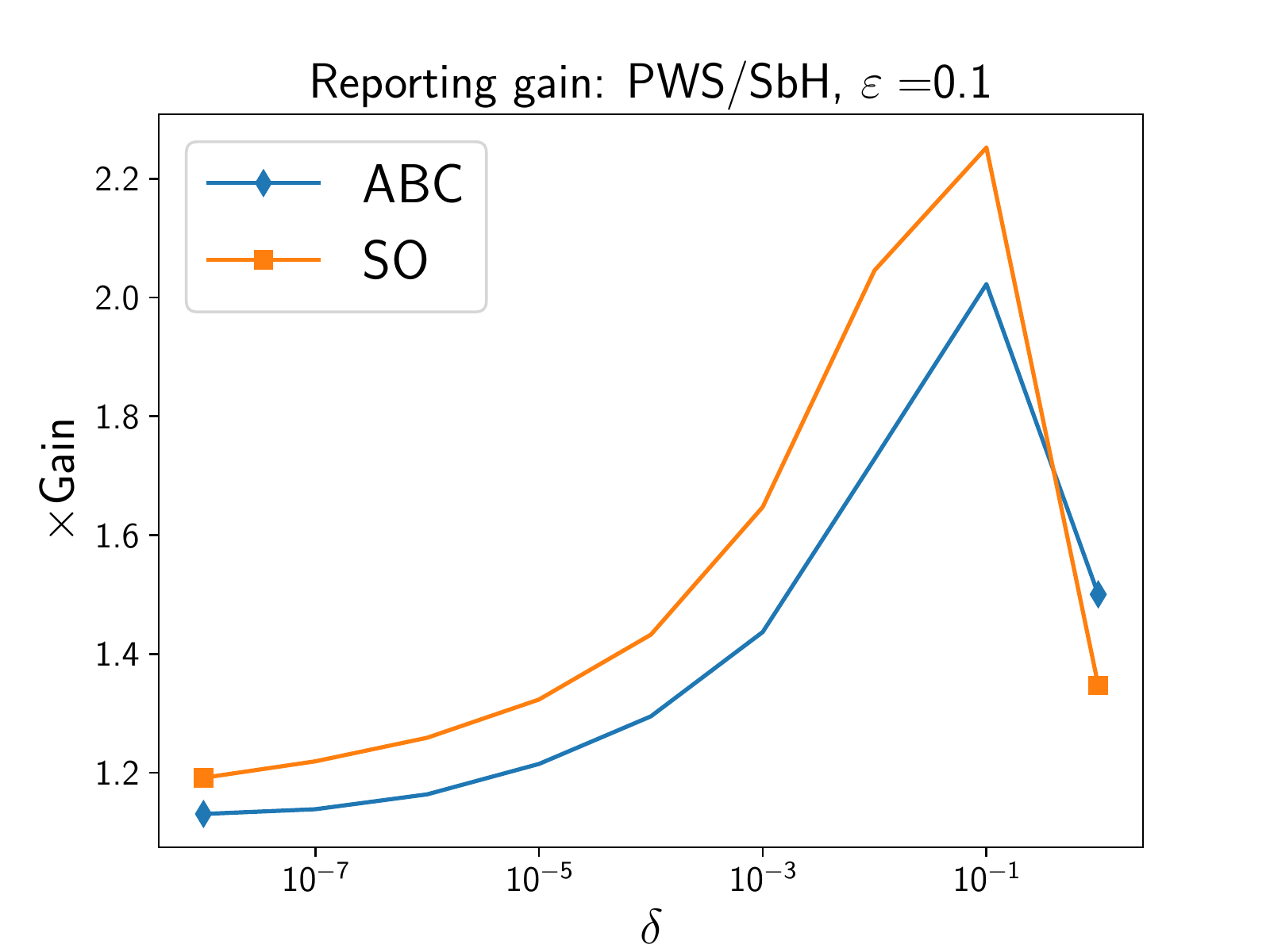}
\includegraphics[width=0.30\textwidth]{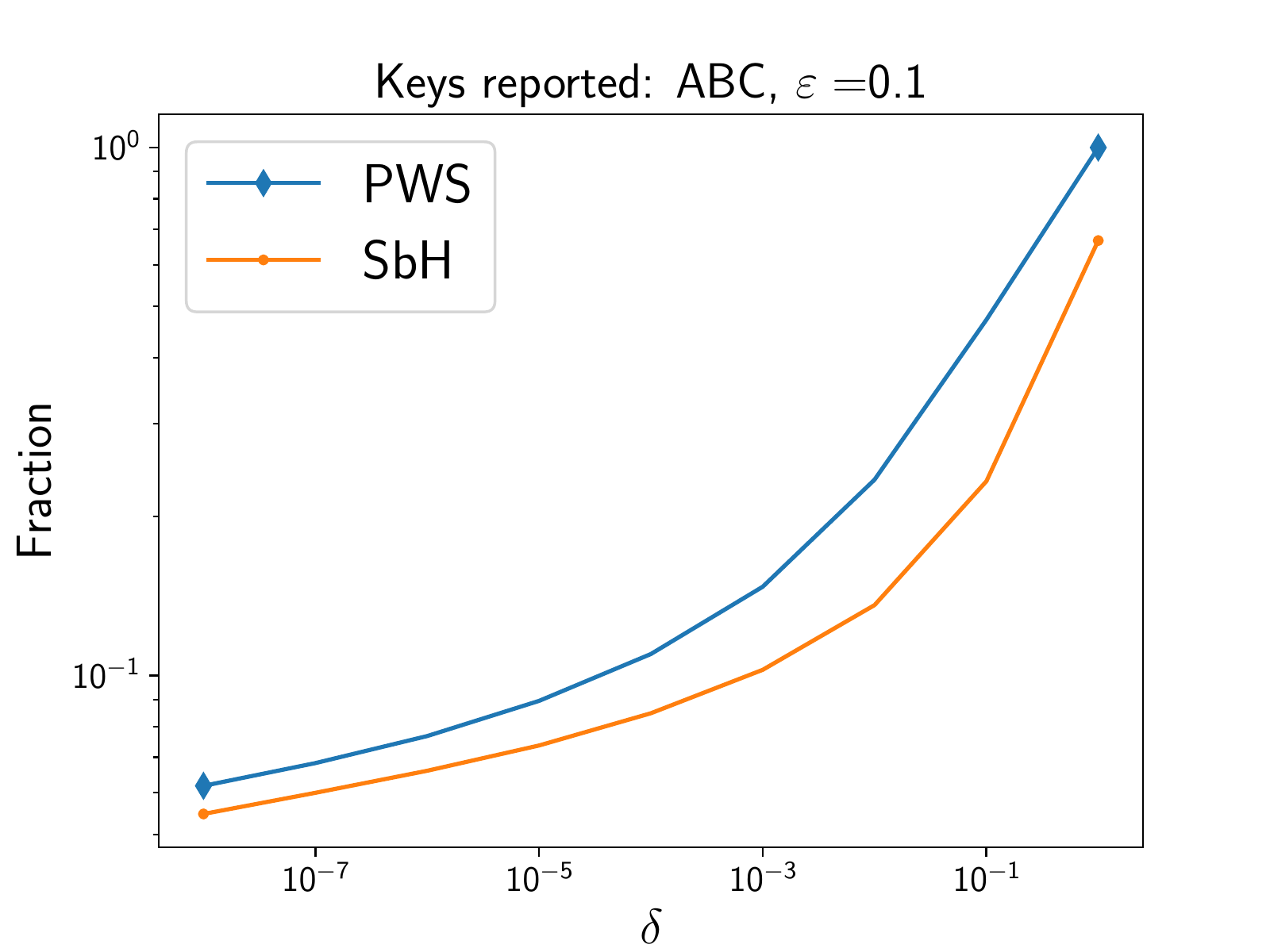}
\includegraphics[width=0.30\textwidth]{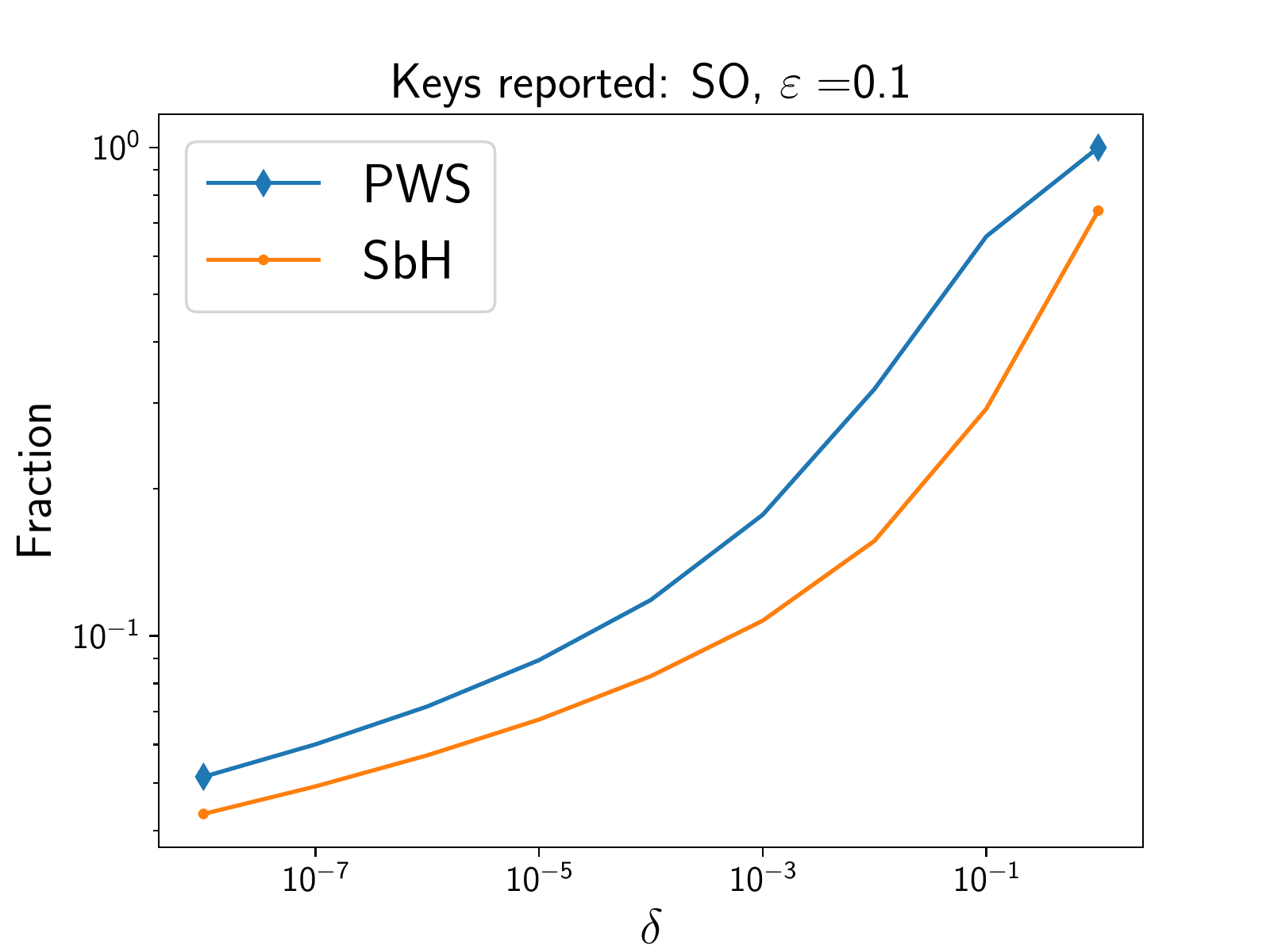}
\caption{Evaluation of PWS on real-world datasets (without sampling). Left: The ratio of reported keys with PWS to SbH. Center and Right: Fraction of total keys reported with sampled-SbH and PWS as we sweep the parameter $\delta$.}
\label{RealWorldNoSampling:plot}
\end{figure}
}
 
\begin{figure}[ht]
\centering
\includegraphics[width=0.30\textwidth]{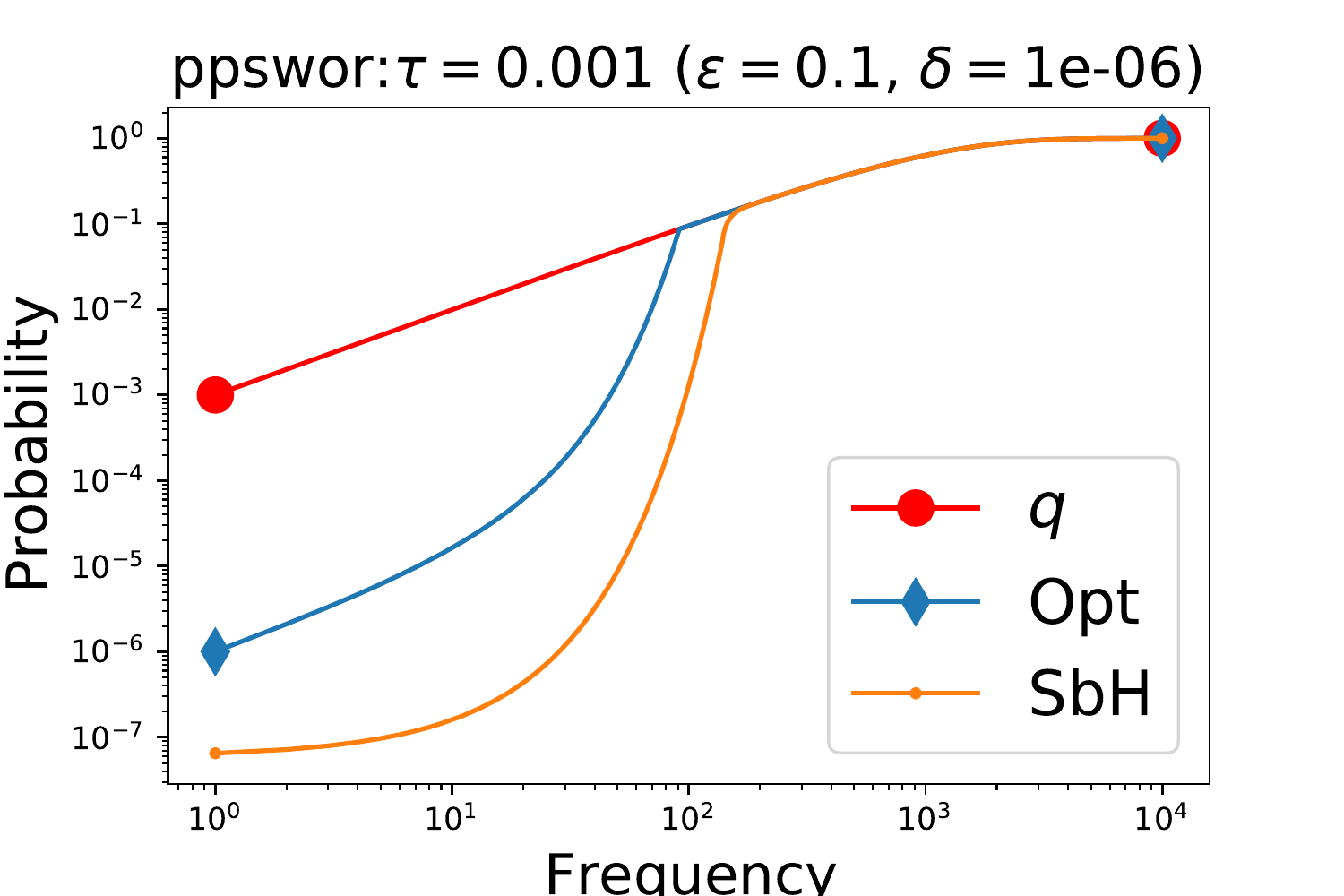}
\includegraphics[width=0.30\textwidth]{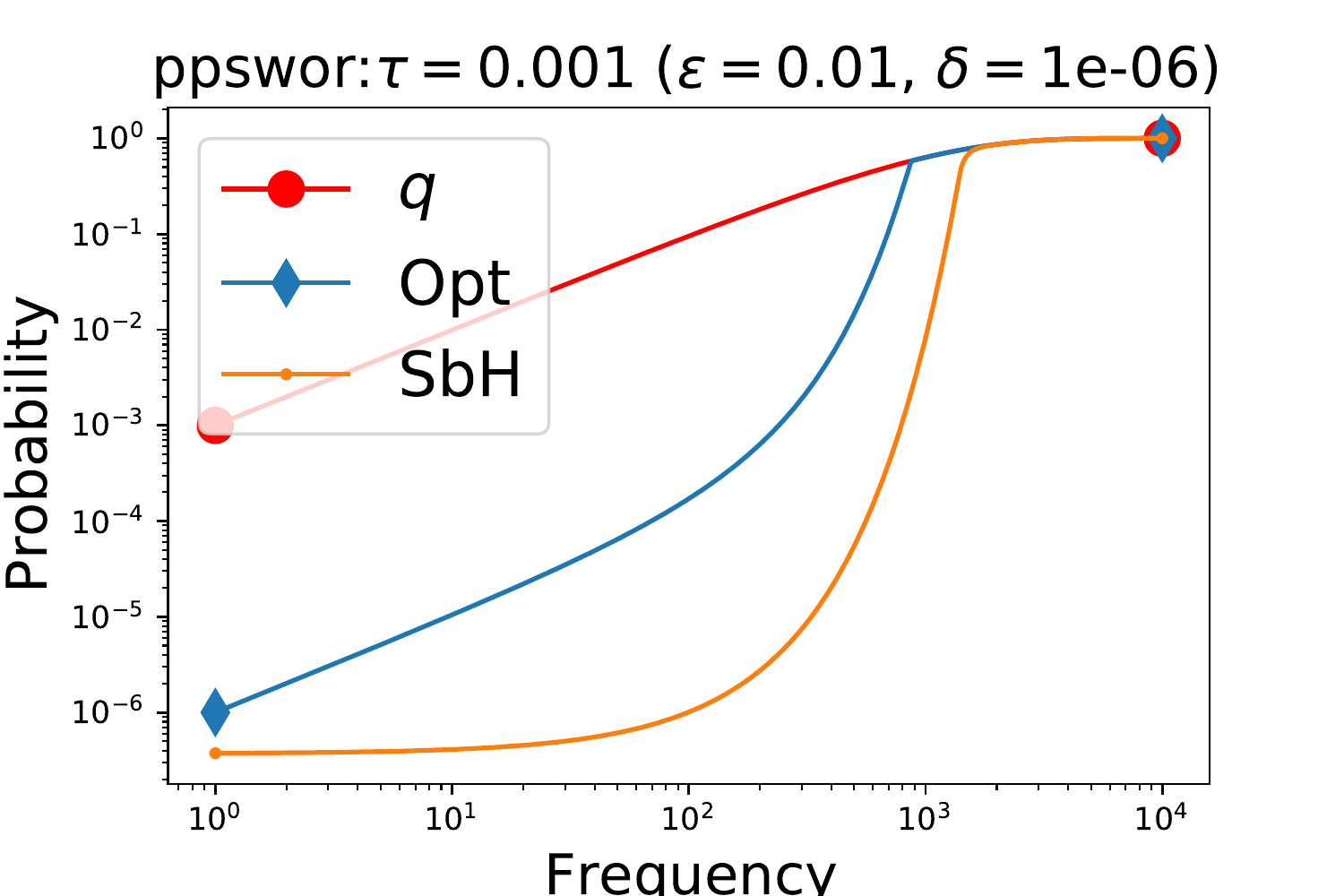}
\includegraphics[width=0.30\textwidth]{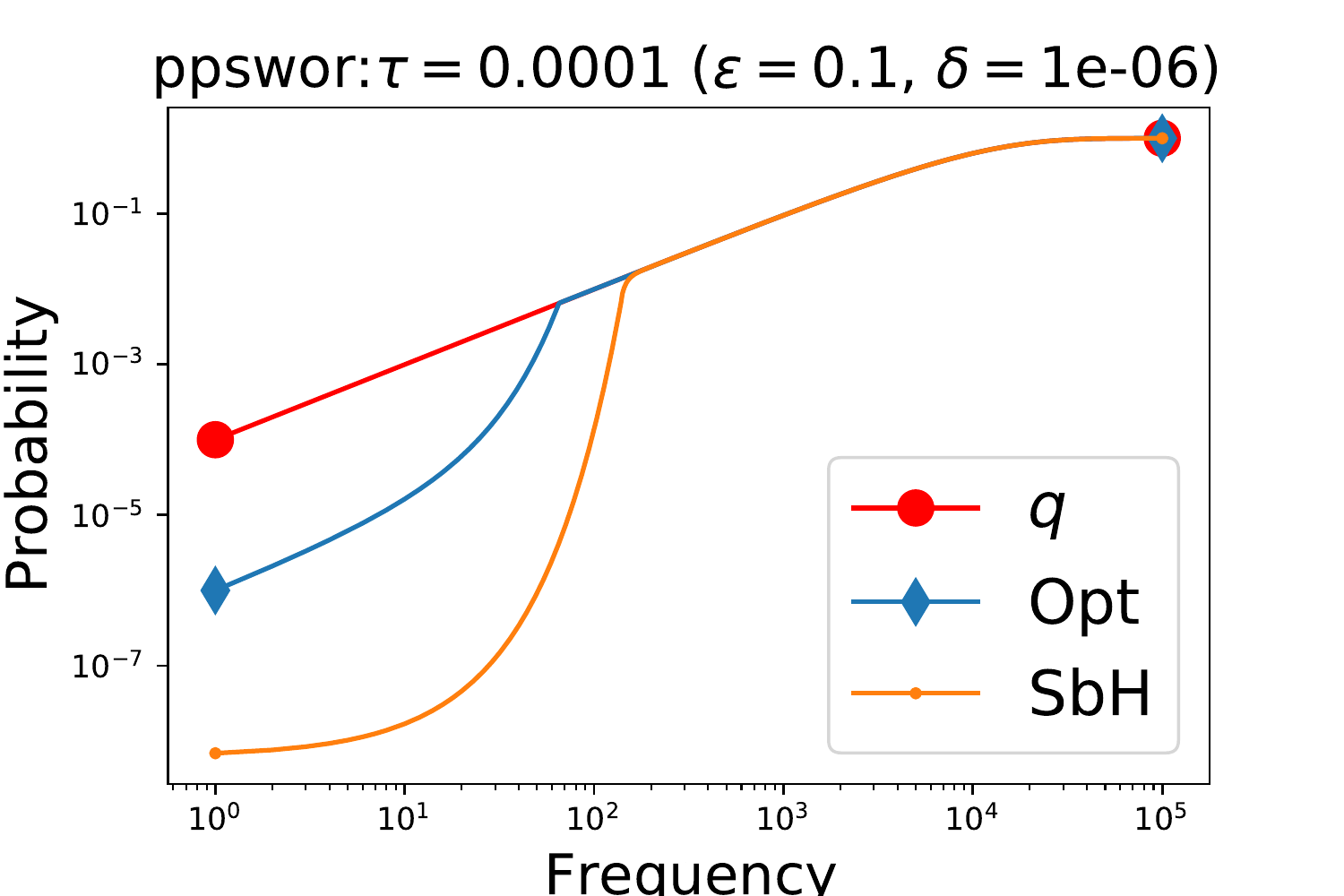}
\caption{Reporting probability as a function of frequency. For ppswor sampling with threshold $\tau$ ($q$), PWS private samples (Opt), and sampled-SbH private samples.}
\label{reporting_sampling:fig}
\end{figure}

\begin{figure}[ht]
\centering
\includegraphics[width=0.30\textwidth]{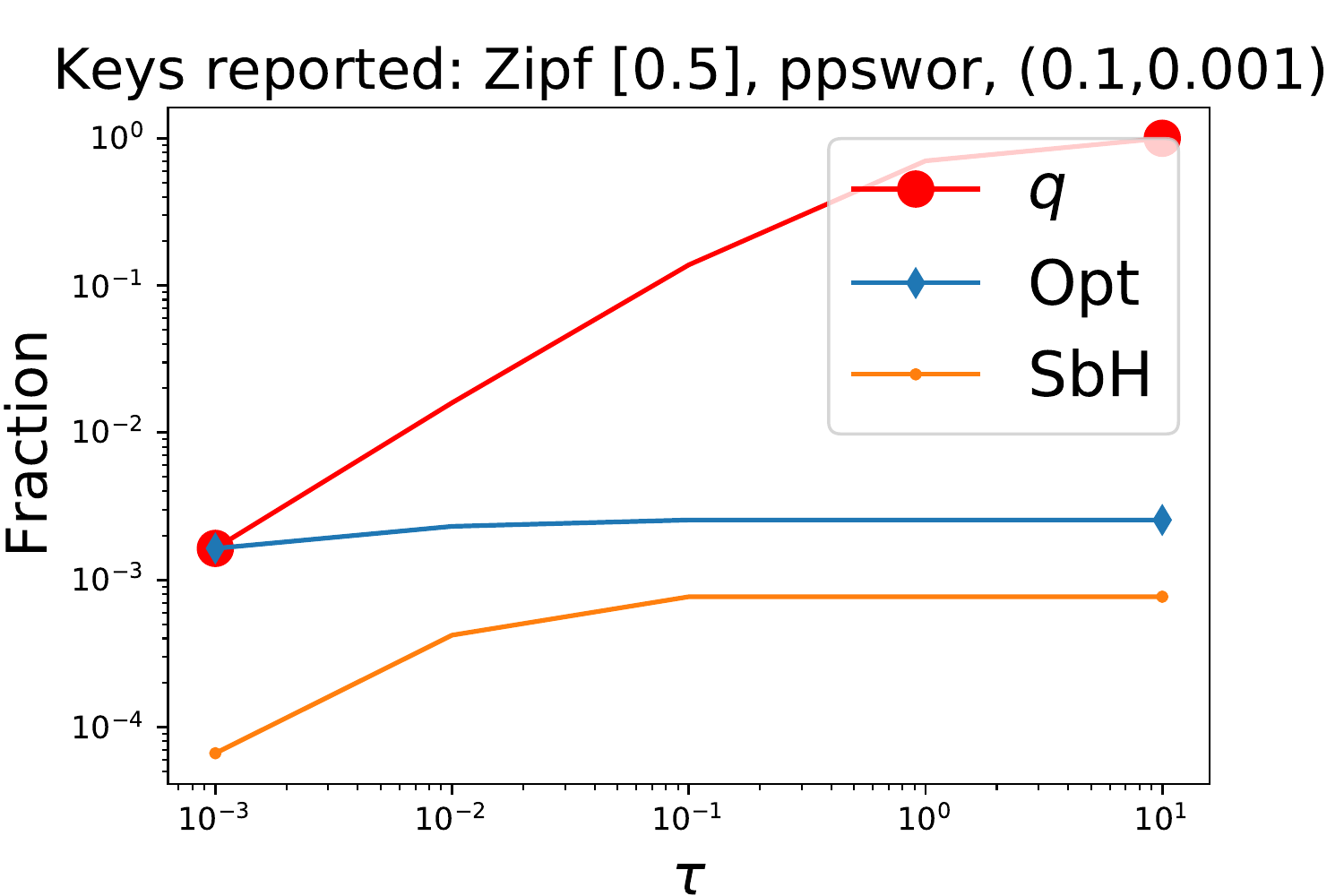}
\includegraphics[width=0.30\textwidth]{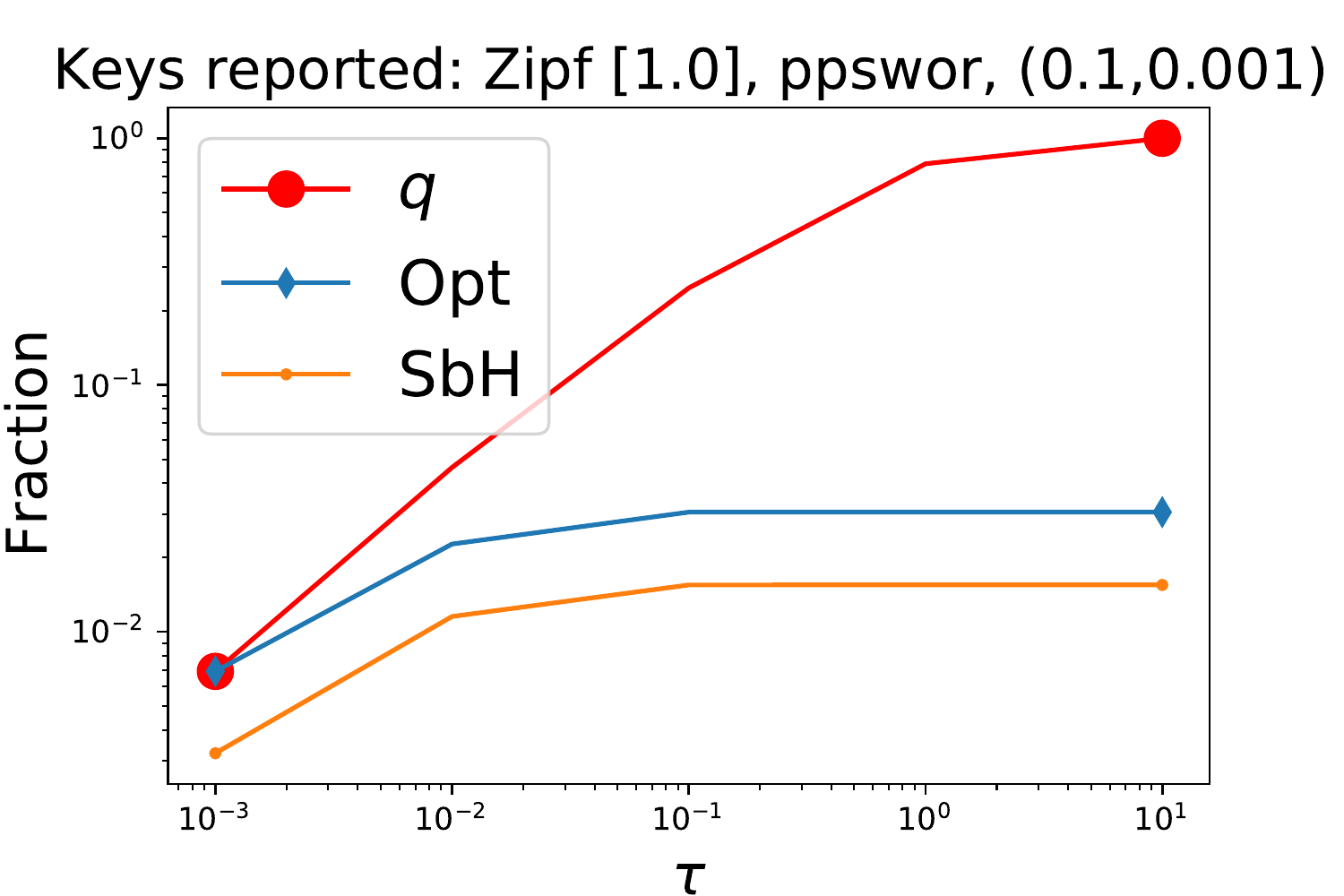}
\includegraphics[width=0.30\textwidth]{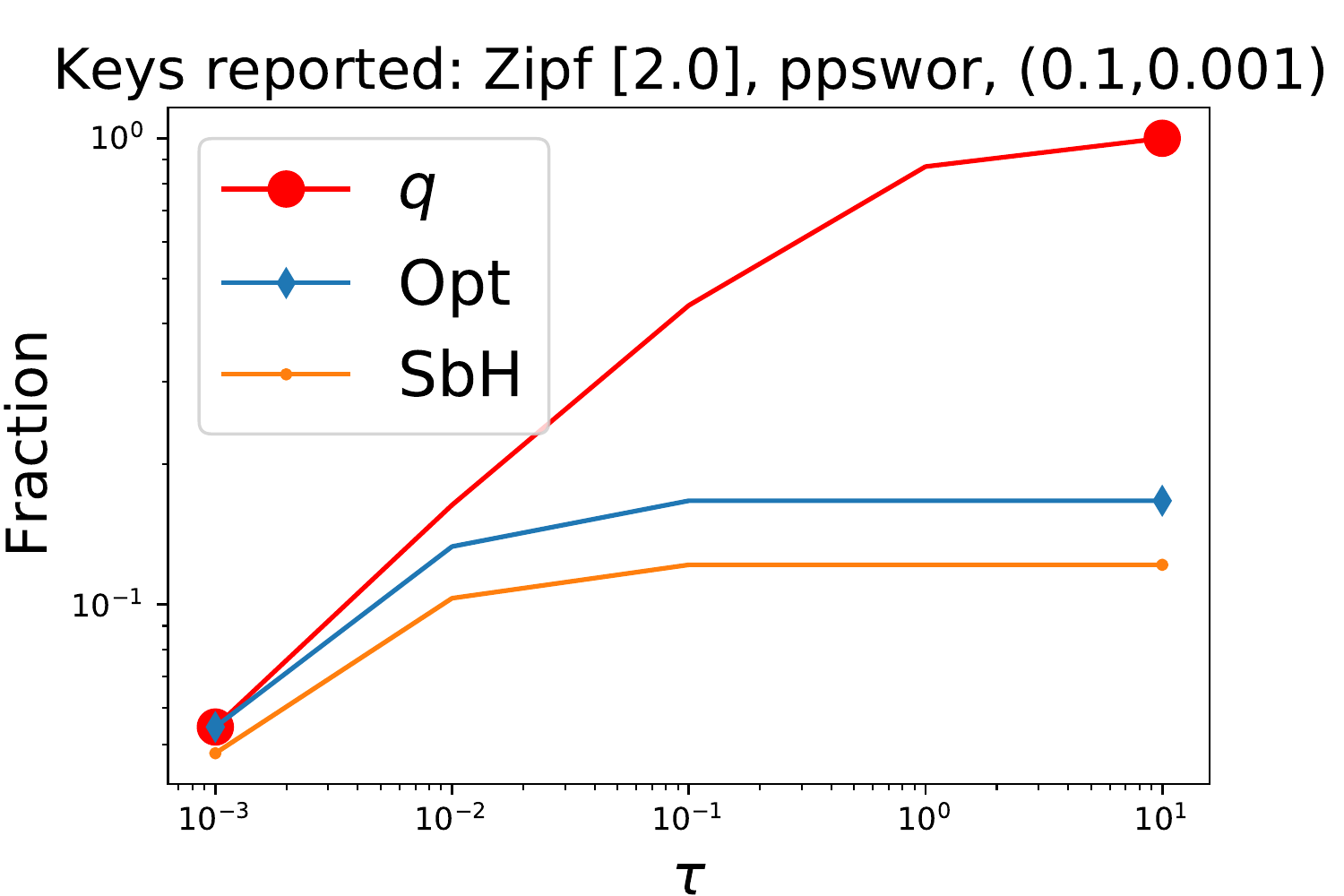}
\caption{Fraction of total keys reported with threshold-ppswor, sampled-SbH, and PWS (Opt), as we sweep the sampling rate $\tau$.  For $\Zipf[\alpha]$, $\varepsilon=0.1$ and $\delta=0.001$ the gains in reporting of PWS over Sampled-SbH are at least 230\% ($\alpha=0.5$), 97\% ($\alpha=1$) and 37\% ($\alpha=2$).}
\label{sweepsamplingZipf:fig}
\end{figure}

 \notinproc{\subsection{Reporting Probabilities with Sampling}}
 \onlyinproc{\paragraph{Reporting Probabilities with Sampling}}
 Figure~\ref{reporting_sampling:fig} shows
 reporting probabilities with PWS (optimal reporting probabilities), sampled-SbH, and non-private sampling, for representative sampling rates and privacy parameters.  
As expected, for sufficiently large frequencies both private methods have reporting probabilities that match the sampling probabilities $q$ of the non-private scheme.  
But PWS reaches $q$ at a lower frequency than sampled-SbH and has significantly higher reporting probabilities for lower frequencies. 
Figure~\ref{sweepsamplingZipf:fig} shows the fraction of keys reported for $\Zipf$ distributions as we sweep the sampling rate (threshold $\tau$).   PWS reports more keys than sampled-SbH and the gain persists also with low sampling rates.  We can see that with PWS, thanks to end-to-end privacy analysis, the reporting loss due to sampling mitigates the reporting loss needed for privacy -- reporting approaches that of the non-private sampling when the sampling rate $\tau$ approaches $\delta$.  Sampled-SbH, on the other hand, incurs reporting loss due to privacy on top of the reporting loss due to sampling.
\notinproc{Figures \ref{RealWorldWithSampling1:plot} and \ref{RealWorldWithSampling2:plot} show the expected fraction of reported keys on the real-world datasets ABC and SO.}

\notinproc{
\begin{figure}[ht]
\centering
\includegraphics[width=0.44\textwidth]{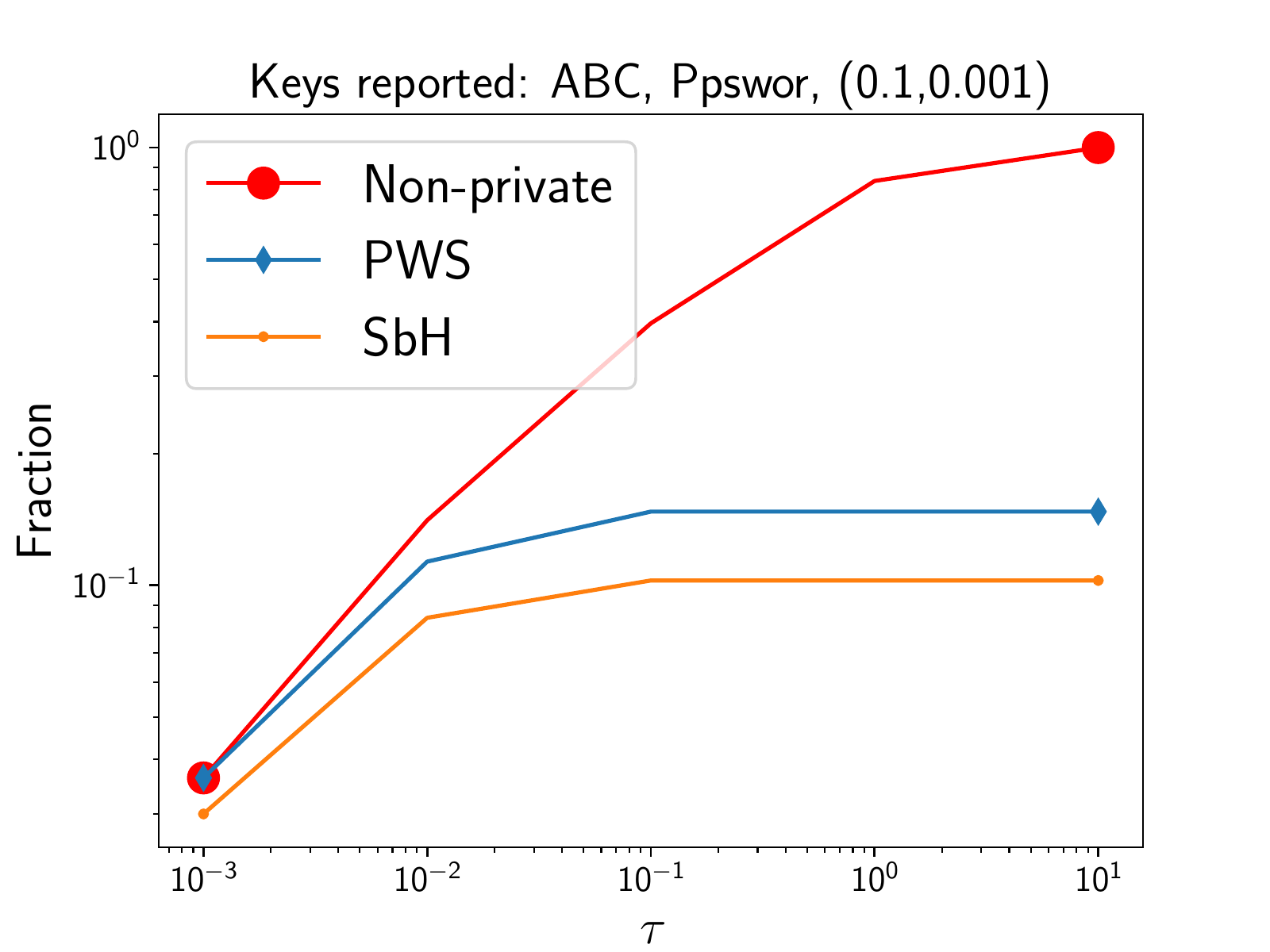}
\includegraphics[width=0.44\textwidth]{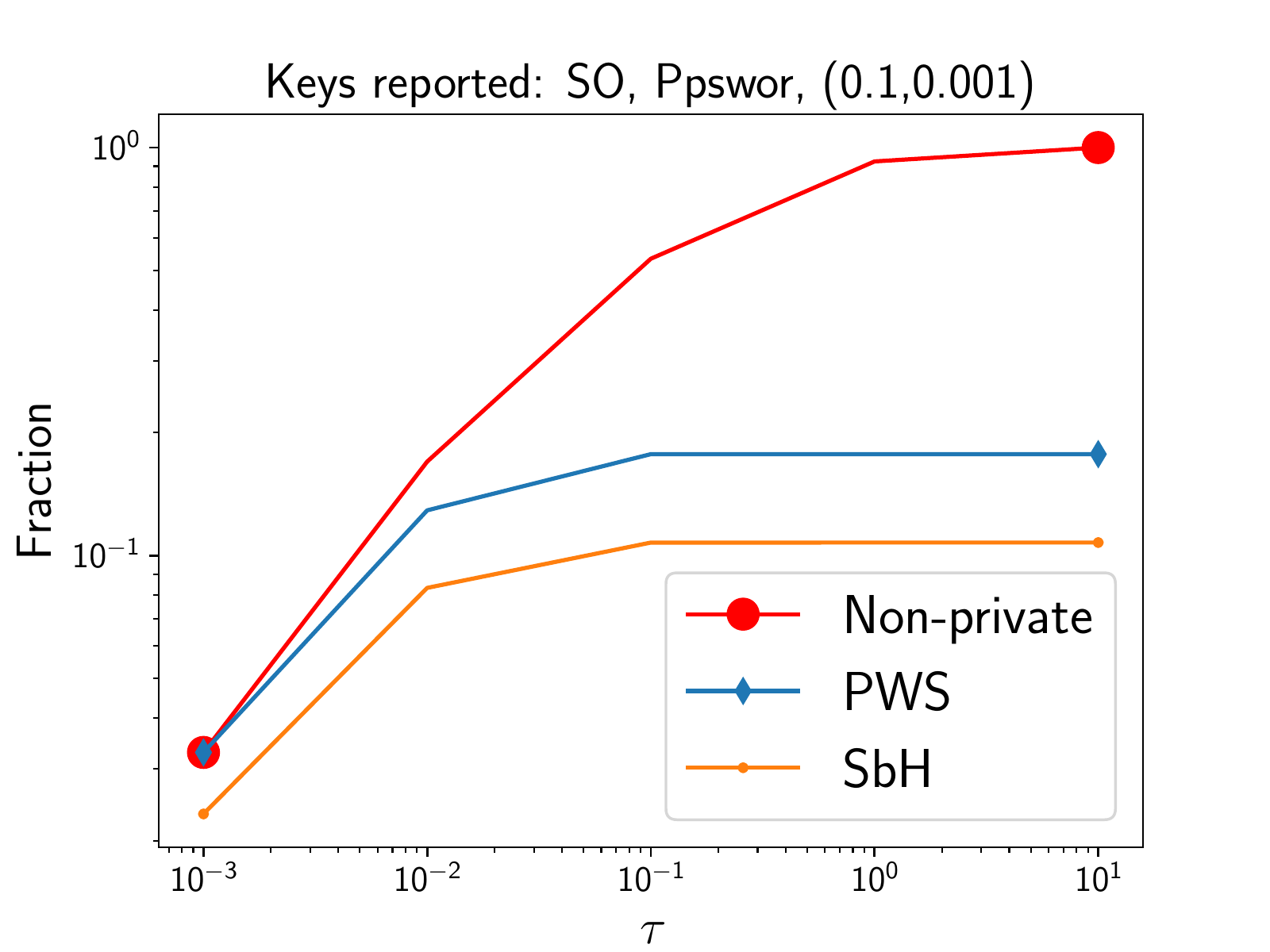}
\caption{Evaluation on real-world datasets (with sampling): fraction of total keys reported with threshold-ppswor (non-private), sampled-SbH, and PWS, as we sweep the sampling rate $\tau$.}
\label{RealWorldWithSampling1:plot}
\end{figure}
\begin{figure}[ht]
\centering
\includegraphics[width=0.44\textwidth]{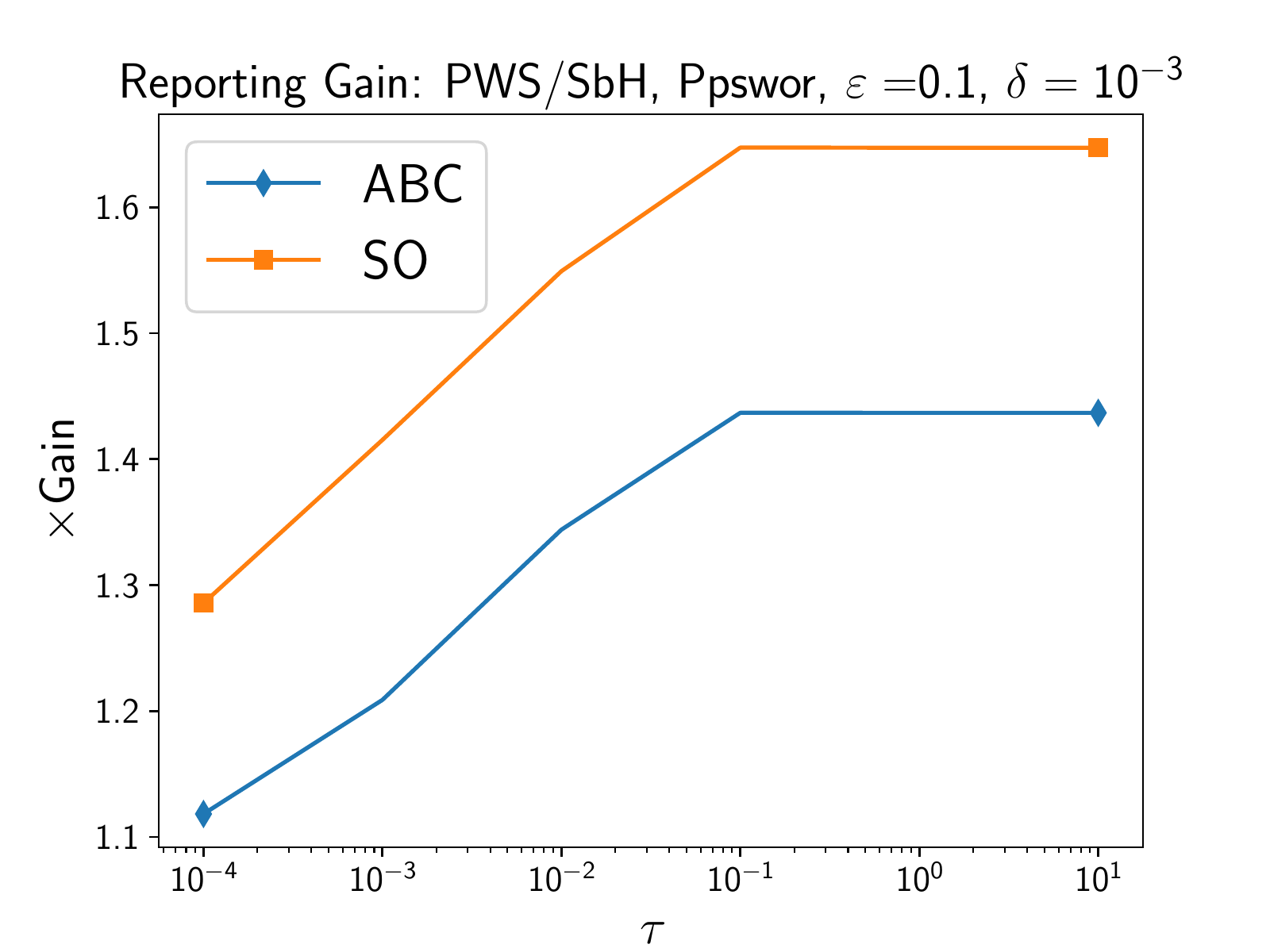}
\includegraphics[width=0.44\textwidth]{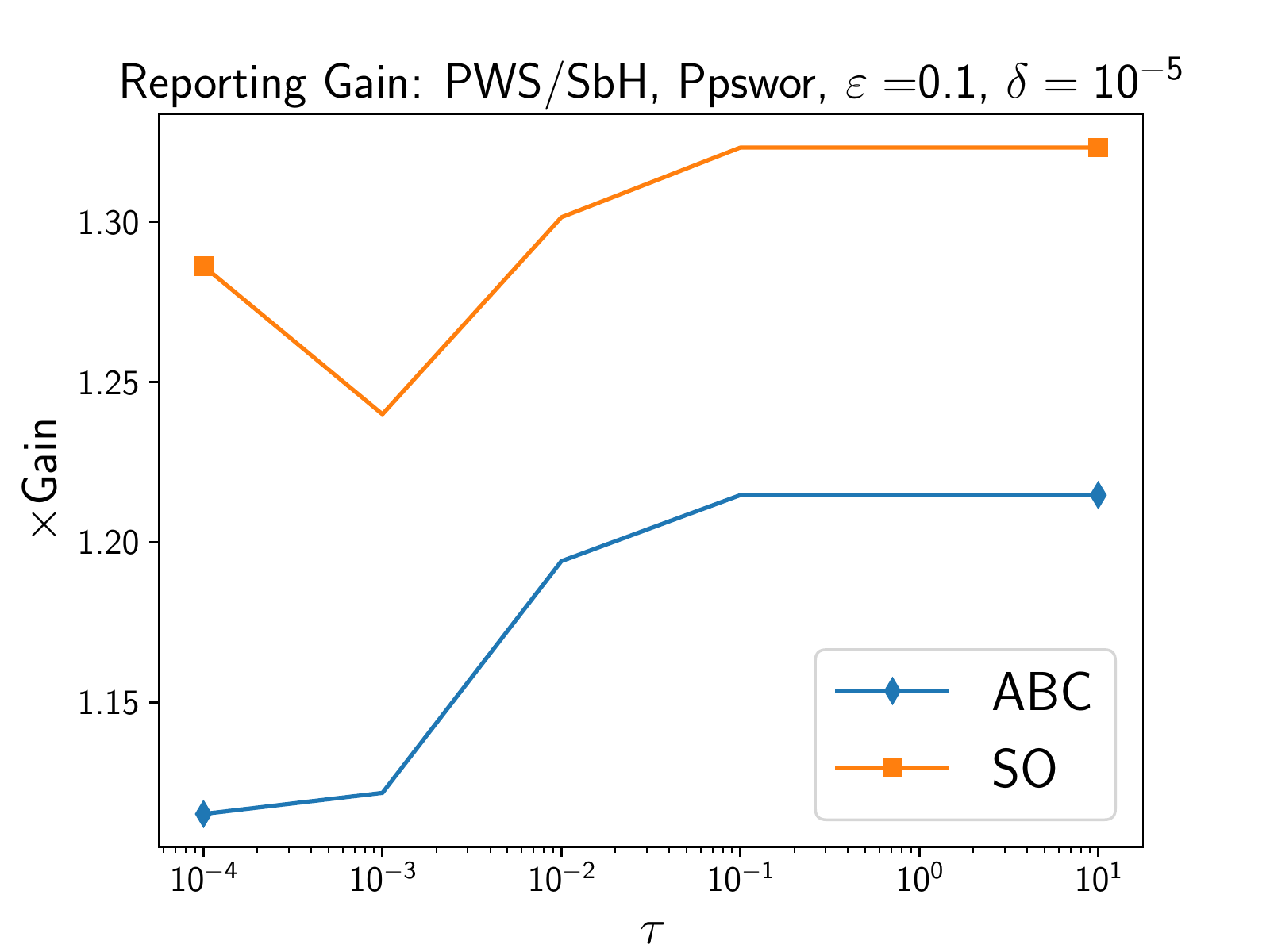}
\caption{Ratio of expected number of keys that are privately reported with PWS to SbH for the ABC and SO datasets, as we sweep the sampling rate $\tau$.}
\label{RealWorldWithSampling2:plot}
\end{figure}
}
 
\begin{figure}[ht]
\centering
\includegraphics[width=0.44\textwidth]{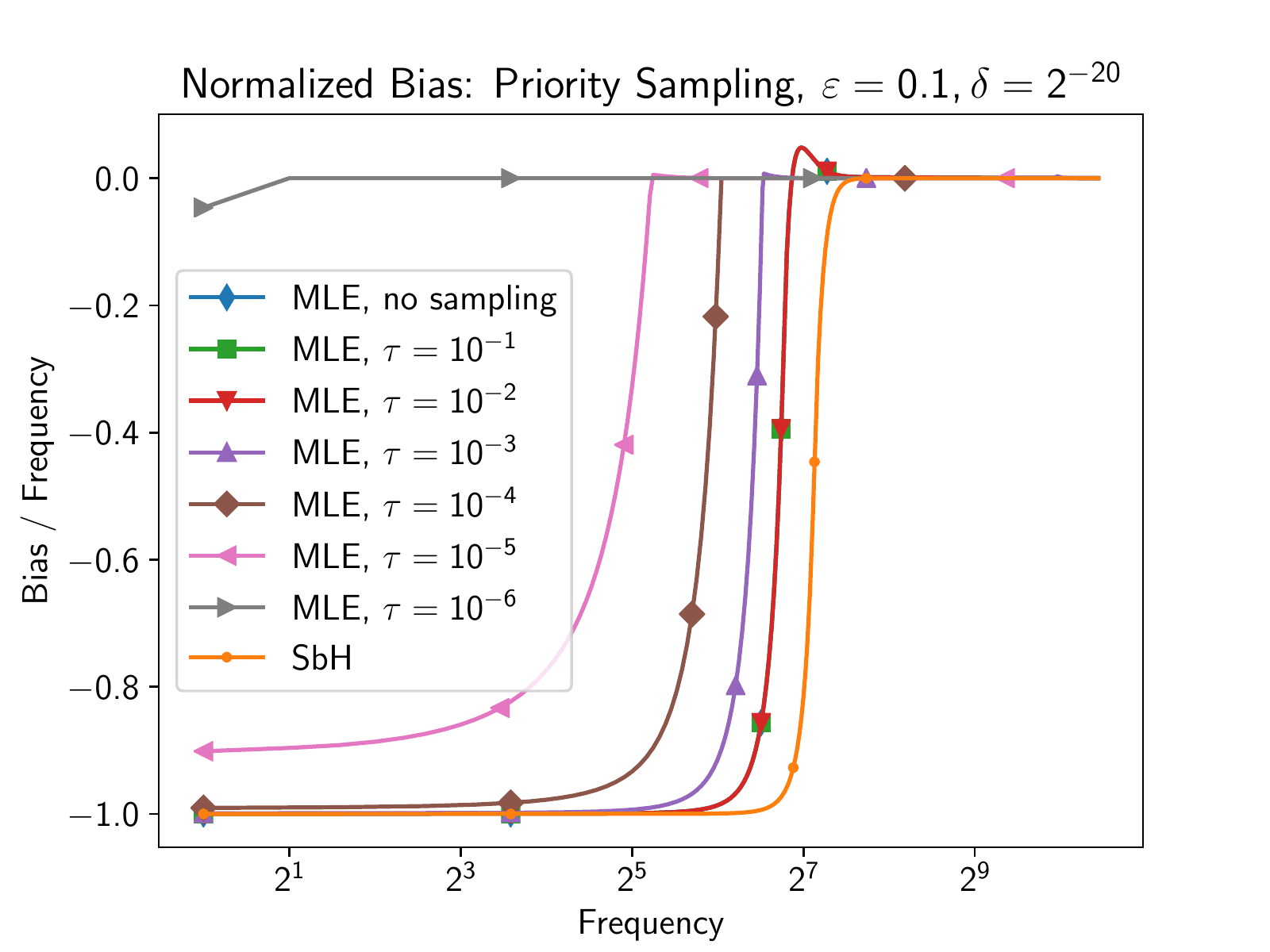}
\includegraphics[width=0.44\textwidth]{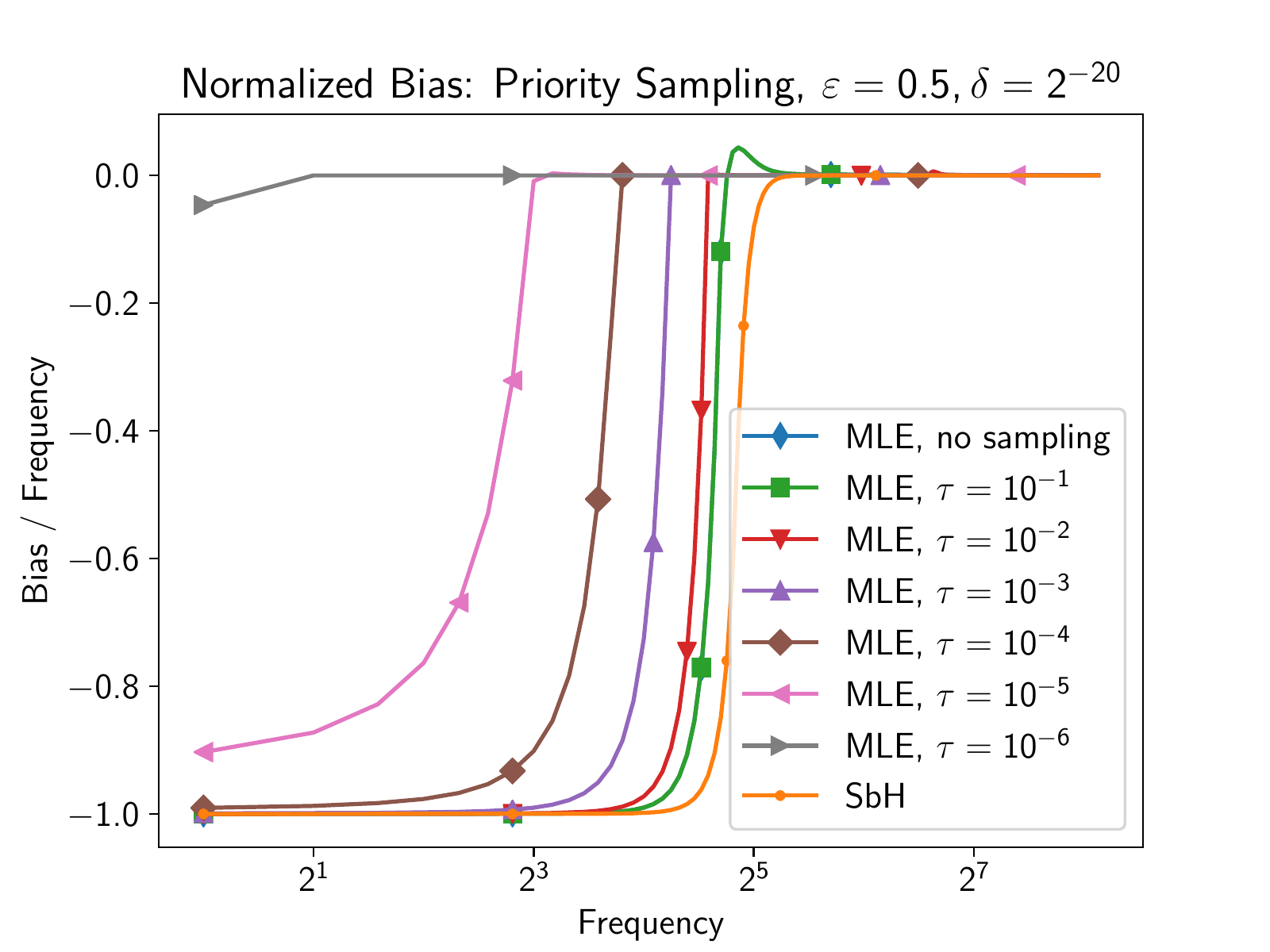}
\includegraphics[width=0.44\textwidth]{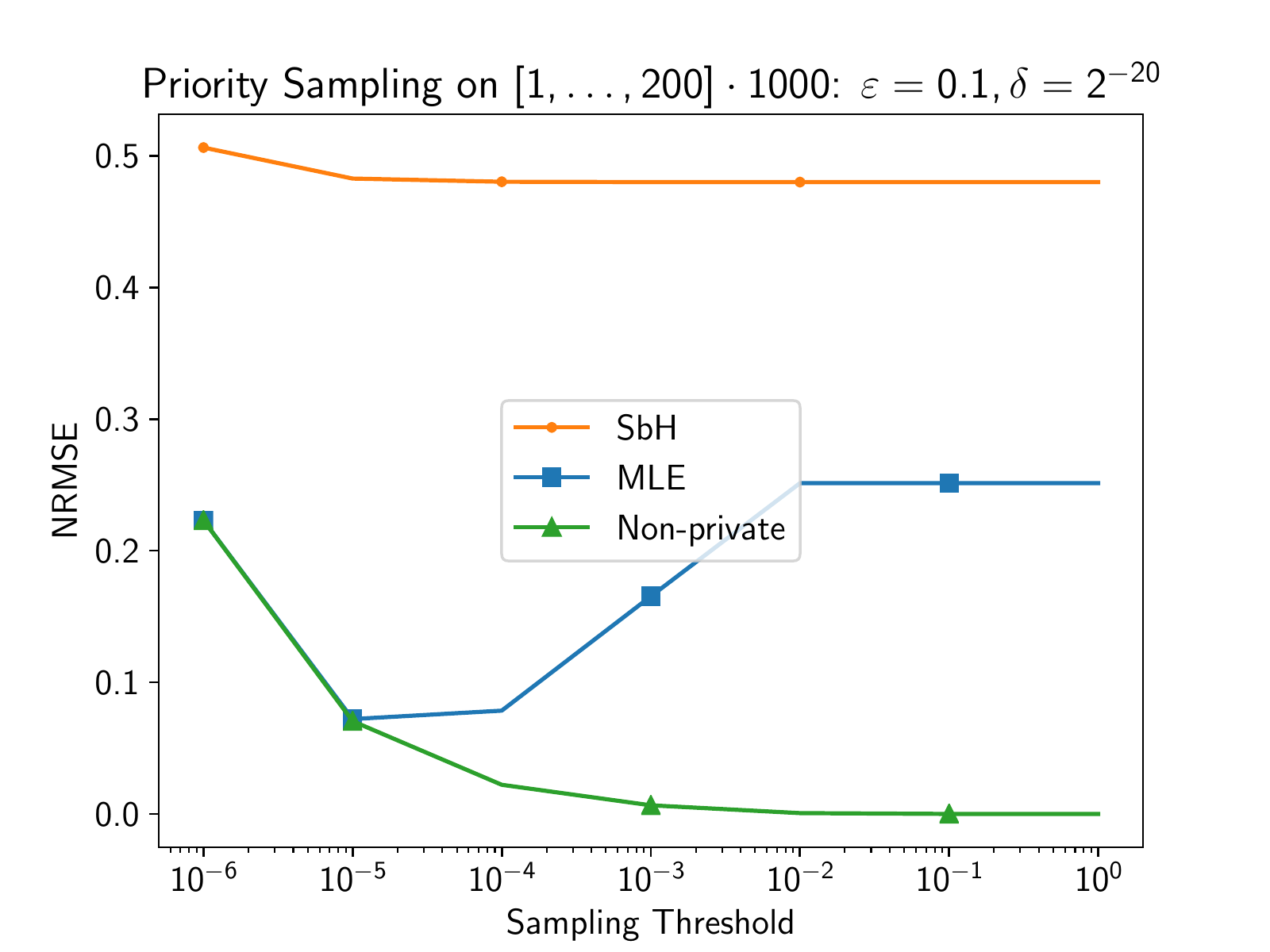}
\includegraphics[width=0.44\textwidth]{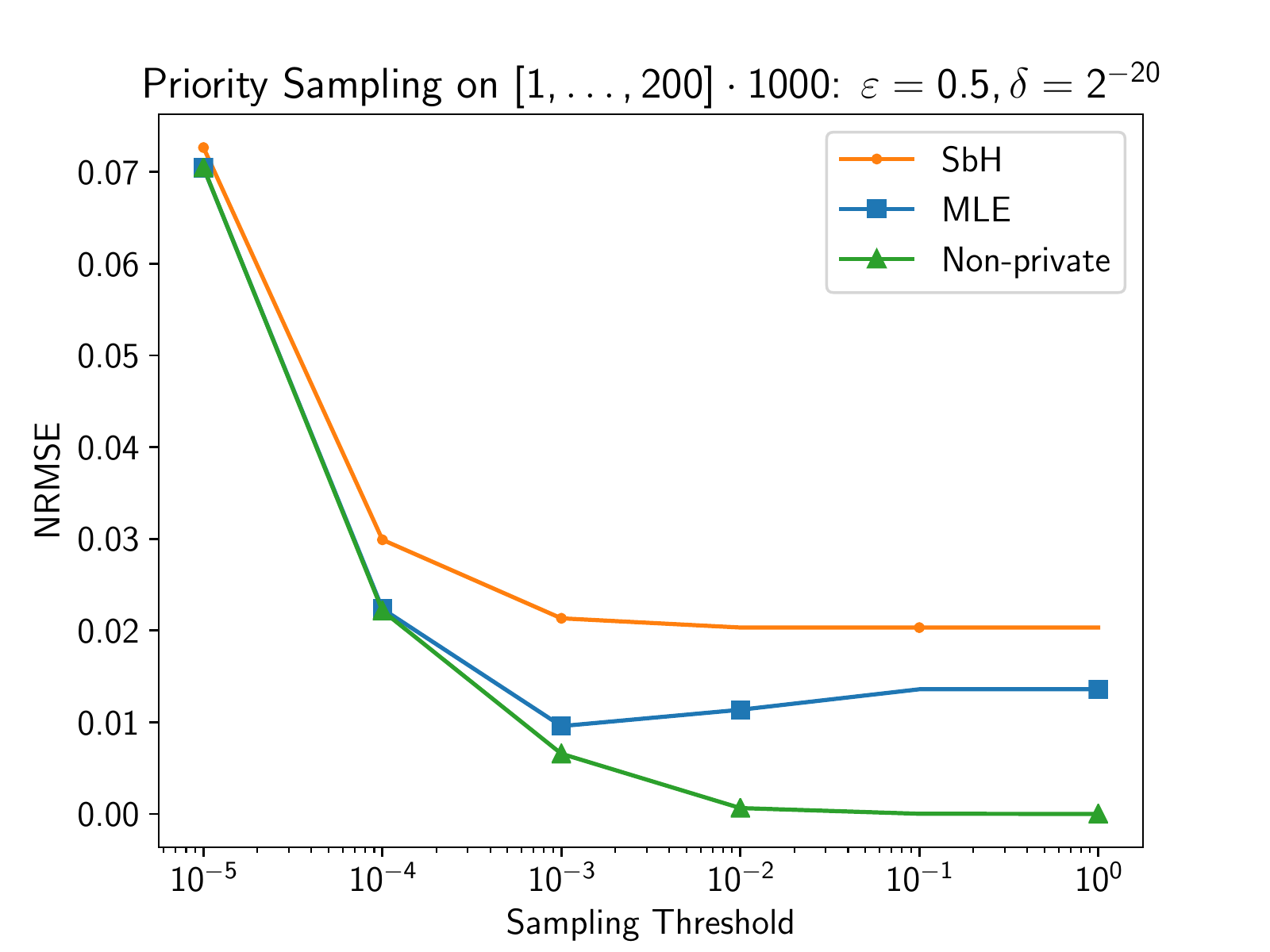}
\caption{ Top: Normalized bias for PWS (MLE) and sampled-SbH as a function of frequency, for different sampling rates.  The bias of the sampled-SbH estimates (shown once) does not change with sampling rate. Bottom: NRMSE as a function of sampling rate for a selection of $2\times 10^5$ keys with frequencies drawn uniformly $[1,200]$.
}
\label{bias_error:fig}
\end{figure}

\notinproc{\subsection{Estimation of Linear Statistics}}
\onlyinproc{\paragraph{Estimation of Linear Statistics}}
We evaluate estimation quality for linear statistics \eqref{qstat:eq} when $g(w)=w$ and $L(x)$ is a selection predicate.  The statistics is simply the sum of frequencies of selected keys.
We compare performance of PWS with the MLE estimator \eqref{MLest:eq}, the baseline sampled-SbH, and for reference, the estimator of the respective non-private sample \eqref{sumest:eq}.
Figure~\ref{bias_error:fig} (top) shows normalized bias $\Bias_i/i$ as a function of the frequency $i$ for the two private methods (the non-private estimator is unbiased and not shown).
With both methods, the bias decreases with frequency and diminishes for $i \gg 2\varepsilon^{-1}\ln(1/\delta)$.
PWS has lower bias at lower frequencies than SbH, allowing for more accurate estimates on a broader range. We can see that with PWS, the bias decreases 
when the sampling rate ($\tau$) decreases 
and diminishes when $\tau$ approaches $\delta$. This is a benefit of the end-to-end privacy analysis. The bias of the baseline method does not change with sampling rate. 

\begin{figure}[!pht]
\centering
\includegraphics[width=0.44\textwidth]{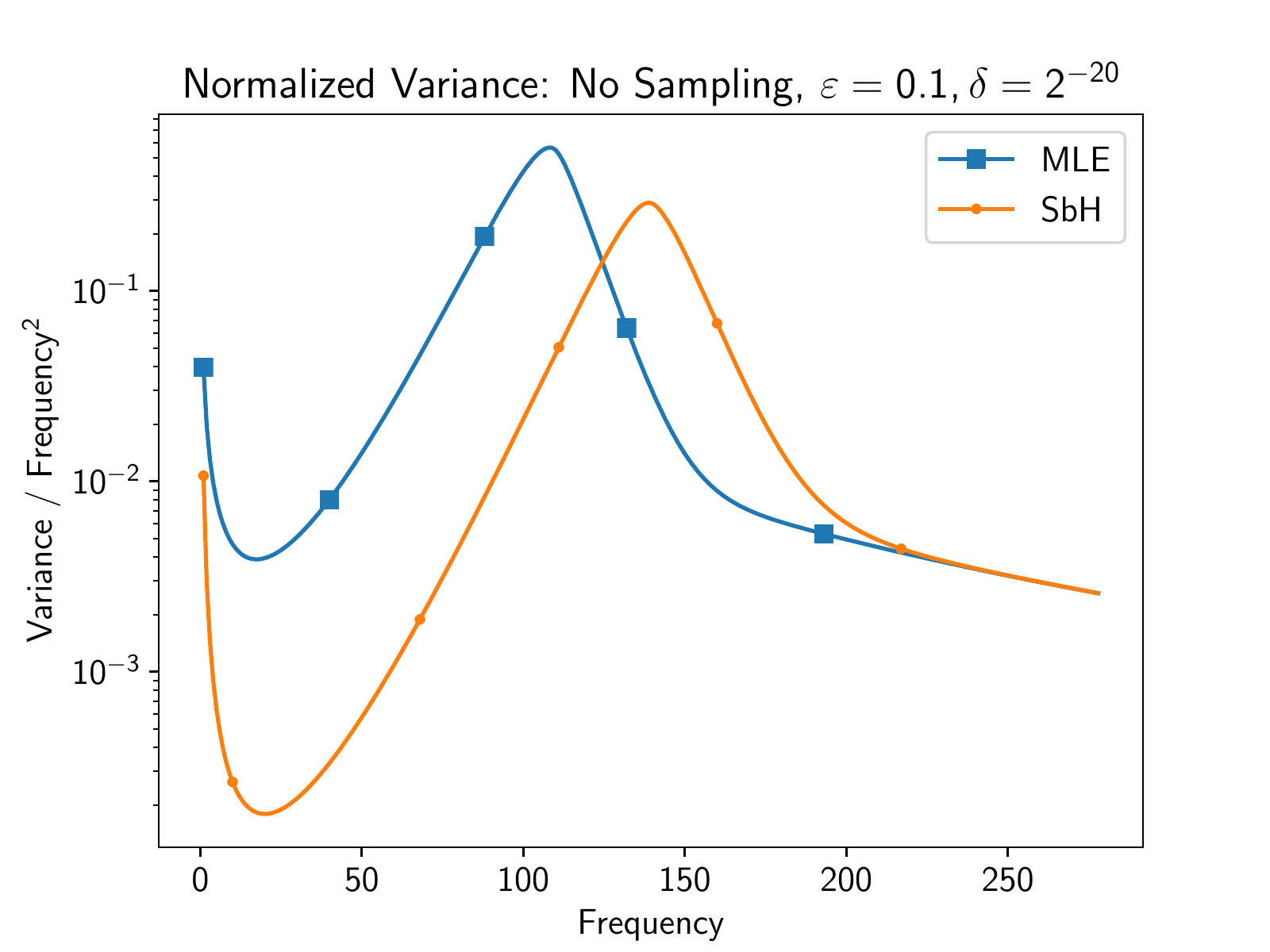}
\notinproc{
\includegraphics[width=0.44\textwidth]{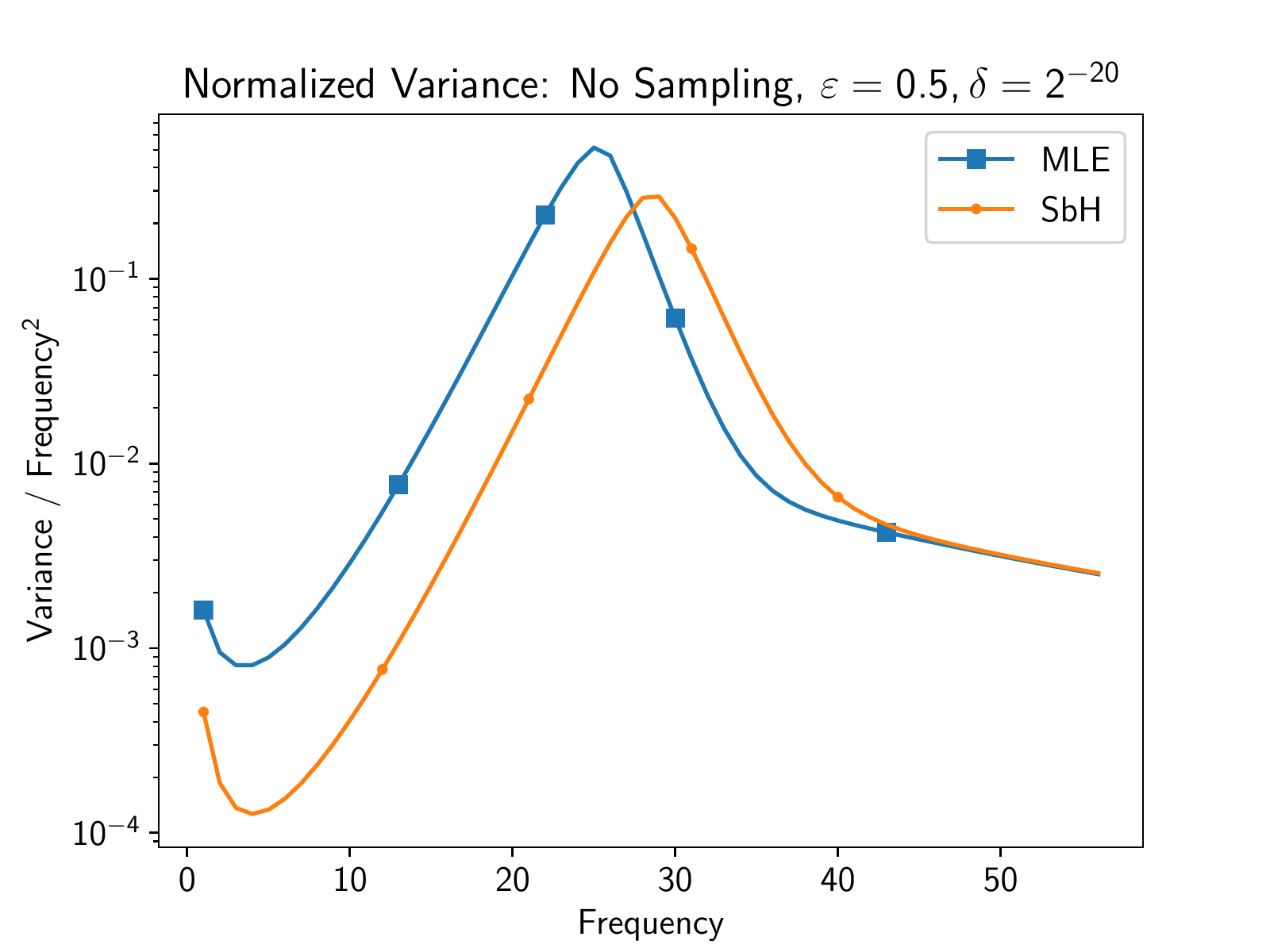}}
\includegraphics[width=0.44\textwidth]{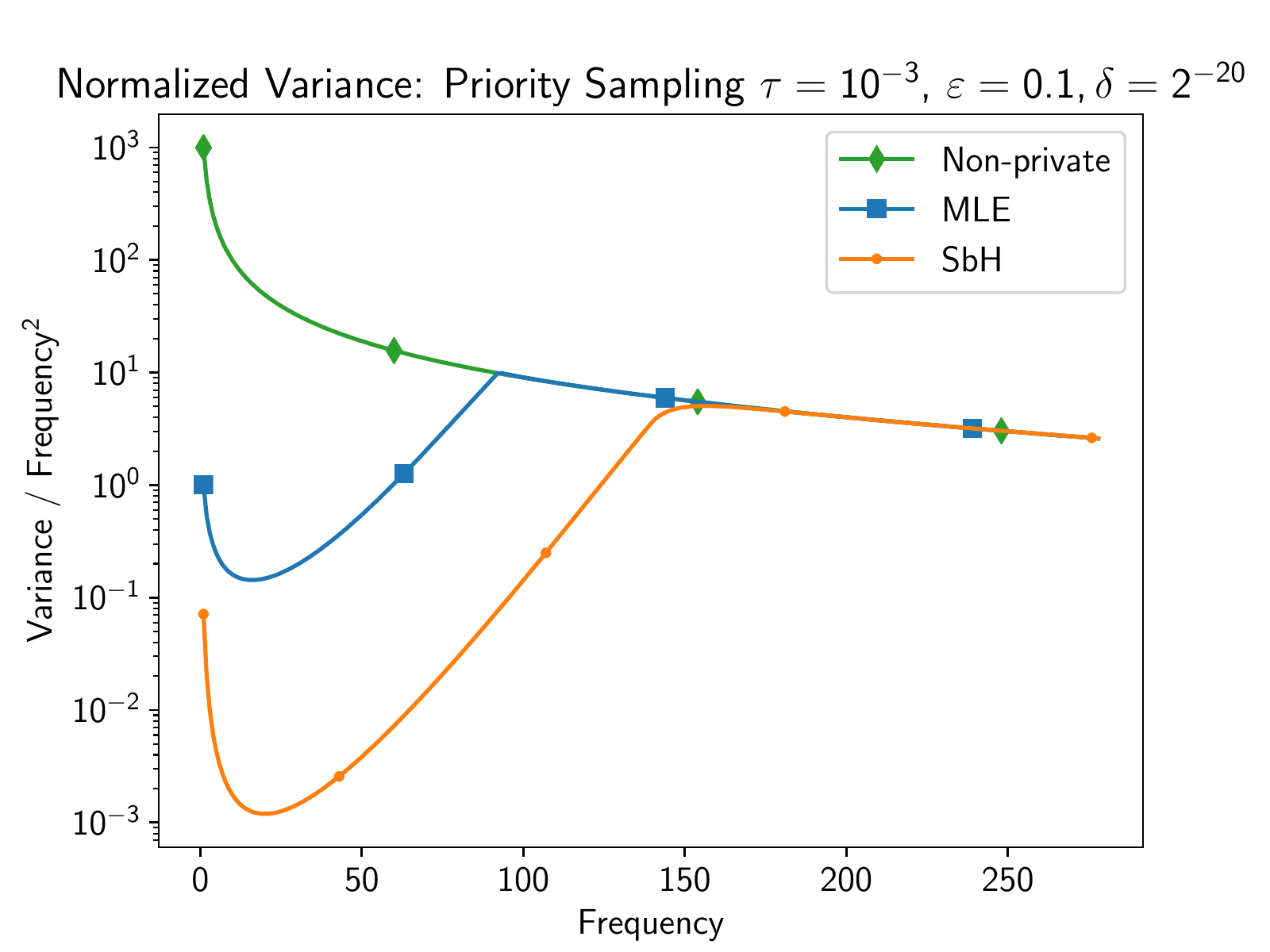}
\notinproc{
\includegraphics[width=0.44\textwidth]{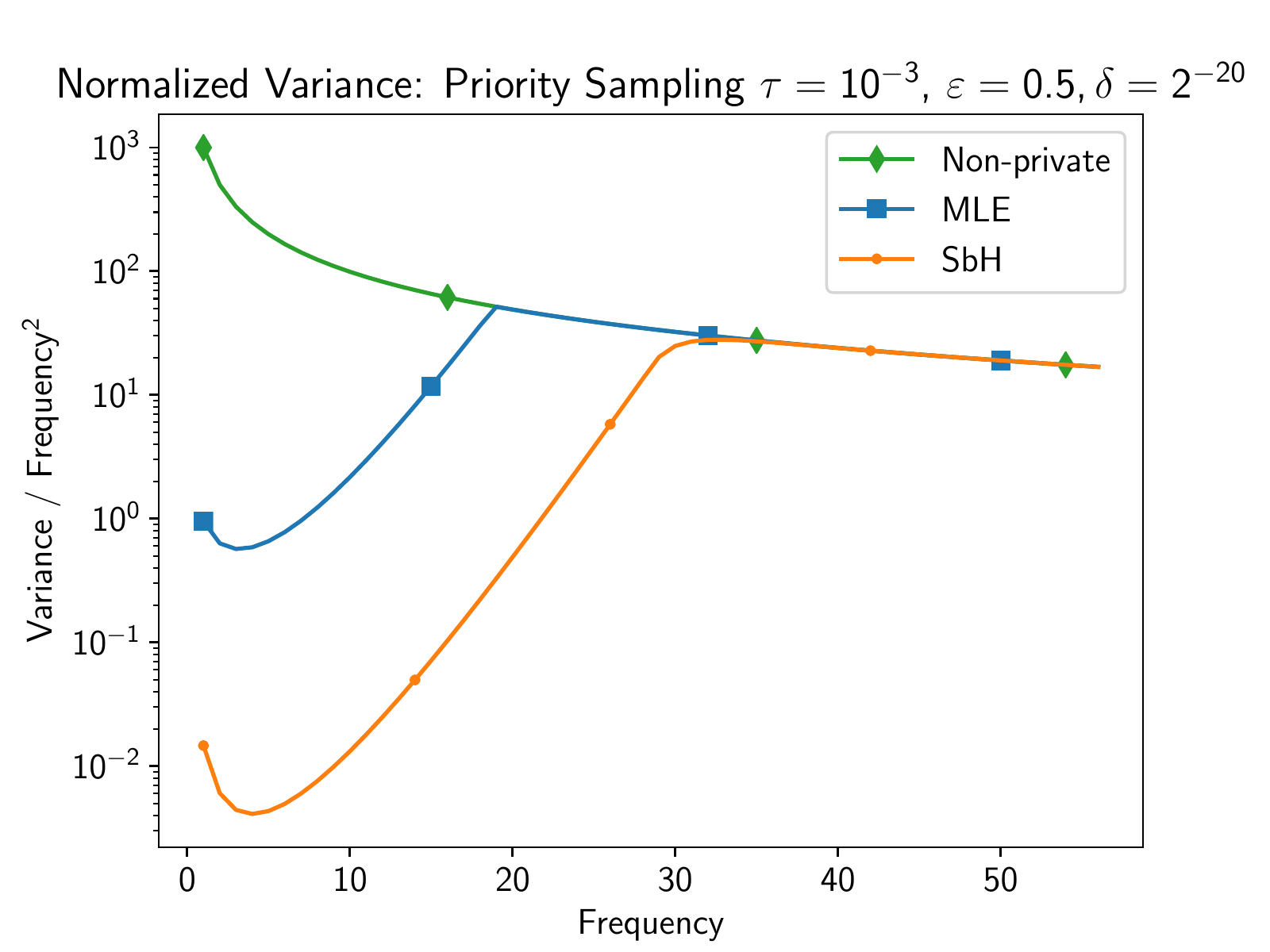}
\includegraphics[width=0.44\textwidth]{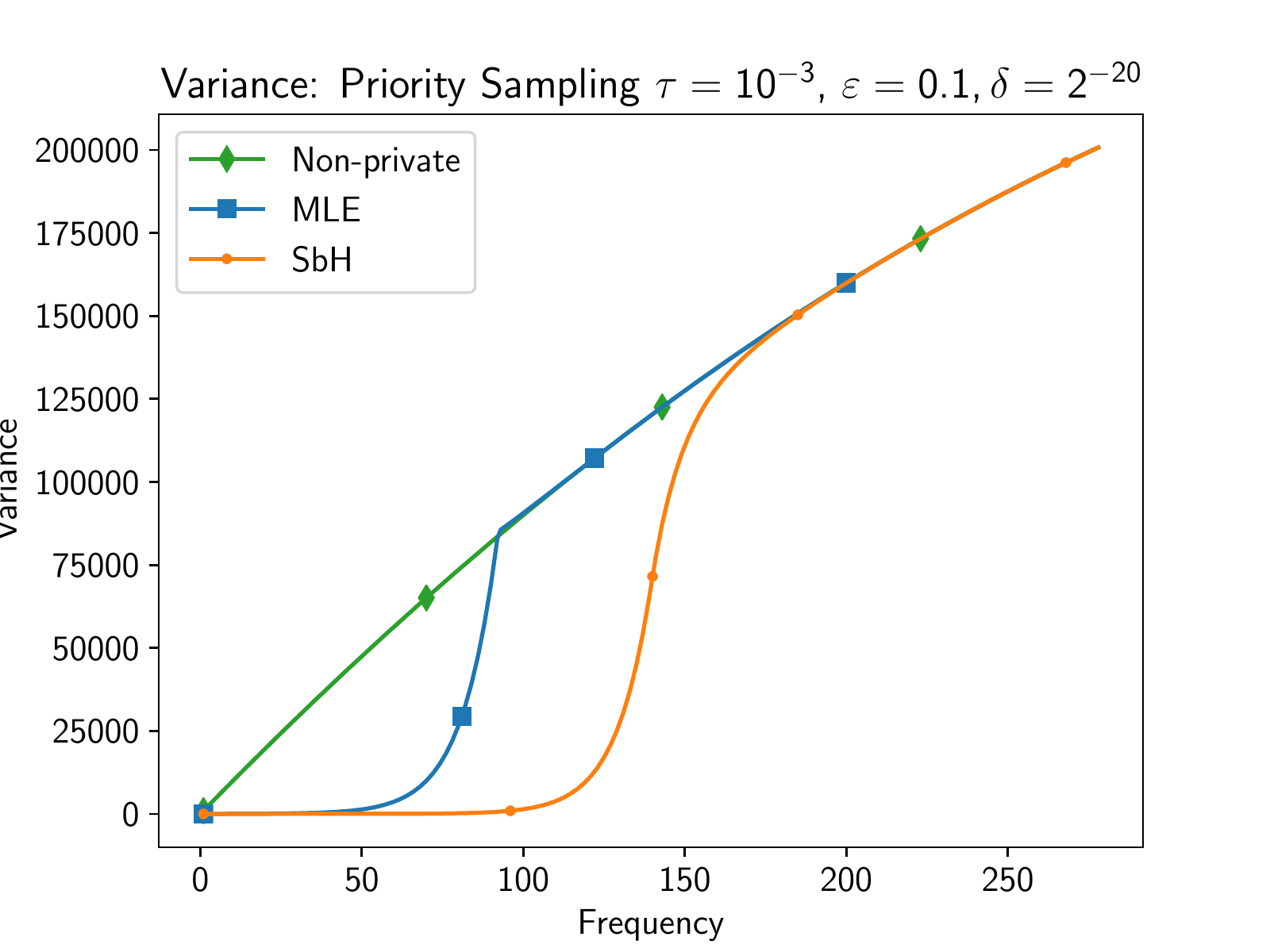}
\includegraphics[width=0.44\textwidth]{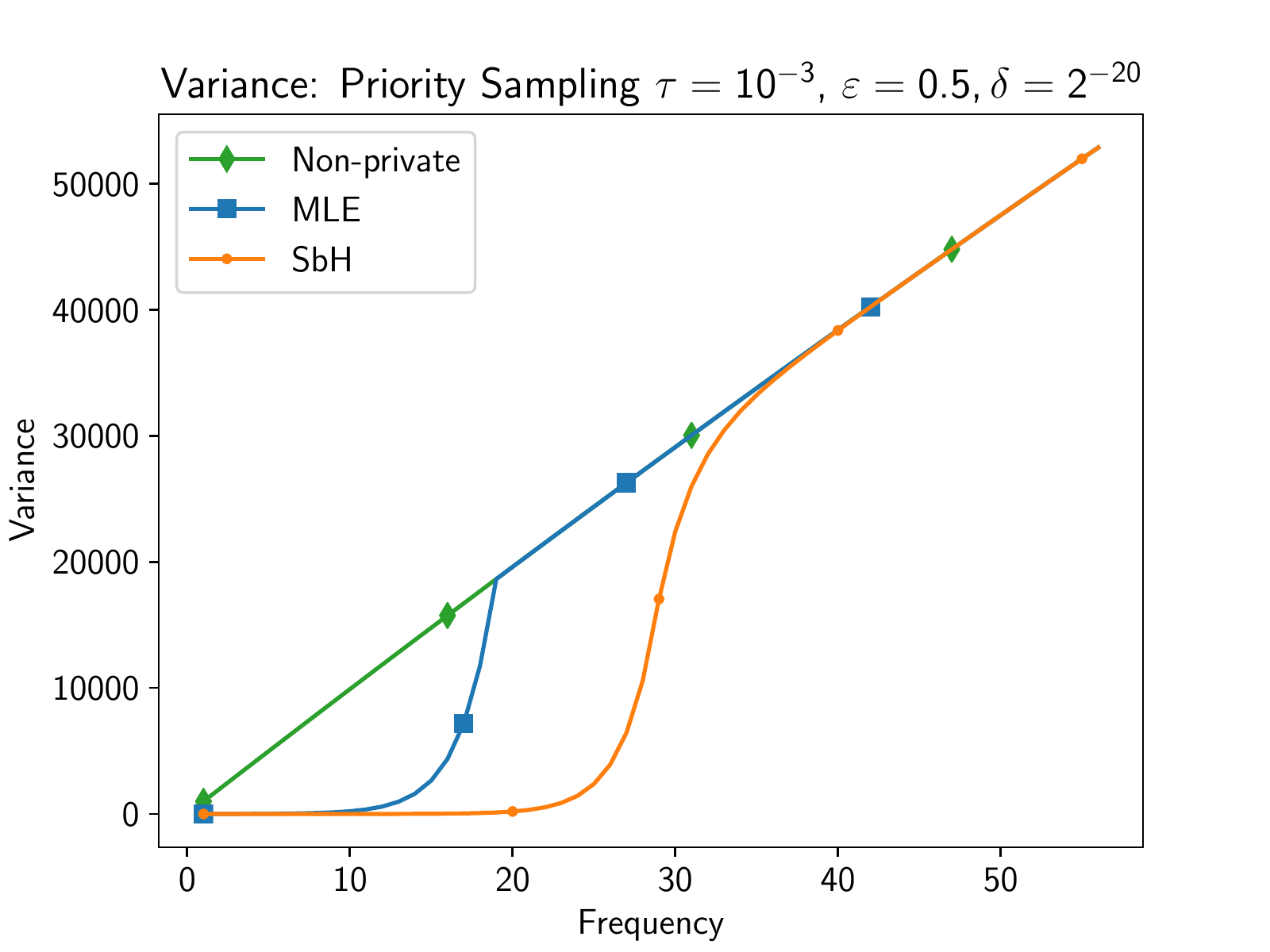}
}
\caption{ Normalized variance $\Var_i/i^2$ \notinproc{and variance $\Var_i$} for PWS (MLE) and sampled-SbH as a function of the frequency $i$.
}
\label{variance:fig}
\end{figure}

Figure~\ref{variance:fig} shows the normalized variance $\Var_i/i^2$ per frequency $i$ for representative parameter settings.
The private methods PWS and sampled-SbH maintain low variance across frequencies:  The value is fractional with no sampling and is of the order of that of the non-private unbiased estimator with sampling.  
In particular this means that the bias is a good proxy for performance and that the improvement in bias of PWS with respect to baseline does not come with a hidden cost in variance. For high frequencies (not shown), keys with all methods are 
included with probability (close to) $1$. The non-private method that reports exact frequencies have $0$ variance whereas the private methods maintain a low variance, but the normalized variance diminishes for all methods.

For statistics estimation, the per-key performance suggest that
when the selection has many high frequency keys, the private methods perform well and are similar to non-private sampling.
When the selection is dominated by very low frequencies, the private methods perform poorly and well below the respective non-private sample.  But for low to medium frequencies,  PWS can provide drastic improvements over SbH and the gain increases with lower sampling rates.  Figure~\ref{bias_error:fig} (bottom) shows the $\NRMSE$ as a function of sampling rate for estimating the sum of frequencies on a selection of $2\times 10^5$ keys with
frequencies uniformly drawn between $1$ and $200$.
We can see that the error of non-private sampling and of sampled-SbH decreases with higher sampling rate.  Note the perhaps counter-intuitive phenomenon that PWS (MLE) hits its sweet spot midway:  This is due to a balance of the two components of the error, the variance which increases and the bias that decreases when the sampling rate decreases.  Also note that PWS significantly improves over SbH also with no sampling ($\tau=1$).  

\vspace{-0.24cm}
\section*{Conclusion}
We presented 
Private Weighted Sampling (PWS), a method to post-process a weighted sample and produce a version that is differentially private.  Our private samples maximize the number of reported keys subject to the privacy constraints and support estimation of linear and ordinal statistics.  We demonstrate significant improvement over prior methods for both reporting and estimation tasks, even for the well studied special case of private histograms (when there is no sampling).

An appealing direction for future work is to explore the use of PWS to design \emph{composable} private sketches, e.g., in the context of coordinated samples. 
Threshold and bottom-$k$ samples of different datasets are {\em coordinated} when using consistent $\{u_x\}$. Coordinated samples generalize MinHash sketches and support estimation of similarity measures and statistics over multiple datasets \cite{BrEaJo:1972,Saavedra:1995,ECohen6f,Broder:CPM00,sdiff:KDD2014,sorder:PODC2014}.

\subsubsection*{Acknowledgments}
Part of this work was done while Ofir Geri was an intern at Google Research. This work was partially supported by Moses Charikar's Google Faculty Research Award. We also thank Chinmoy Mandayam for bringing the related paper \cite{GhoshRS:sicomp2012} to our attention.

\bibliographystyle{plain}
\bibliography{main,cycle}

\onlyinproc{\end{document}}
\appendix

\section{Proofs: Sanitized Keys} \label{proofkeybasic:sec}
 
We establish that $(\pi_i)_{i\geq 1}$ as computed by Algorithm~\ref{alg:reportkeys} are maximum under the DP constraints and are non-decreasing.
\begin{proof} [Proof of Lemma~\ref{basicpi:lemma}]
We use the notation $\overline{\pi_i} := 1-\pi_i$ for the probability of a key not being included in $C(A())$.  
Since key inclusions in the (original or sanitized) sample are independent, 
the probability of a particular sanitized sample $Z$ (set of keys, possibly empty) has the product form 
 \[P_{\boldsymbol{w}}(Z) := \prod_{x\in Z} \pi_{w_x} \prod_{x\not\in Z} \overline{\pi_{w_x}} \enspace . \]
 
 Since $\pi_i=q_i p_i$ is the product of two probabilities, one that is given ($q_i$) and one that we set ($p_i$) then our solution for $(\pi_i)_{i\geq 1}$ is realizable if and only if it satisfies the constraints for all $i$:
\begin{equation} \label{qbound:eq}
\pi_i \leq q_i \enspace .
\end{equation}

We now consider the DP constraints.
Consider two neighboring datasets $\boldsymbol{w}$ and
 $\boldsymbol{w}'$ and the two cases (i) For some $i\geq 1$ there is a key $x$ such that $w_x=i$ and $w'_x = i-1$  (ii) For some $i\geq 0$ there is a key $x$ such that  $w_x = i$ and $w'_x = i+1$.
 
 We consider sets $T$ of possible outputs that we partition to outputs $T^+$ that include the key $x$  and outputs $T^-$ that do not include $x$. 
 We use the notation 
 \begin{align*}
    Q^+ &= \sum_{Z\in T^+} P_{\boldsymbol{w}_{-x}}(Z\setminus\{x\})\\
  Q^- &= \sum_{Z\in T^-} P_{\boldsymbol{w}_{-x}}(Z)\\
 \end{align*}
 for the respective sums over these sets of outputs of the probability projected on keys other than $x$.
 
 The general DP constraints on a set of outputs $T$ have one of the following form corresponding to our two cases:
 \begin{align}
 Q^+ \pi_i + Q^- \overline{\pi_i} &\leq e^\epsilon (Q^+ \pi_{i-1} + Q^- \overline{\pi_{i-1}}) + \delta \label{firstset:eq}\\
 Q^+ \pi_i + Q^- \overline{\pi_i} &\leq e^\epsilon (Q^+ \pi_{i+1}  + Q^- \overline{\pi_{i+1}}  )+  \delta \enspace .\label{secondset:eq}
 \end{align}
 
We observe, assuming monotonicity of $\pi_i$,  that constraints \eqref{firstset:eq}  are strictest when $Q^-$ is as small as possible.  
This because $\overline{\pi_i}\leq \overline{\pi_{i-1}}$ implies 
$Q^- \overline{\pi_i}\leq e^\varepsilon Q^- \overline{\pi_{i-1}}$ (using $e^\varepsilon\geq 1$), so the larger $Q^-$ is, the less strict the inequality becomes.
This
is achieved when $T^{-} = \emptyset$ and thus $Q^-=0$, that is, $T$ only includes outputs that include $x$.  Similarly, assuming $\pi_i \geq \pi_{i-1}$, the constraints are strictest when $Q^+$ is as large as possible.  That is $T^+$ includes all possible outcomes on keys other than $x$ and thus $Q^+=1$.   A similar argument shows that constraints \eqref{secondset:eq} are strictest when $Q^+=0$ and $Q^-=1$.

We obtain that the DP constraints simplify to
 \begin{align}
 i\geq 1 \text{:}\,\; \pi_i &\leq e^\epsilon \pi_{i-1} + \delta \label{prehalf:eq}\\
 i\geq 0 \text{:}\,\;  \overline{\pi_i} &\leq e^\epsilon \overline{\pi_{i+1}} +  \delta \enspace .\label{posthalf:eq}
 \end{align}
 We show below that the solution $(\pi_i)_{i\geq 1}$ of the simplified constraints (as subset of all constraints) is non-decreasing  and hence any solution to the full set of constraints must also be non-decreasing and thus the assumption that led to the simplification is valid.
 
 We observe that the constraints \eqref{prehalf:eq} and \eqref{posthalf:eq} are upper bounds on $\pi_i$ in terms of $\pi_{i-1}$ and the feasibility constraint \eqref{qbound:eq} is also an upper bound on $\pi_i$. Therefore, each iterate $\pi_i$ computed in Algorithm~\ref{alg:reportkeys} attains the maximum possible value by the constraints, provided that $\pi_{i-1}$ is at its maximum value. The claim that each $\pi_i$ is maximized follows by induction.
 
 Finally, to establish that $(\pi_i)_{i\geq 1}$ are non-decreasing we show that each term in the minimum that determines $\pi_{i+1}$ is at least $\pi_{i}$:
 \begin{itemize}
\item \eqref{qbound:eq}:  $q_{i+1} \geq q_i \geq \pi_i$
\item \eqref{prehalf:eq}: $e^\varepsilon \pi_i + \delta \geq \pi_i$
\item \eqref{posthalf:eq}: $1+ e^{-\varepsilon}(\pi_i+\delta-1) = (1- e^{-\varepsilon}) + e^{-\varepsilon}(\pi_i+\delta) \geq (1- e^{-\varepsilon})\pi_i + e^{-\varepsilon}(\pi_i+\delta) = \pi_i + e^{-\varepsilon}\delta \geq \pi_i$
\end{itemize}

  \end{proof}

Consider the iterates $\pi_i$ and the corresponding constraints sequence of the minimum among the three constraints: \eqref{qbound:eq}, \eqref{prehalf:eq}, and~\eqref{posthalf:eq}. 
We will slightly abuse notation and use these references to constraints in expressions.  We first consider the relation between \eqref{prehalf:eq} and \eqref{posthalf:eq}:
\begin{lemma} \label{twoandthree:lemma}
The constraints sequence has all positions with \eqref{prehalf:eq} preceding all positions with  \eqref{posthalf:eq}. In the typical settings of $ \epsilon \ll 1$, the highest position $i$ before the transition has $\pi_i \leq  (1-\delta)/2$.
\end{lemma}


\begin{proof}
The ratio of constraints as a function of $x=\pi_i$ is:
\[\frac{\text{\eqref{prehalf:eq}}}{\text{\eqref{posthalf:eq}}} = \frac{e^{\varepsilon}x+\delta}{1+e^{-\varepsilon}\left(x+\delta-1\right)} \enspace .\]
This is an increasing function for $x\in (0,1]$. 
Therefore the iterates $\pi_i$ are such that initially \eqref{prehalf:eq} is smaller (ratio is lower than $1$) and then 
\eqref{posthalf:eq} is smaller (ratio is larger than $1$).
Solving for the crossing point (ratio equal to one) we get \[x = \frac{1-\delta}{1+e^{\varepsilon}} \approx \frac{1-\delta}{2+\varepsilon} 
\enspace ,\]
using the first order approximation $e^{z}\approx 1+z$ which holds when $z \ll 1$. Also note that $x\leq (1-\delta)/2$.
\end{proof}

A subsequence of $(\pi_i)$ where all constraints are \eqref{prehalf:eq} or all \eqref{posthalf:eq} has a compact form:
\begin{lemma} \label{onlypreorpost:lemma}
The iterates $\pi_{i+1} = e^\varepsilon \pi_i + \delta$ on a sub-sequence with only \eqref{prehalf:eq} that starts at $i_0$  can be compactly expressed for $i>i_0$:
\begin{equation} \label{presubsequence:eq}
    \pi_i =  \pi_{i_0} e^{(i-i_0)\varepsilon} +\delta\frac{e^{(i -i_0)\varepsilon} -1}{e^\varepsilon-1}\enspace.
\end{equation}
Similarly, for a sub-sequence with only \eqref{posthalf:eq} constraints
where $\overline{\pi_{i+1}} = e^{-\varepsilon}(\overline{\pi_{i}} -\delta)$
we get:
\begin{equation} \label{complement_constraints:eq}
    \overline{\pi_{i}} = \overline{\pi_{i_0}} e^{-(i-i_0)\varepsilon} - \delta e^{-\varepsilon}  \frac{1- e^{-(i-i_0)\varepsilon}}{1- e^{-\varepsilon}}\enspace .
\end{equation}
 \end{lemma}
\begin{proof}
The iterates form a geometric series.
\end{proof}

We now establish the closed-form expressions of the solution $\pi^*_i$ that corresponds to $q_i=1$ for all $i$. 
\begin{proof} [Proof of Lemma~\ref{privatehist:lemma}]
Since $q_i=1$ for all $i$, the constraint sequence includes only \eqref{prehalf:eq} and \eqref{posthalf:eq} constraints (until the minimum exceeds $1$, in which $\pi_i=1$ at this position and all subsequent positions).  From Lemma~\ref{twoandthree:lemma} we know it has the form \eqref{prehalf:eq}$^*$\eqref{posthalf:eq}$^*$.  We have $\pi^*_1 = \delta$.

 We have $\pi^*_1 = \delta$ and hence while the constraint \eqref{prehalf:eq} holds. Using \eqref{presubsequence:eq} (Lemma~\ref{onlypreorpost:lemma})  we have
 $\pi^*_i = \delta \frac{e^{i\varepsilon} -1}{e^\varepsilon-1}$.
 From the proof of Lemma~\ref{twoandthree:lemma}, we
 have \eqref{prehalf:eq} in the constraint sequence until
 $\pi^*_i > \frac{1-\delta}{1+e^{\varepsilon}}$.
 From our choice of $L$, we have
 \[
 \pi^*_L = \delta \frac{e^{L\varepsilon} -1}{e^\varepsilon-1} = \frac{1-\delta}{1+e^{\varepsilon}}\ .
 \]
 Therefore, both \eqref{prehalf:eq} and \eqref{posthalf:eq} hold at position $L$.
  We have $\pi^*_{L+1} = e^\varepsilon \pi^*_L + \delta$.  From our choice of $L$ we
 have $\overline{\pi^*_{L+1}} = \pi^*_L$ and $\overline{\pi^*_{L}} = \pi^*_{L+1}$.
 We apply \eqref{complement_constraints:eq} (Lemma~\ref{onlypreorpost:lemma}) with $i_0 = L$ to obtain the claim.
 Note the symmetry of the solution where for $1\leq i\leq L$, $\overline{\pi^*_{2L+1-i}} = \pi^*_{i}$.
 

\end{proof}

We are now ready to bound the number of positions $i$ where $\pi_i < q_i$:
\begin{proof} [Proof of Lemma~\ref{totaltwothree:lemma}]
We have $\pi_i < q_i$ if and only if the $i$th position in the constraint sequence has \eqref{prehalf:eq} or \eqref{posthalf:eq}.

The sequence of $\pi_i$ is non-decreasing with at most one transition from 
 \eqref{prehalf:eq} to \eqref{posthalf:eq}.  For $i$ such that $q_i\leq \delta$ we have the constraint \eqref{qbound:eq}. Hence $\pi_i \geq \delta = \pi^*_1$ at the first position with $\pi_i < q_i$.
  Each application of the minimum of \eqref{prehalf:eq} and \eqref{posthalf:eq} increases $\pi_i$.  The total increase is larger when the initial value is larger.
 Let $h$ be the $i$th position with \eqref{prehalf:eq} or  \eqref{posthalf:eq}. Because of the monotone increase we can show by induction that $\pi_h \geq \pi^*_i$.  
 Since there are $2L+1$ positions in $(\pi^*_i)_{i\geq 1}$ with value $\pi^*_i<1$,  there can be at most $2L+1$ positions in $(\pi_i)_{i\geq 1}$ with $\pi_i< q_i$.
\end{proof}

With threshold sampling (say by moments of frequency) we have a closed form for $q_i$.
Using this and Lemma~\ref{onlypreorpost:lemma} we can express the solution $\pi_i$ with computation that depends on the number of transitions in the constraint sequence.  The following Lemma bounds the number of such transitions.  The proof of  Lemma~\ref{niceq:lemma} follows as a special case:
\begin{lemma}
For threshold ppswor with $p=1$ the constraints sequence has the regular-expression form
\[
\text{\eqref{prehalf:eq}$^*$} \text{\eqref{posthalf:eq}$^*$} \text{\eqref{qbound:eq}$^*$} \enspace .
\] 
For priority sampling with $p\leq 1$ and for priority with $p\geq 1$ when $\tau\geq \delta$, all \eqref{qbound:eq} constraints must follow all \eqref{prehalf:eq} constraints in the constraint sequence.
\end{lemma}
\begin{proof} 
From Lemma~\ref{twoandthree:lemma} there is at most one transition from \eqref{prehalf:eq} to \eqref{posthalf:eq}.

We now consider the relation between \eqref{qbound:eq} and \eqref{prehalf:eq}. 
When $\pi_i=q_i$, we will have \eqref{qbound:eq}$\leq$\eqref{prehalf:eq}  if 
\begin{equation} \label{ratio:eq}
    \rho(i,\delta) := \frac{q_{i+1}-\delta}{q_i}\leq e^\varepsilon \enspace .
\end{equation}

We need to establish that once \eqref{ratio:eq} holds for $i=i_0$, it continues to hold for $i\geq i_0$.   Equivalently, establishing the claim for all $\epsilon>0$ is equivalent to establishing that $\rho(i,\delta)$ is non-increasing with $i$ when $\rho(i,\delta)> 1$.
That is, that $\rho(i,\delta)$ is non-increasing, equivalently, that the partial derivative satisfies
\begin{equation} \label{rhononincreasing:eq}
    \frac{\partial\rho(i,\delta)}{\partial i} \leq 0
\end{equation}
when 
\begin{equation} \label{diffcond:eq}
    q_{i+1}-q_i \geq \delta \enspace .
\end{equation}
In some of the derivations i will be convenient to work with the continuous form of \eqref{diffcond:eq}:
\begin{equation} \label{diffcondcont:eq}
\frac{\partial q_i}{\partial i} \geq \delta\enspace .
\end{equation}

Consider ppswor threshold sampling with $p=1$.  Recall that
$q_i=1-e^{-\tau i}$.  Therefore, the condition \eqref{diffcond:eq} is $e^{-\tau i}(1-e^{-\tau})\geq \delta$.  It suffices to check \eqref{rhononincreasing:eq}  when $\delta + e^{-\tau} \leq 1$.
Substituting and solving \eqref{rhononincreasing:eq} we obtain that the derivative is negative when
$\delta + e^{-t} \leq 1$.  Therefore, there can be at most one transition from \eqref{prehalf:eq} to \eqref{qbound:eq} for ppswor with $p=1$.

We next consider priority threshold sampling $q_i = \min\{1,\tau i^p\}$.
For $i^p\tau  \geq 1$, $q_i=1$ and \eqref{diffcond:eq} does not hold.
Therefore it suffices to consider $q_i = \tau i^p < 1$ and
\[\rho(i,\delta) = \frac{(i+1)^p - \frac{\delta}{\tau}}{i^p}\enspace .\]

When $p\leq 1$, $(i+1)^p-i^p \leq 1$ and thus  condition \eqref{diffcond:eq} does not hold when $\tau\leq \delta$.
Hence it suffices to consider $p\geq 1$ or $\delta < \tau$.
Using the continuous form \eqref{diffcondcont:eq} with  $q_i = \tau i^p$ we get that it is satisfied when 
\begin{equation} \label{fromcont:eq}
    i^{p-1} \geq \delta/(p\tau)\enspace .
\end{equation}

By solving \eqref{rhononincreasing:eq} we get
\[
(i+1)^{p-1} > \frac{\delta}{\tau}\enspace .
\]
This holds for all $i\geq 1$ and $p\geq 1$ when $\delta\leq \tau$.
Since it suffices to for the solution to hold for \eqref{fromcont:eq}, we obtain the claim for $p\leq 1$.
Combining, we obtain that there can be at most one transition from \eqref{prehalf:eq} to \eqref{qbound:eq} for priority sampling with $p\leq 1$ and for $p\geq 1$ when $\tau\geq\delta$.

We next consider the relation between \eqref{qbound:eq} and \eqref{posthalf:eq}.  
We define
\[
\overline{\rho}(i,\delta) := \frac{\overline{q_{i}} - \delta}{\overline{q_{i+1}}}\enspace .
\]
When $\pi_i = q_i$, we will have
\eqref{qbound:eq}$\leq$\eqref{posthalf:eq}  if 
\[
\overline{\rho}(i,\delta) \leq e^{\varepsilon}\enspace .
\]
To establish the claim for all $\varepsilon >0$ it is equivalent to establish that 
$\overline{\rho}(i,\delta)$ is non-increasing when it is greater than $1$.
Equivalently, that $\overline{q_i} - \overline{q_{i+1}} = q_{i+1}-q_i > \delta$ (same as \eqref{diffcond:eq} and \eqref{diffcondcont:eq}) implies that
$\overline{\rho}(i,\delta)$ is non-increasing.  That is,
\begin{equation} \label{rhobardecreasing:eq}
\frac{\partial{\overline{\rho}(i,\delta)}}{\partial i}  \leq 0\enspace .
\end{equation}

For ppswor with $p=1$ we get 
\[
\overline{\rho}(i,\delta) = \frac{e^{-\tau i} -\delta}{e^{-\tau(i+1)}}
\]
and that \eqref{rhobardecreasing:eq} holds for all $\delta \geq 0$ and $\tau >0$.

\end{proof}

\section{Proofs: Sanitized Keys and Frequencies} \label{proofssanitizedKF:sec}

We establish properties of the values $(\pi_{i,j})$ computed by Algorithm~\ref{alg:reportfreq}.
  \begin{proof} [Proof of Theorem~\ref{theorem:reported-freq-dist}]
  We write the DP constraints in terms of sets $T$ of potential outputs on pairs of neighboring datasets $\boldsymbol{w}$ and $\boldsymbol{w}'$.  We follow the proof of Lemma~\ref{basicpi:lemma}. 
  Consider two neighboring datasets $\boldsymbol{w}$ and
 $\boldsymbol{w}'$ and the two cases: (i)~For some $i\geq 1$ there is a key $x$ such that $w_x=i$ and $w'_x = i-1$. (ii)~For some $i\geq 0$ there is a key $x$ such that  $w_x = i$ and $w'_x = i+1$.  
 
 We consider a set of outputs $T$.
   A potential output $Z\in T$ is a set of key value pairs $(y,j)$ where key $y$ is reported with value $j$.  For purposes of this proof we partition $T$ to sets $T_j$ according to the output on key $x$. If $(x,j)\in Z$ we place $Z\in T_j$ and if key $x$ is not in $Z$ we place $Z$ in $T_0$.
  
  We denote by $Q_j$ the respective combined probability of outputs $T_j$ when projected on all keys other than $x$ (equivalently, the probability of $T_j$ when key $x$ is removed from the dataset).   The general DP constraints have the form
  \[
  \Pr[C(A(\boldsymbol{w}))\in T] \leq e^\varepsilon  \Pr[C(A(\boldsymbol{w}'))\in T]+ \delta \enspace .
  \]
  We have
  \[
  \Pr[C(A(\boldsymbol{w}))\in T] = \sum_{j=0}^i Q_j \pi_{i,j}\enspace .
  \]
  
  From the two choices of the neighboring dataset $\boldsymbol{w}'$ we have one of:
  \begin{align*}
      \Pr[C(A(\boldsymbol{w}'))\in T] &= \sum_{j=0}^{i-1} Q_j \pi_{i-1,j} \\
  \Pr[C(A(\boldsymbol{w}'))\in T] &= \sum_{j=0}^{i+1} Q_j \pi_{i+1,j} \enspace .
  \end{align*}
  
  
  The corresponding DP constraints are:
      \begin{align}
  i\geq 1 \text{ : }\;    \sum_{j=0}^i Q_j \pi_{i,j} \leq e^\epsilon \sum_{j=0}^{i-1} Q_j \pi_{i-1,j} + \delta \label{down:eq} \\
  i\geq 1 \text{ : }\;    \sum_{j=0}^i Q_j \pi_{i,j} \leq e^\epsilon \sum_{j=0}^{i+1} Q_j \pi_{i+1,j} + \delta  \enspace .\label{up:eq}
  \end{align}
  
  Considering any particular set of values $\pi_{i,j}$, the strictest constraints of the form \eqref{down:eq} would have $Q_j=1$ for $j$ where $\pi_{i,j} > \pi_{i-1,j}$ and $Q_j=0$ otherwise.
  Similarly for constraints of the form~\eqref{up:eq}, the strictest would have $Q_j=1$ for $j$ where $\pi_{i,j} > \pi_{i+1,j}$ and $Q_j=0$ otherwise. By "strictest" constraints we mean that if the values $\{\pi_{i,j}\}$ satisfy these selected constraints, the satisfy all DP constraints.
  
  Therefore, taking the union of all these "strictest" sets of constraints over all $\{\pi_{i,j}\}$ we obtain that  without loss of generality it suffices to solve for constraints of the following form: For $i\geq 1$ and all $J\subset \{0,\ldots,i\}$: 
  \begin{align}
 \sum_{j\in J} \pi_{i,j} &\leq e^\epsilon \sum_{j\in J} \pi_{i-1,j} + \delta \label{upc:eq} \\
 \sum_{j\in J} \pi_{i,j} &\leq e^\epsilon \sum_{j\in J} \pi_{i+1,j} + \delta \enspace . \label{downcp:eq}
 \end{align}
  We re-arrange the set of constraints \eqref{downcp:eq} and get the equivalent set for all $i$ and $J\subset \{0,\ldots,i\}$:
  \begin{equation}\label{downc:eq}
  \sum_{j\in J} \pi_{i,j} \geq e^{-\varepsilon} \left(\sum_{j\in J} \pi_{i-1,j} - \delta\right)\ .
 \end{equation}
For sets $J$, the constraints
  \eqref{upc:eq} and \eqref{qbound2:eq} determine upper bounds on $\sum_{j\in J}\pi_{i,j}$ as determined by $\sum_{j\in J} \pi_{i-1,j}$ for $j\in J$. 
The constraints 
    \eqref{downc:eq} determine lower bounds.

The solution is constructed by increasing $i$, so that row $i$ is set after rows $h<i$ are set.
We first set $\pi_{i,0} = 1-\pi_i$. We set other entries so that 
\begin{equation} \label{sumcond:eq}
\sum_{j=1}^i \pi_{ij} = \pi_i
\end{equation}
(we will establish inductively that we can always satisfy \eqref{sumcond:eq}).  
Note that such settings imply that \eqref{qbound2:eq} are satisfied.  
At the high level, we set the entries $\pi_{i \bullet}$ using two passes:
The first pass is performed in increasing order by $j\geq 1$ and we greedily compute the minimum values $D_{i,j}$ we can have for $\pi_{i,j}$ so that constraints of type \eqref{downc:eq} are satisfied.
The second pass is performed in decreasing order by $j\leq i$ and greedily sets $\pi_{i,j}$ to the maximum value we can while satisfying the constraints \eqref{upc:eq} and \eqref{qbound2:eq}.  The second pass finishes at an entry $h$
so that $\sum_{j=1}^{h-1} D_{i,j} + \sum_{j=h}^i \pi_{i,j} = \pi_i$.  We set all entries $1\leq j<h$ to $\pi_{i,j} \gets D_{i,j}$.  A subtlety is that during the passes we only use the respective constraints \eqref{downc:eq} and \eqref{upc:eq} for $J$ of prefix or suffix forms, but we establish that the constructed solution satisfies these constraints for all $J$.

We now elaborate on the first pass.
The value $D_{i,j}$ for $j\geq 1$ is set to the minimum needed for $\pi_{i,j}$
so that a setting of $\pi_{i,h} \gets D_{i,h}$ for $h<  j$
satisfies
\eqref{downc:eq} for $J=\{0,\ldots,j\}$ and $J=\{1,\ldots,j\}$ (We refer to such $J$ as being in {\em prefix form}).  Specifically, for
$j=1,\ldots,i-1$ in order we get:
  \begin{align*}
  D_{i,j} \gets \max\{0 &, e^{-\varepsilon} \left(\sum_{h=1}^j \pi_{i-1,h} - \delta\right) -\sum_{h=1}^{j-1} D_{i,h} + \\
  & \max\{0,e^{-\varepsilon}\pi_{i-1,0} - \pi_{i,0}\} \} \enspace .
  \end{align*}
Let the breakpoint $b_1$ be maximum such that  
\[
\sum_{j=1}^{b_1} \pi_{i-1,j} + \max\{0,e^{-\varepsilon} \pi_{i-1,0}-\pi_{i,0}\} \leq \delta \enspace .\]
Then $D_{i\bullet}$ has $D_{i,j}=0$ for $j\leq b_1$ and
$D_{i,j}= e^{-\varepsilon} \pi_{i,j}$ for $j> b_1+1$.
It follows that
\begin{equation}\label{LBbelow:eq}
\forall j\geq 1,\, D_{i,j} \leq e^{-\varepsilon} \pi_{i-1,j} \enspace .
\end{equation}
Also note (from the form of the constraints) that for all $j\geq 1$,
$\sum_{h=1}^j D_{i,j}$ is a lower bound on $\sum_{h=1}^j \pi_{i,j}$ for all $\pi_{i\bullet}$ that satisfy \eqref{downc:eq} on all $J$ with prefix form.  That is, we could not get a lower prefix with a non-greedy setting of $D_{i,j}$.

We now provide details on the second pass. We visit entries $j\leq i$ in decreasing order and set
$\pi_{i,j} > D_{i,j}$ as follows.
Let \[D_i \gets \sum_{h=1}^{i-1} D_{i,h}\enspace .\] 
From \eqref{LBbelow:eq} we get 
\begin{equation} \label{LBsumbelow:eq}
    D_i \leq \sum_{h=1}^{i-1} \pi_{i-1,h} = \pi_{i-1} \leq \pi_i\enspace .
\end{equation}
We compute the initial probability mass to allocate $R\gets \pi_i-D_i$.  From \eqref{LBsumbelow:eq}
we get that $R\geq 0$.   While allocating we maintain that
$\sum_{h=j}^i (\pi_{i,h}-D_{i,h}) \leq R$.
For $j$, we first compute the maximum value $U$ we can have for $\pi_{i,j}$ given values we already set for $\pi_{i,h}$ for $h>j$ so that the constraint \eqref{upc:eq} with $J = \{j,\ldots,i\}$ is satisfied:
  \begin{equation*}
   U \gets e^\varepsilon \sum_{h=j}^{i-1}\pi_{i-1,h} + \delta -\sum_{h=j+1}^i \pi_{i,h}\enspace .
  \end{equation*}
We then compute the increase $\Delta \gets U-D_{i,j}$.  If $\Delta\leq R$
we set $\pi_{i,j} \gets U$ and $R \gets R- \Delta$.  Otherwise, we set $\pi_{i,j}\gets D_{i,j}+R$ and $R\gets 0$.  The pass terminates when $R=0$ and we let $b_2$ be the $j$ value when the pass terminates. ($\pi_{i,j}$ for $1\leq j< b_2$ are set to $D_{i,j}$).
The solution has the following structure.  If initially $R\leq \delta$, the second pass stops at $b_2=i$. Otherwise, $\pi_{i,i}=\delta$ and for $b_2<j \leq i-1$, $\pi_{i,j} = e^\varepsilon \pi_{i-1,j}$.
Note that (due to the form of the constraints) is that the construction maximizes
$\sum_{h=j}^i \pi_{i,j}$ for all $\pi_{i\bullet}$ that satisfy \eqref{upc:eq} on all $J$ with suffix form.

  We next establish that the constructed $\pi_{i\bullet}$ satisfy \eqref{sumcond:eq}.  For that, we need to show that we can always exhaust $R$.
 Note that if the situation is that current values have
  $\sum_{j=1}^i \pi_{i,j} < \pi_i$ we must have slack and $U>\pi_{i,j}$ in the constraints. This since $\sum_{j=1}^i \pi_{i,j} < \pi_i \leq e^\varepsilon \pi_{i-1}$. We obtain $U=\pi_i-\sum_{j=2}^i \pi_{i,j} > \pi_{i,1}$ when processing $j=1$.
  
  Finally, we need to establish that the constructed $\pi_{i\bullet}$ satisfies constraints \eqref{downc:eq} and \eqref{upc:eq} for all $J$.
  
  From the construction, we have that any solution with $\pi_{i,j}\geq D_{i,j}$ for all $j$, and in particular, the one constructed, satisfy \eqref{downc:eq} for all $J$ that has the form
  $\{0,\ldots,h\}$ or $\{1,\ldots,h\}$ for some $h$.  Consider now an arbitrary $J$.  All indices with $j>b_1$ have $\pi_{i,j} \geq e^{-\varepsilon} \pi_{i-1,j}$. Therefore $J'\gets J\setminus \{b_1+1,\ldots,i\}$ is such that if \eqref{downc:eq} holds for $J'$, it must hold for $J$.
  The entries $1\leq j<b_1$ have $\pi_{i,j}=0$.  Thus $J'' \gets J'\cup \{1,\ldots,b_1-1\}$ is such that if \eqref{downc:eq} holds for $J''$, it must hold for $J'$.  Now note that $J''$ has a prefix form  starting at $0$ or $1$ and ending at $b_1-1$ or $b_1$.  Therefore, \eqref{downc:eq} holds for $J''$ and hence, also holds for $J$.
  
  From the construction, we know that \eqref{upc:eq} is satisfied for any $J$ that has a suffix form
  $(h,\ldots,i)$ for $h\geq b_2$.
  Consider now an arbitrary $J$.  Now recall that for all $j<i$, $\pi_{i,j} \leq e^\varepsilon \pi_{i-1,j}$.  Therefore, if \eqref{upc:eq} holds for $J'=J\setminus \{0,\ldots,i-1\}$, it must hold for $J$.  But either  $J'=\{i\}$ or $J'$ is empty and in both cases \eqref{upc:eq} holds.
      \end{proof}

\begin{proof}[Proof of Lemma~\ref{formij:lemma}]
We need to show that $(\pi^*_{i,j})_{j= \max\{1,i-2L\}}^i$ 
is a suffix of the following sequence of length $2L+1$:
\begin{equation} \label{form:eq}
\delta\cdot(1,e^\varepsilon,\ldots,e^{\varepsilon(L-1)},e^{\varepsilon L},e^{\varepsilon(L-1)},\ldots,e^\varepsilon,1)
\end{equation}

We inductively verify that the construction in Algorithm~\ref{alg:reportfreq} is such that if form \eqref{form:eq} holds for $\pi^*_{i,\bullet}$, it also holds for $\pi^*_{i+1,\bullet}$.
We use properties of the solution outlined in the proof of Theorem~\ref{theorem:reported-freq-dist}.

Note that for $i\leq 2L+1$, $\pi^*_i$ is equal to the sum of the length-$i$ suffix of the sequence and for $i> 2L+1$ we have $\pi^*_i=1$.

For $i\leq L$: the algorithm first sets lower bounds on $(\pi^*_{i+1,j})_{j=1}^i$ that are pointwise below 
the respective $\pi^*_{i,j}$.  
The setting of final values will set $\pi^*_{i+1,i+1}\gets \delta$ and for $h=1,\ldots,\min\{i,L\}$:
\[ \pi^*_{i+1,i+1-h}\gets e^{\varepsilon} \pi_{i,i+1-h}\enspace .
\]

For $L+1\leq i \leq 2L$:
We have that $\overline{\pi^*_{i+1}} = \pi^*_{i+1,0} =  e^{-\varepsilon}(\pi^*_{i,0} -\delta)$. 
Therefore, the lower bounds the algorithm computes for $(\pi^*_{i+1,j})_{j=1}^i$ are respectively
$e^{-\varepsilon} \pi^*_{i,j}$.  This is consistent with the prefix for $j\leq i-L$. The setting of final values will be according to the suffix until 
the sum is equal to $\pi^*_{i+1}$, which will be the case when $j=i-L$.

For $i\geq 2L+1$:  In this case $\pi^*_i=1$ and $\pi^*_{i+1} =1$ and therefore $\pi^*_{i,0}=0$.
We have from our assumption $\pi^*_{i,j}=0$ for $j < i-(2L)$.  Therefore, the respective lower bounds on
$\pi^*_{i+1,j}$ are $0$.  We have $\pi^*_{i,i-2L}=\delta$  and thus the lower bound $\pi^*_{i+1,i-2L}$ is $0$ and we obtain lower bounds  
The lower bounds on $\pi^*_{i+1,i+1-h}$ for $1 \leq h < 2L$  will respectively be
$e^{-\varepsilon} \pi^*_{i,i+1-h}$.
The setting of final values will set $\pi_{i+1,i+1}\gets \delta$ and set for $h=1,\ldots,L$:
\[ \pi_{i+1,i+1-h}\gets e^{\varepsilon} \pi_{i,i+1-h}
\]
At that point we allocated all of the probability mass of $1$ and have the claimed form \eqref{form:eq} for row $i+1$.
\end{proof}

\section{Frequency Sanitizer with Maximum Separation} \label{general:sec}

  We express a general form of frequency sanitizers and then establish that the sanitizer of Algorithm~\ref{alg:Creport} can be discretized and has the stated properties.
  
\subsection{General Form of Frequency Sanitizers}
We express a general form of {\em frequency sanitizers} (frequency reporting schemes),
in order to facilitate a discussion of when a scheme is optimal.
We consider the sparse case, where a scheme never reports keys with frequency $0$.
A scheme is designed for a given non-decreasing sampling probabilities $(q_i)_{i\geq 1}$ and privacy parameters $(\varepsilon,\delta)$ as in Algorithm~\ref{alg:sanitizer} and is  specified by
 a sequence $(f_i)_{i\geq 0}$ of probability density functions (PDF) over the support $\Re_{\geq 0}$.
 The PDF $f_i$ is the distribution of end-to-end reported values for a key with frequency $i\geq 0$. Each $f_i$ has discrete probability mass
 at the point $x=0$, which corresponds to the event that the key is not reported.
 To simplify presentation, we denote the value of the discrete point mass by $f_i(0)$ and assume that there is continuous mass density on $(0,\infty)$.  Note that for realizability we must have $f_i(0) \geq 1-q_i$. In particular, the function $f_0$, which specifies the reporting function for a key with frequency $0$, has a discrete mass of $1$ at $x=0$ and is $0$ elsewhere (in our sparse case, keys with frequency $0$ are not sampled or reported). 
 The end-to-end sampling and sanitization is equivalent to drawing independently, for each key $x$ with frequency $i$ in the dataset, $j_x\sim f_i$, and reporting  $(x,j_x)$ if $j_x>0$.
 
 The sanitization of a set of {\em sampled} key and frequency pairs is performed as in Algorithm~\ref{alg:sanitizer}, with the following distributions $f_i^{-}$ playing the role of the discrete $p_{i\bullet}$:  $f_i^{-}(0) := (f_i(0)-(1-q_i))/q_i$ and $f_i^{-}(x) = f_i(x)/q_i$ otherwise. The sanitization processes 
 each (sampled) key $x$ with frequency $i$ by drawing independently a sanitized frequency $j_x\sim f_i^{-}$ and if $j_x>0$, reporting the pair $(x,j_x)$. The reader can verify that the end-to-end distributions $(f_i)$ are equivalent to sampling with $(q_i)$ and sanitizing with $(f^{-}_i)$.
For each $f_i$, we denote the respective Cumulative Distribution Function (CDF) by $F_i$  and the inverse CDF by the sets $F_i^{-1}(\alpha) := \{ x \mid \{F_i(x) = \alpha\}\}$,  defined for $\alpha\in [f_i(0), 1)$.

  \begin{lemma} \label{DPforf:lemma}
A frequency sanitizer specified by $(f_i)_{i\geq 0}$ is $(\varepsilon,\delta)$-DP if and only if for any measurable set $J \subset \Re_{\geq 0}$
    \begin{align}
\forall i\geq 0, \forall J\subset  \Re_{\geq 0},\ & \int_J f_i(x) dx \leq e^{\varepsilon} \int_J f_{i-1}(x)dx + \delta \label{cprehalf:eq}\\
  \forall i\geq 0, \forall J\subset  \Re_{\geq 0},\ & \int_J f_i(x)dx \geq e^{-\varepsilon}(\int_J f_{i-1}(x) - \delta)  \label{cposthalf:eq}
  \end{align}  
  (Note that we allow $J$ to include or exclude the discrete point mass at $x=0$) 
  \end{lemma}
  \begin{proof}
  Variant on the proof of Theorem~\ref{theorem:reported-freq-dist} 
  \end{proof}

\subsection{A Refined Frequency Sanitizer}

Algorithm~\ref{alg:Creport} specifies PDFs $(f_i)$ for a frequency sanitizer. The scheme is a refinement of the scheme of Algorithm~\ref{alg:reportfreq} and, as we shall see, for any $(q_i)$ and $(\varepsilon,\delta)$, it maximally  separates sanitized values for different frequencies. 
At a high level, the sub-optimality in the $(\pi_{i,j})$ of Algorithm~\ref{alg:reportfreq} stems from frequencies $i$ for which 
(i)~$R$ is not exactly exhausted at $U$ for some output value 
or (ii)~there is no prefix of discretized outputs $\pi_{i\bullet}$ where exactly 
$\delta = \sum_{j=1}^{b_i} \pi_{ij} + \max\{0, e^{-\varepsilon} \pi_{i-1,0} - \pi_{i,0} \}$.   The scheme of Algorithm~\ref{alg:Creport} (which can also be discretized) introduces additional outputs so that the above breakpoints align wholly with outputs.
Note that for the special case of $(\pi^*_{i,j})$ with integral $L$, Algorithm~\ref{alg:reportfreq} (Lemma~\ref{formij:lemma}) yields the same scheme as Algorithm~\ref{alg:Creport}. This because $b_i$ and $c_i$ are always integral and no new breakpoints are introduced by Algorithm~\ref{alg:Creport}.

\begin{theorem} \label{fsanitizedfreq:theorem}
The $(f_i)$ computed by Algorithm~\ref{alg:Creport} and the sanitizer they specify satisfy:
\begin{enumerate}
    \item 
    For all $i$, $f_i(0) = 1-\pi_i$.
    \item
The sanitizer is $(\varepsilon,\delta)$-DP. 
\item
For each $i$, there is a $c_i\geq 0$ such that subject to the above and $f_{i-1}$, 
$\int_J f_i(x) dx$ is maximized for all $J=(z,i-1]$ for $z\geq c_i$ and is  minimized for all $J=[0,z]$ for $z\leq c_i$.
\end{enumerate}
\end{theorem}
  \begin{proof}
 We first establish a more limited claim, that for each $i$ there is $c_i$ so that the DP constraints are satisfied for all intervals of the form
  $(x,i]$ for $x>c_i$ and $[0,x]$ or $(0,x]$ for all $x\leq c_i$.  We will also show that under the conditions above, $\int_J f_i(x) dx$ is maximized for all $J=(z,i-1]$ for $z\geq c_i$ and is  minimized for all $J$ of the form $[0,z]$ for $z\leq c_i$.
 
  Each $f_i$ is constructed from $f_{i-1}$ as follows.  
    
    From constraint~\ref{posthalf:eq}, the maximum probability mass that can be placed on $J$ for which $\int_J f_{i-1}(x)dx=0$ is $\min\{\delta,\pi_i\}$.  The construction places this maximum amount on $(i-1,i]$.
    
  We construct a function $f_L:(0,i-1]$ that are the minimum values needed to satisfy
  \eqref{cposthalf:eq} for all intervals  $[0,z]$  or $(0,z]$ for $z\in (0,i-1]$.
  We show that indeed $f_L$ has this property:
  The respective constraints are
  \begin{align*}
      \overline{\pi_i} + \int_{0+}^z f_i(x) dx \geq e^{-\varepsilon}(\overline{\pi_{i-1}} +  \int_{0+}^z f_{i-1}(x) dx -\delta)\\
      \int_{0+}^z f_i(x) dx \geq e^{-\varepsilon}(\int_{0+}^z f_{i-1}(x) dx -\delta)\enspace ,
  \end{align*}
  (where with abuse of notation we use $\int_{0+}^z$ to exclude the discrete mass point at $x=0$).
   Combining, we get
   \begin{equation}
       \int_{0+}^z f_i(x) dx \geq \max\{e^{-\varepsilon}\overline{\pi_{i-1}}-\overline{\pi_i},0 \}+ e^{-\varepsilon}(\int_{0+}^z f_{i-1}(x) dx -\delta)\enspace .
   \end{equation}
   Note that the right hand side could be negative but note it is non-decreasing.  We compute a point $b_i$ so that it is positive for all $z> b_i$ and at most $0$ for $z< b_i$.
  The function $f_L(x)$ specified in the algorithm is equal to the maximum of the two constraints
  \[
  \int_0^z f_L(x) dx  = \max\{e^{-\varepsilon}\overline{\pi_{i-1}}-\overline{\pi_i},0 \}+ e^{-\varepsilon}(\int_{0+}^z f_{i-1}(x) dx -\delta)\enspace .
    \]
    
    Finally, we note that given the placement on $(i-1,i]$,  constraints~\eqref{cposthalf:eq} on intervals $(z,i]$ are satisfied if and only if
    $\int_z^{i-1} f_{i}(x) dx \leq e^{\varepsilon}(\int_z^{i-1} f_{i-1}(x) dx$.  The solution sets them at maximum value (equality) for all $z \geq c_i$.
    
We next establish that    
the frequency sanitizer specified by the $(f_i)$ is $(\varepsilon,\delta)$-DP (satisfies the DP constraints (Lemma~\ref{DPforf:lemma}).

 From the construction, the functions
$(f_i)$ satisfy that for all $i$, there is $c_i \geq 0$, so that 
\begin{align*}
    \forall x\in (c_i,\infty], & f_i(x)\geq e^\varepsilon f_{i-1}(x) \\
    \forall x\in (0,c_i], & f_i(x) \leq e^{-\varepsilon} f_{i-1}(x)\enspace .
\end{align*}

We already established that $(f_i)$ satisfies the constraints \eqref{cprehalf:eq} on sets $J$ that are intervals of the form $(z,\infty)$ and satisfies the constraints \eqref{cposthalf:eq} on intervals of the form
$[0,z)$ (included the discrete point mass at $x=0$) or $(0,z)$ (do not include the discrete point mass).

We will show that the constraints are satisfied for any $J$.
For each set $J$, define the partition $J\setminus\{0\}= J^+ \cup J^-$ where $J^+ = J \cap (c_i,\infty)$ and $J^- = J \cap (0,c_i)$.
Consider the constraint \eqref{cprehalf:eq} for $J$.  
Noting that we always have $\overline{\pi_i} \leq \overline{\pi_{i-1}}$, the constraint will hold for $J$ if it holds for $J^+$.  In turn, the constraint holds for $J^+$ if it holds for $(c_i,\infty)$.  
Consider the constraint \eqref{cposthalf:eq} for $J$.  
If $\overline{\pi_i} > e^{-\varepsilon} \overline{\pi_{i-1}}$ or if $0\not\in J$, the constraints holds for $J$ if it holds for $J^-$.  In turn, it holds for $J^-$ if it holds for $(0,c_i]$.
Otherwise, the constraint holds for $J$ if it holds for $\{0\} \cup J^-$.  In turn, it will hold for $\{0\} \cup J^-$ if it holds for $[0,c_i]$.
\end{proof}

\begin{lemma}
The functions $(f_i)_{i=1}^m$ are piecewise constant on $(0,\infty)$ and have 
at most $3m$ distinct breakpoints in total.
\end{lemma}
\begin{proof}
The specification of $f_i$ on $(0,i]$  is constructed from $f_{i-1}:(0,i-1]$.  A constant value is assigned on $(i-1,i]$.  The construction computes two points $b_i$ and $c_i$ so that $f_{i-1}(x)=0$ on $(0,b_i]$, is a constant 
($e^{-\varepsilon}$) times $f_{i-1}$ on $x\in (b_i,c_i]$ and is a constant 
($e^{\varepsilon}$) times $f_{i-1}$ on $x\in (c_i,i-1]$.  Therefore, $f_i$ has at most 3 more breakpoints than $f_{i-1}$. 
The breakpoints are $\{i\}_{i=1}^m \cup \{b_i\}_{i=1}^m \cup \{c_i\}_{i=1}^m$.
\end{proof}
The sanitizer can be discretized by collapsing intervals between consecutive breakpoints to
discrete output points with the respective probability mass.  The discretization does not impact estimation or privacy:  We can always map back from the discrete sanitized frequencies to "simulate" the respective continuous ones (and vice versa) by drawing uniformly at random from a corresponding interval.

\section{Maximum Separation and Ordinal Statistics}  \label{maxseparation:sec}
      
      \subsection{Maximum Separation}

We define a measure of separation between distributions $f_{i_1}$ and $f_{i_2}$ at a certain quantile value $\alpha$ and show that the $(f_i)$ constructed by Algorithm~\ref{alg:Creport} maximize it pointwise. 

For $(f_i)$ we define for all $i_1 , i_2$ and $\alpha\in (f_{i_2}(0),1)$ 
the respective functions 
\begin{equation}
    T_{i_1,i_2}(\alpha) := F_{i_1}(\inf F_{i_2}^{-1}(\alpha))\ .
\end{equation}

Intuitively, lower values of $T_{i_1,i_2}(\alpha)$ for $i_1>i_2$ and higher values for $i_1<i_2$ mean we can separate better the data frequencies $i_1,i_2$ from their respective sanitized values and have higher probability of a pair being concordant.
We show that the $(f_i)$ constructed by Algorithm~\ref{alg:Creport} yield a scheme that simultaneously optimizes  all $T_{i_1,i_2}(\alpha)$ values:
 \begin{theorem} \label{genfreqreporting:alg}
 The $(f_i)$ constructed by Algorithm~\ref{alg:Creport}, for all $\alpha\in (f_{i_2}(0),1)$, minimize $T_{i_1,i_2}(\alpha)$ for all $i_1>i_2$ and maximize $T_{i_1,i_2}(\alpha)$ for all $i_1<i_2$, $\alpha$, over all DP frequency sanitizers.
\end{theorem}
\begin{proof}
The construction of $f_{i}:(0,i-1]$ from $f_{i-1}:(0,i-1)$ results in  $T_{i,i-1}$  that is invariant to the actual distribution of $f_{i-1}$.
From Theorem~\ref{fsanitizedfreq:theorem},
$f_i(0)$ is at a minimum (that depends only on $f_{i-1}(0)$ and the mass of $1-f_i(0)$ is pushed as high as possible for any prefix.
We now consider $T_{i_1,i_2}$ for $i_1>i_2$ and note that it can be expressed in terms of $T_{i_1, i_1+1},\ldots, T_{i_2-1,i_2}$, using (repeatedly if needed):
\begin{align*}
    T_{i+2,i}(\alpha) &= F_{i+2} (\inf F^{-1}_{i}(\alpha)) \\
    &= F_{i+2} (F^{-1}_{i+1} (F_{i+1} (F^{-1}_i(\alpha)))) \\ 
    &= F_{i+2} (F^{-1}_{i+1}(T_{i+1,i}(\alpha))) \\
    &= T_{i+2,i+1}(T_{i+1,i}(\alpha))\enspace .
\end{align*}
Finally, note from the expressions that when $T_{i+1,i}(\alpha)$ are at a maximum for all $i$ and $\alpha$ then so is $T_{i_1,i_2}$ for all $i_1>i_2$ and $\alpha$.
\end{proof}

\subsection{Ordinal Statistics}
\begin{proof} [Proof of Corollary~\ref{ordinalstat:coro}] 
Conveniently, we can express properties of the order induced by sanitized frequencies using only $T$ and $(f_i(0))_{i\geq 1}$:

The probability that a pair of keys with frequencies $i_1 > i_2$ is concordant is
\begin{equation}
 f_{i_2}(0)(1-f_{i_1}(0))+ \frac{1}{2} f_{i_2}(0) f_{i_1}(0) +  \int_{f_{i_2}(0)}^1 (1-T_{i_1,i_2}(\alpha)) d\alpha  \enspace .  
\end{equation}
For a key $x$, keys $Y$ with $w_y > w_x$ for $y\in Y$ and keys $Z$ with $w_z < w_x$ for $z\in Z$, the probability that all $\{(x,y)\}_{y\in Y}$ and $\{(x,z)\}_{z\in Z}$ pairs are concordant is
\begin{align}
 f_{w_x}(0) & \prod_{y\in Y} (1-f_{w_y}(0)) \frac{1}{2^{|Z|}}\prod_{z\in Z} f_{w_z}(0) + \nonumber \\
 \int_{f_{w_x}(0)}^1 f_{w_x}(\alpha) & \prod_{y\in Y}(1-T_{w_x,w_y}(\alpha))\prod_{z\in Z} T_{w_x,w_z}(\alpha) d\alpha\enspace .
\end{align}
The probability that a set of keys with frequencies $i_1<i_2<\cdots<i_k$ is concordant is a sum over expressions that have a form of a constant that depends on $\{f_{i_j}(0)\}_{j\in [k]}$ times an expression of the form 
$\int_{f_{i_1}(0)}^1 \int_{T_{i_2,i_1}(\alpha_1)}^1 \cdots d \alpha_k\cdots d\alpha_1$.

 
 Note that all these expressions are non-decreasing in $T_{i_1,i_2}(\alpha)$ for $i_1>i_2$ and non-increasing for $i_1 < i_2$.  Therefore,  theses expressions are maximized for the $(f_i)$ of Algorithm~\ref{alg:Creport}.
 
 The expected Kendall-$\tau$ rank correlation increases with the expected number of concordant pairs.  Since the scheme maximizes the expected number of concordant pairs, it also maximizes the expected Kendall-$\tau$ rank correlation. 
\end{proof}

\section{The Biased-Down Estimator} \label{biaseddown:sec}

 \begin{equation} \label{Biasdownest:eq}
    a_j \gets \min_{i \mid \pi_{i,j} > 0}  \frac{g(i) - \sum_{h=1}^{j-1} a_h \pi_{i,h}}{\pi_i - \sum_{h=1}^{j-1} \pi_{i,h}} \enspace .
\end{equation}
The sequence $(a_j)$ is non-decreasing and is guaranteed not to over-estimate\notinproc{ (see details in Appendix~\ref{biaseddown:sec})}.

Note (from the structure of $\pi_{i,j}$) that the minimum is over at most $2L((\varepsilon,\delta)$ values of $i$ and each sum can be over at most $2L(\varepsilon,\delta)$ positive $\pi_{i,h}$ entries
(between $i\pm L(\varepsilon,\delta)$).

\begin{lemma}
The estimator expressed by the sequence \eqref{Biasdownest:eq}
 \begin{equation}
    a_j \gets \min_{i \mid \pi_{i,j} > 0}  \frac{g(i) - \sum_{h=1}^{j-1} a_h \pi_{i,h}}{\pi_i - \sum_{h=1}^{j-1} \pi_{i,h}} \enspace .
\end{equation}
is biased-down and non-decreasing.
\end{lemma}
\begin{proof}
The estimate $a_j$ is always set to be at most the value needed to have an unbiased estimate for $g(i)$ when $a_i=a_j$ for $i\geq j$.  Therefore, the estimate can only be biased down.

Let 
$r_{i,j} = \frac{g(i) - \sum_{h=1}^{j-1} a_h \pi_{i,h}}{\pi_i - \sum_{h=1}^{j-1} \pi_{i,h}}$
and recall that $a_j$ is set to the minimum over applicable $i$ of $r_{i,j}$.
Now note that $r_{i,j}$ is non-decreasing with $j$ because $a_j$ is always set to be at most
$r_{i,j}$.  Since for each $j$ we take a minimum over a set of values that can only be larger, $a_j$ may only increase. 
\end{proof}

\section{Limitations of Private Non-negative Unbiased Estimation} \label{mustnegative:sec}
We show that private weighted sampling schemes with optimal key reporting generally do not admit non-negative and unbiased estimation of frequencies.  The lemma below considers the case when there is no sampling, but the argument extends to sampling schemes where $\pi_i = \pi^*_i$ for an appropriate prefix of the sequence.
\begin{lemma} \label{mustnegative:lemma}
Consider  $q \equiv 1$ and (any) keys and frequencies sanitizer with optimal $\pi^*_i$ reporting of keys.  Then there is no unbiased and nonnegative estimator for frequencies. 
\end{lemma}
\begin{proof}
Consider a sanitized keys and frequencies scheme for $q=1$.
The scheme reports a key with frequency $i$ with (optimal) probability $\pi^*_i$.  When a key is reported, the scheme reports a token as a sanitized frequency.
Let $\mathcal{T}[i]$ be the distribution on output tokens for a key with frequency $i$. To make this a distribution we use the special output $\bot$ for the (probability $\overline{\pi^*_i}$) event that the key is not reported.
Using our notation we have 
\[\Pr_{t\sim \mathcal{T}[i]}[t \not=\bot] = \pi^*_i\enspace .\]

Consider a token $t$ that has positive probability to be reported with $i=1$, that is, 
$\Pr_{z\sim \mathcal{T}[1]}[z=t]>0$.  We argue that for 
any $h\leq L(\varepsilon,\delta)$ (where $L$ is as defined in \eqref{Ldef:eq}) 
\begin{equation} \label{blowup:eq}
    \Pr_{z\sim \mathcal{T}[h]}[z=t] = e^{(h-1)\varepsilon} \Pr_{z\sim \mathcal{T}[1]}[z=t]\enspace .
\end{equation}
The argument follows from the privacy constraints for maintaining the maximum key reporting probabilities of $\pi^*_i$. The maximum probability with frequency $h$ for tokens that are not reported for frequency $h-1$ is $\delta$. Therefore, to have $\pi^*_{h} = \pi^*_{h-1} e^\varepsilon + \delta$ the reporting probability of each token reported for $h-1$ must increase by a factor of at least $e^\varepsilon$. 

We now consider estimation. A general estimator for this scheme returns an estimate with expected value $a_t$ for output token $t$.  Note that any unbiased estimator must be $0$ when a key is not reported ($a_\bot = 0$) and for all $h$ we have:
\begin{equation} \label{unbiased:eq}
    h = \E_{z\sim \mathcal{T}[h]} a_z
\end{equation}

Let $T_1$ be the set of possible output tokens $t \not=\bot$ such that
$\Pr_{z\sim \mathcal{T}[1]}[z=t]>0$.
We have
$\Pr_{z\sim \mathcal{T}[1]}[z\in T_1] = \pi^*_1 = \delta$ and from unbiasedness \eqref{unbiased:eq} with $h=1$:
\[\E_{z\sim \mathcal{T}[1]} a_z = \frac{1}{\delta} \enspace .\]
Consider now the estimates for a key with frequency $h\leq L(\varepsilon,\delta)$.  We use
\eqref{unbiased:eq} to obtain: 
\[
h=\E_{z\sim \mathcal{T}[h]} a_z =  \E_{z\sim \mathcal{T}[h]} I_{z\not\in T_1} a_z + \E_{z\sim \mathcal{T}[h]} I_{z\in T_1} a_z
\]
We will show that we can have that the second term is larger than $h$ which will mean that the first term is negative.
We use \eqref{blowup:eq}:
\begin{align*}
 \E_{z\sim \mathcal{T}[h]} I_{z\in T_1} a_z &= \sum_{t\in T_1} \Pr_{z\sim \mathcal{T}[h]}[z = t] a_t \\
    &= \sum_{t\in T_1} \Pr_{z\sim \mathcal{T}[1]}[z = t] e^{\varepsilon(h-1)} a_t \\
    &= e^{\varepsilon(h-1)} \sum_{t\in T_1} \Pr_{z\sim \mathcal{T}[1]}[z = t] a_t \\
    &= e^{\varepsilon (h-1)} \E_{z\sim \mathcal{T}[1]} a_z = e^{\varepsilon (h-1)}\enspace .
\end{align*}
We obtain that when $h < e^{\varepsilon (h-1)}$ holds, which is the case 
for example when $\varepsilon=1$ and $h=2$, we have $a_t<0$ on some tokens $t$.  This because  the contribution to the expectation of the estimate of frequency $h$ that is only due to outputs $T_1$ already exceed the value $h$.  Therefore, we must have negative values $a_t<0$ on at least some tokens $t\not\in T_1$.
\end{proof}

\section{SbH Baseline: Expressions} \label{expressSbH:sec}
  In this section we derive expressions for inclusion probabilities, bias, and error for the baseline method of Stability-based Histograms~\cite{BunNS19} (SbH) (see Section~\ref{SbH:sec}).  We use these expressions in our empirical and analytical evaluation.

 In this section we treat the weighted sampling probabilities $q_i$ as a continuous function  for $i\geq 0$ and use estimators that are continuous functions of a reported $j$. For consistency with other parts of the paper we maintain the discrete indices notation $i,j$.  For a key with frequency $i$, we express the probability density $\phi_{i,j}$ that the key is sampled and reported with frequency $j \geq T$.
  The distribution $\Lap[1/\epsilon]$ is a combination of $(1/2) \Exp[\varepsilon]$ and $(-1/2) \Exp[\varepsilon]$.
  \begin{align*}
  j \geq i \text{ : }\; & \phi_{i,j} = \frac{1}{2} q_j \varepsilon e^{-\varepsilon (j-i)} \\
  j\leq i \text{ : }\; & \phi_{i,j} = \frac{1}{2}  q_j \varepsilon e^{-\varepsilon (i-j)} \enspace .
  \end{align*}
  
  The respective overall reporting probability for a key with frequency $i$ is
  \[ \phi_i = \int_T^\infty \phi_{i,j} dj \enspace .\]
For estimation, we follow \eqref{invprob:est}.  For reported frequency $j$ to estimate $g(i)$ for a key with frequency $i$ we use:
  \[a_j := \frac{g(j)}{q_j} \enspace .\]
  Note that since this is applied after the privacy transform, the estimator is biased. But for keys with frequencies where $g(j)$ is likely to be close to $g(i)$ and $q_j$ close to $q_i$ this estimate would be closer to a direct use of \eqref{invprob:est} on the original data.
  The expected value and MSE of the estimate for a key with frequency $i$ are:
  \begin{align*}
      \E_i =& \int_T^\infty a_j \phi_{i,j} dj \\
      \MSE_i =& (g(i))^2 \cdot (1-\phi_i)\\
      &+ \int_T^\infty (a_j -g(i))^2 \phi_{i,j} dj\enspace .
  \end{align*}
  Using these per-frequency expressions, we can express the MSE and bias of sum estimators for linear statistics as in Section~\ref{estssanitized:sec}.
  
  \subsection{Explicit expressions}

  Substituting $\phi_{i,j}$ and $a_j$ we obtain explicit expressions in terms of $(q_i)_{i\geq 1}$ and $g()$: 
  
  \begin{equation}\label{phiexpress:eq}
      \phi_i = \begin{cases}
      i > T \text{ : }&  \frac{1}{2} \varepsilon  \left( \int_i^\infty q_j e^{-\varepsilon (j-i)} d j  +  \int_T^i q_j e^{-\varepsilon (i-j)} d j \right) \\
      \multicolumn{2}{c}{ =  \frac{1}{2} \varepsilon \left( e^{\varepsilon i} \int_i^\infty q_j e^{-\varepsilon j} d j   +  e^{-\varepsilon i}  \int_T^i q_j e^{\varepsilon j} d j \right)}\\
      i \leq T \text{ : }\;& \frac{1}{2} \varepsilon e^{\varepsilon i}  \int_T^\infty q_j e^{-\varepsilon j} d j \enspace .
    \end{cases}
   \end{equation}

  The expected value $\E_i$ of the estimate of $g(i)$ is:
   \begin{equation} \label{Eexpress:eq}
      \E_i = \begin{cases}
      i > T \text{ : }&  \\
      \multicolumn{2}{c}{\frac{1}{2} \varepsilon \left( e^{\varepsilon i} \int_i^\infty q_j a_j e^{-\varepsilon j} d j   +  e^{-\varepsilon i}  \int_T^i q_j a_j e^{\varepsilon j} d j \right)}\\
      \multicolumn{2}{c}{ = \frac{1}{2} \varepsilon \left( e^{\varepsilon i} \int_i^\infty g(j) e^{-\varepsilon j} d j   +  e^{-\varepsilon i}  \int_T^i g(j) e^{\varepsilon j} d j \right)}\\
      i \leq T \text{ : }&  \frac{1}{2} \varepsilon e^{\varepsilon i}  \int_T^\infty g(j) e^{-\varepsilon j} d j \enspace .
      \end{cases}
   \end{equation}
   Note that $\E_i$ (and hence the bias $\E_i-g(i)$) does not depend on the sampling $q$.
   We express the expected value for the special case when $g(i) = i$:
   \begin{equation} \label{SbHphiE:eq}
      \E_i = \begin{cases}
      i > T \text{ : }\;&  \frac{1}{2} \varepsilon  e^{\varepsilon i} \varepsilon^{-2} e^{-\varepsilon i}(\varepsilon i + 1) \\ 
      \multicolumn{2}{c}{ +  \frac{1}{2} \varepsilon  \varepsilon^{-2} e^{-\varepsilon i} (e^{\varepsilon i}(\varepsilon i -1) - e^{\varepsilon T}(\varepsilon T -1) )} \\
      & = i -\frac{1}{2} e^{-\varepsilon(i-T)}(T-\frac{1}{\varepsilon}) \\
      i \leq T \text{ : }\;&  \frac{1}{2} \varepsilon e^{\varepsilon i}  \varepsilon^{-2} e^{-\varepsilon T}(\varepsilon T +1) \\
            & =  \frac{1}{2} e^{-\varepsilon(T-i)}(T+ \frac{1}{\varepsilon})\enspace .
      \end{cases}
     \end{equation}

  The MSE (general $q$ and $g$) is:
  \begin{align}
      \MSE_i &= (1-\phi_i)g(i)^2 + \label{MSEexpress:eq} \\
      & \int_T^\infty{q_j(a_j-g(i))^2\PDF_{\Lap[1/\varepsilon]}(j - i)dj}\nonumber \\
      &=    -2g(i)\E_i + (g(i))^2\phi_i + (g(i))^2(1-\phi_i) + \nonumber\\
      & \int_T^\infty{\frac{(g(j))^2}{q_j}\PDF_{\Lap[1/\varepsilon]}(j - i)dj}\nonumber \\
      &= (g(i))^2 -2g(i)\E_i + \nonumber\\
      & \int_T^\infty{\frac{(g(j))^2}{q_j}\PDF_{\Lap[1/\varepsilon]}(j - i)dj}\nonumber \\
      &= \quad(g(i))^2 -2g(i)\E_i + \nonumber\\
       &\begin{cases}
      i > T \text{ :}& \frac{1}{2} \varepsilon  e^{\varepsilon i} \int_i^\infty \frac{g(j)^2}{q_j}  e^{-\varepsilon j} d j   +  \\ 
        &   \frac{1}{2} \varepsilon e^{-\varepsilon i}  \int_T^i \frac{g(j)^2}{q_j}  e^{\varepsilon j} d j \nonumber \\
        i \leq T \text{ :}&  \frac{1}{2} \varepsilon e^{\varepsilon i}  \int_T^\infty  \frac{g(j)^2}{q_j} e^{-\varepsilon j} d j \enspace .\nonumber
        \end{cases}
        \end{align}

  \subsection{Expressions for Private histograms ($q \equiv 1$)} 
We now express the reporting probabilities $\phi_i$ for the case where no sampling is subsequently performed ($q \equiv 1$).  From \eqref{phiexpress:eq} we obtain:
    \begin{equation} \label{SbHphi:eq}
      \phi_i = 1 - \CDF_{\Lap[\frac{1}{\varepsilon}]}(T - i)= \begin{cases}
      i \geq T \text{ :}& 1 - \frac{1}{2\delta}  e^{-(i-1)\varepsilon} \\
      i < T \text{ : }&  \frac{1}{2} \delta e^{\varepsilon(i-1)}
      \end{cases}
  \end{equation}

 By substituting $q\equiv 1$ and $g(i)=i$ in \eqref{MSEexpress:eq} we get
 {\small
 \begin{equation} \label{SbHphiMSE:eq}
      \MSE_i = \begin{cases}
      i > T \text{ : }\;& \frac{2}{\varepsilon^2} -  e^{-\varepsilon(i-T)}(\frac{1}{2}T^2-iT+\frac{i-T}{\varepsilon} + \frac{1}{\varepsilon^2} )\\
      i \leq T \text{ :}& i^2 + e^{-\varepsilon(T-i)}(\frac{1}{2}T^2-iT -\frac{i-T}{\varepsilon} + \frac{1}{\varepsilon^{2}})
  \end{cases}
 \end{equation}      
}

    \subsection{Expressions with sampling}
    
    The sampling schemes we consider are parameterized by $\tau>0$.
    For threshold ppswor sampling
  $q_j = 1-e^{-\tau f(j)}$ or threshold Poisson $q_j = \min\{1, \tau f(j)\}$.
  
      We express $\phi_i$ for ppswor threshold sampling and function of frequency $f(w)=w$:
        {\small
  \begin{equation} \label{SbHphiq:eq}
      \phi_i = \begin{cases}
      i \geq T \text{ :}& 
      1-\frac{1}{2}e^{-\varepsilon(i-T)}-\frac{\varepsilon}{2(\varepsilon+\tau)}e^{-i\tau}\\
      & +I_{\varepsilon\not=\tau} \frac{\varepsilon}{2(\varepsilon-\tau)}(e^{-\varepsilon(i-T) -\tau T} - e^{-\tau i})\\
      & -I_{\varepsilon=\tau}(\frac{1}{2}\varepsilon(i-T)e^{-\varepsilon i}) \\
      i \leq T \text{ : }\;&  \frac{1}{2} e^{-\varepsilon(T-i)}(1 - \frac{\varepsilon}{\varepsilon+\tau} e^{-T\tau})
      \end{cases}
  \end{equation}
 }

 We now consider priority sampling with threshold $\tau$ and $f(w)=w$. We start from expressing the inclusion probability $\phi_i$. Recall that in the non-private case, the inclusion probability of $i$ is $q_i=\min\{\tau i, 1\}$.
 
 If $T \geq \frac{1}{\tau}$,
 \begin{align*}
 \phi_i &= \int_T^\infty{q_j\PDF_{\Lap[\frac{1}{\varepsilon}]}(j - i)dj} \\
 &= \int_T^\infty{\PDF_{\Lap[\frac{1}{\varepsilon}]}(j - i)dj} \\
 &= 1 - \CDF_{\Lap[\frac{1}{\varepsilon}]}(T - i) \\
 &= \begin{cases}
      i \geq T \text{ :}& 1 - \frac{1}{2\delta}  e^{-(i-1)\varepsilon} \\
      i < T \text{ : }&  \frac{1}{2} \delta e^{\varepsilon(i-1)}
      \end{cases}
 \end{align*}
 
 Otherwise, $T < \frac{1}{\tau}$.
 \begin{align*}
 \phi_i &= \int_T^\infty{q_j\PDF_{\Lap[\frac{1}{\varepsilon}]}(j - i)dj} \\
 &= \int_T^{\frac{1}{\tau}}{\tau j \PDF_{\Lap[\frac{1}{\varepsilon}]}(j - i)dj}\\
 &\quad + \int_{\frac{1}{\tau}}^\infty{\PDF_{\Lap[\frac{1}{\varepsilon}]}(j - i)dj} \\
 &= \tau \int_T^{\frac{1}{\tau}}{j \cdot \frac{1}{2}\varepsilon e^{-\varepsilon|j - i|} dj} + 1 - \CDF_{\Lap[\frac{1}{\varepsilon}]}(\frac{1}{\tau} - i)
 \end{align*}
 
To compute the inclusion probability, we consider three cases:
\begin{enumerate}
    \item $i \leq T$. In that case,
    \[
    \phi_i = \frac{\tau}{2}\left(\left(T + \frac{1}{\varepsilon}\right)e^{(i - T)\varepsilon} - \frac{1}{\varepsilon}e^{(i - \frac{1}{\tau})\varepsilon}\right).
    \]
    \item $T < i < 1/\tau$. In that case,
    \[
    \phi_i = \tau \left(i - \frac{1}{2\varepsilon}e^{\varepsilon(i - \frac{1}{\tau})} - \frac{1}{2}\left(T - \frac{1}{\varepsilon}\right)e^{\varepsilon(T - i)}\right).
    \]
    \item $i \geq 1/\tau$. In that case,
    \[
    \phi_i = 1 - \frac{\tau}{2\varepsilon}e^{\varepsilon(\frac{1}{\tau} - i)} - \frac{\tau}{2}\left(T - \frac{1}{\varepsilon}\right)e^{\varepsilon(T - i)}.
    \]
\end{enumerate}

To compute the MSE, we use Eq.~\eqref{MSEexpress:eq}, and need to compute $\int_T^\infty{\frac{(g(j))^2}{q_j}\PDF_{\Lap[1/\varepsilon]}(j - i)dj}$. In our implementation, we wrote functions that evaluate the integrals:
\[
\int{xe^{\varepsilon x}dx} = \frac{1}{\varepsilon^2}e^{\varepsilon x}(\varepsilon x - 1) + C
\]
\[
\int{xe^{-\varepsilon x}dx} = - \frac{1}{\varepsilon^2}e^{-\varepsilon x}(\varepsilon x + 1) + C.
\]
\[
\int{x^2e^{\varepsilon x}dx} = \frac{1}{\varepsilon^3}e^{\varepsilon x}(\varepsilon^2 x^2 -2\varepsilon x + 2) + C
\]
\[
\int{x^2e^{-\varepsilon x}dx} = - \frac{1}{\varepsilon^3}e^{-\varepsilon x}(\varepsilon^2 x^2 + 2\varepsilon x + 2) + C.
\]

Then we considered the following cases in order to compute the MSE. If $i \leq T$, we need to compute $\frac{1}{2}\varepsilon e^{\varepsilon i}\int_T^\infty{\frac{j^2}{\min\{\tau j, 1\}}e^{-\varepsilon j}dj}$, and we have two cases:
\begin{enumerate}
    \item $1/\tau < T$. In that case, the integral $\int_T^\infty{\frac{j^2}{\min\{\tau j, 1\}}e^{-\varepsilon j}dj}$ becomes $\int_T^\infty{j^2e^{-\varepsilon j}dj}$.
    \item $1/\tau \geq T$. In that case, the integral $\int_T^\infty{\frac{j^2}{\min\{\tau j, 1\}}e^{-\varepsilon j}dj}$ becomes
    \[
    \frac{1}{\tau}\int_T^{\frac{1}{\tau}}{je^{-\varepsilon j}dj} + \int_{\frac{1}{\tau}}^\infty{j^2e^{-\varepsilon j}dj}.
    \]
\end{enumerate}

Similarly, if $i > T$, we need to compute
\begin{align*}
\frac{1}{2}\varepsilon e^{\varepsilon i}\int_i^\infty{\frac{j^2}{\min\{\tau j, 1\}}e^{-\varepsilon j}dj}
+ \frac{1}{2}\varepsilon e^{-\varepsilon i}\int_T^i{\frac{j^2}{\min\{\tau j, 1\}}e^{\varepsilon j}dj}.    
\end{align*}
and consider the three cases: (i) $1/\tau < T$, (ii) $T \leq 1/\tau < i$, and (iii) $i \leq 1/\tau$.


\end{document}